\documentclass[11pt,a4paper]{article}
\usepackage[hyperref]{acl2019}
\usepackage{times}
\usepackage{latexsym}
\usepackage{graphicx}
\usepackage{amsmath}
\usepackage[utf8]{inputenc}
\usepackage{xspace}
\usepackage{paralist}
\usepackage{booktabs}

\usepackage{tikz}
\usepackage{tikz-qtree}

\usepackage{url}
\usepackage[normalem]{ulem}

\aclfinalcopy 
\def\aclpaperid{BlackboxNLP 31}  

\newcommand\eg{e.g.\ }
\newcommand\ie{i.e.\ }
\newcommand\footurl[1]{\footnote{\url{#1}}}

\newcommand{\word}{\emph{input state}\xspace}
\newcommand{\words}{\emph{input states}\xspace}
\newcommand{\state}{\emph{output state}\xspace}
\newcommand{\states}{\emph{output states}\xspace}

\def\RR#1{{\color{blue}RR: \it #1}}
\def\DM#1{{\color{red}DM: \it #1}}
\def\JL#1{{\color{magenta}JL: \it #1}}
\def\JLrepl#1#2{{\color{magenta}JL: \sout{#1} \it #2}}
\def\DEL#1{{\color{green}SMAZAT: \it #1}}

\def\JL#1{}
\def\JLrepl#1#2{}
\def\RR#1{}
\def\DM#1{}
\def\DEL#1{}

\title{From Balustrades to Pierre Vinken:\\Looking for Syntax in Transformer Self-Attentions}

\author{David Mare\v{c}ek \and Rudolf Rosa\\
  Charles University, Faculty of Mathematics and Physics \\
  Institute of Formal and Applied Linguistics \\
  Malostransk\' e n\' am\v est\' i 25, 118 00 Prague, Czech Republic \\
  \texttt{\{marecek, rosa\}@ufal.mff.cuni.cz}}

\date{}

\begin{document}
\maketitle
\begin{abstract}
We inspect the multi-head self-attention in Transformer NMT encoders for three source languages, looking for patterns that could have a syntactic interpretation. In many of the attention heads, we frequently find sequences of consecutive states attending to the same position, which resemble syntactic phrases.
We propose a transparent deterministic method of quantifying the amount of syntactic information present in the self-attentions,
based on automatically building and evaluating phrase-structure trees from the phrase-like sequences.
We compare the resulting trees to existing constituency treebanks, both manually and by computing precision and recall.
\end{abstract}

\section{Introduction}



The classical approach to Natural Language Processing used to be complex
pipelines, e.g.~\cite{tectomt:2010, manning:2014, apertium:2011},
consisting of multiple steps of linguistically motivated analyses,
such as part-of-speech tagging or syntactic parsing, using explicit
intermediate representations (\eg dependency trees) to abstract over the underlying texts.

In recent years, this has changed with the introduction of deep neural
end-to-end models, which take raw text as input and produce the desired
output directly. Any intermediate representations of the text
may
emerge during the training of the neural
network, and are hidden to us.


We focus on the encoder part of the Transformer architecture \cite{vaswani2017attention}, applied to neural machine translation (NMT),
as
visualizations presented by the authors suggest that its attention heads capture various phenomena such as syntax, semantic roles or anaphora links.

In this work, we analyze the syntactic properties of the self-attention heads both qualitatively and quantitatively. For the quantitative evaluation, we devise a new technique that quantifies
the amount of syntactic information by explicitly building constituency trees from the attentions and comparing them
with the standard syntactic trees.


%
%
%
%


Section~\ref{sec:visualization} briefly describes the Transformer encoder architecture and the way we
visualize the self-attention matrices using heatmaps.
In Section~\ref{sec:analysis}, we present our findings from an extensive manual inspection of the
heatmaps, identifying several common patterns, including the baluster-like
structures which seem to resemble syntactic phrases.
To avoid
confirmation bias, we proceed by devising a
linguistically uninformed tree extraction algorithm
(Section~\ref{sec:parsing}), which
builds a constituency tree
based solely on the assumption that the balusters
correspond to syntactic phrases.
We analyze the resulting parse trees and compare them with standard
syntactic
trees, both manually and via automatic evaluation. 
%
In Section~\ref{sec:head-selection}, we follow the hypothesis that only some of the attention heads
are ``syntactic'', and try to identify them.

\DEL{ both supervisedly and heuristically
in Section~\ref{sec:head-selection},
but we do note that we may now again be performing wishful observations, as we
are seeking for the syntax quite eagerly. \JL{týhle větě nerozumim}}

\section{Related Work}
\label{sec:related-work}
Initial analyses of syntax captured by neural networks focused on RNNs.
\citet{shi:padhi:knight:2016} examine how much syntax is learned by RNN encoder by freezing
its
weights and using a decoder to predict syntactic trees.
\citet{adi:2016} examine sentence vector representations by training auxiliary classifiers to take sentence encodings and predict attributes like word order.
\citet{linzen2016assessing} assess the ability of LSTMs to learn syntax by predicting verbal numbers. 
\citet{blevins-levy-zettlemoyer:2018} measure the amount of syntax in RNNs by predicting part-of-speech tags and constituent labels.

In the last year, related studies appeared also for the Transformer architecture.
\citet{tang:2018} show the Transformer networks perform better than RNNs on
word sense disambiguation.
\citet{zhang:2018} show that language models use more syntactic and morphological information than translation models.


Recently, \citet{hewitt:2019} tried to find syntactic structures in contextual word representations by training simple models on annotated parse trees, concluding that syntactic trees are embedded both in BERT \citep{bert} and ELMo \citep{elmo} models.
This is also supported by \citet{belinkov:2019}, who successfully trained probes to extract linguistic structures, including syntactic dependencies, from various trained neural networks.

Most existing works train probing models on annotated data (e.g.\ treebanks).
However, such a model may learn
to predict the linguistic structure not because it is captured by the network, but because it can be predicted from features
preserved from the input,
as has been already noted e.g.\ by \citet{belinkov:survey}.
In our work, we try to avoid that risk by not using annotated data for the predictions, but rather looking for structures explicitly present in the network representations.




\RR{Tady je zakomentovaný relwork z Hewitta}


In a study closely related to ours,
\citet{raganato:2018} also observe syntax-like patterns in Transformer encoder self-attentions, and try to extract syntactic trees without using annotated data (except for taking the root node from the gold annotation).
However, they construct dependency trees, while we observe phrase-like rather than dependency-like structures.
Moreover, their findings are somewhat inconclusive,
as the accuracy of the resulting trees is close to the
baseline,
while our results are clearly positive.
A similar approach was already suggested (but not evaluated) in \citep{marecek:blackbox2018}.

\JL{asi by se taky hodilo zmínit, že Transormerem jde dělat i parsing https://arxiv.org/pdf/1807.03819.pdf, https://arxiv.org/pdf/1706.05137.pdf
a navíc teď  vyšel BERT, kterej ukazuje, že když dobře naučíš transformíra, tak v tý reprezentaci 
najdeš cokoli}

\section{Transformer NMT Encoder}
\label{sec:visualization}

In the Transformer architecture, \citet{vaswani2017attention} came up with several important improvements over the classical attention,
including \emph{multi-headed} attention. It features
a set of independent attention heads, each deciding on its own to which states to attend.
This allows each of the heads to specialize to provide a different type of information or feature (similarly \eg to CNN filters).
The encoder typically uses six multi-head self-attention sub-layers.
Each state
on a given layer (\state) is computed from a concatenation of the result of applying a set of attention heads to the states on the previous layer (\words),
passed through a feed-forward layer.
This may allow the encoder to do more advanced multi-step processing, such as aggregating the information about several subwords into one position and then attending to this position on the higher layers.



Another notable feature of the Transformer encoder is the use of residual connections, which transport the source subword embeddings forward, bypassing the self-attention mechanism, and get averaged with the outputs of the self-attention.
This ensures that the \state at each position retains a significant amount of the corresponding source subword embedding, supporting the usual shortcut of assuming that the hidden states can be thought of as representations of the underlying subwords (in the context of the sentence).

\subsection{Encoder Self-Attention Visualization}

We focus
on exploring multi-head self-attentions of the encoder.
We use a natural visualization of self-attention heads using square matrix heatmaps (Figure~\ref{fig:heatmaps}), going from black (attention weight = 0) to white (attention weight = 1). The subwords that correspond to the rows and columns are printed alongside the matrix.
The rows correspond to \states, and the columns to \words;
as the \states attend to \words, the softmaxed attention weights on each row sum to 1.
\DEL{For each sentence, for each layer, and for each head, there is a square matrix of attention weights.}
\DEL{It follows that there cannot be too many bright squares in one row, but there can be many bright squares in one column.}


Note that the visualizations may be deceiving in several aspects.
It is important to understand that the fact that a given head at a given position on a given layer attends to a position of a specific subword does \emph{not} mean that the resulting hidden state will simply contain the representation of that subword, for several reasons:
\begin{compactitem}
\item The input to the self attention is the output of the previous layer, \ie a hidden state, presumably but not necessarily representing the subword at this position to some extent, and usually mixing in information about other subwords in the sentence.
\DEL{\item The Transformer self-attention mechanism does not apply the attention weights directly to the input states, but rather to the so-called \emph{values} of the states, obtained via a linear transformation}
\DEL{\item The output of the self-attention is averaged with the state from the previous layer through residual connections}
\item The hidden states emitted from each layer are the outputs of a feed forward network that takes a concatenation of outputs from all of the heads on that layer as input, and can thus mix them, ignore them, only use parts of them, etc.
\end{compactitem}
%

 

\subsection{Experiment Setup}
We analyze the Transformer NMT encoders for the following three languages: English (en), French (fr), and German (de). We selected those particular languages because they are available in the Europarl corpus\footnote{\url{http://data.statmt.org/wmt18/translation-task/training-parallel-ep-v8.tgz}} \cite{europarl} comprising large high-quality multiparallel data, and because constituency syntax parse trees can be obtained for them by the Stanford parser \citep{stanfordparser} out-of-the-box.\footurl{https://nlp.stanford.edu/software/lex-parser.html}

As we
want to explore
a state-of-the-art setup, we use the Transformer 
model~\cite{vaswani2017attention}
as reimplemented by \citet{neuralmonkey:2018} in the Neural Monkey framework\footnote{\url{https://github.com/ufal/neuralmonkey}} in standard setting:
6 encoder and decoder layers, 16 attention heads, embedding size of 512, hidden-layers' size of 4096, dropout 0.9, and batch size 30.

We train the translator for all 6 source-target language pairs (en-fr, en-de, fr-en, fr-de, de-en, de-fr).\footnote{We
intersect the
English-German and English-French parallel corpora using English as pivoting language.}
From the Europarl corpus, we take first 1,000 sentences as development data, last 1,000 sentences as evaluation data, and the remaining 486,272 sentences for training. Table~\ref{tab:bleu-scores} lists the BLEU scores of the systems.
All inspections and evaluations, both manual and automatic, have been performed on the evaluation data.

\begin{table}
    \centering
    \begin{tabular}{rl|rl|rl}
         en-de & 33.5 & en-fr & 45.2 & fr-de & 24.3 \\
         de-en & 39.8 & fr-en & 42.1 & de-fr & 32.9 \\
    \end{tabular}
    \caption{BLEU scores measured on the test data.}
    \label{tab:bleu-scores}
\end{table}

The data are
tokenized by the Stanford Tokenizer\footurl{https://nlp.stanford.edu/software/tokenizer.shtml}
to make the tokens
consistent with the constituency trees with which we
will
compare our results.
We then build a shared dictionary of 100,000 BPE subword units
\cite{bpe} on the concatenated training data of all three languages,
append an
EOS symbol to
each sentence, and train the translation model.

\DM{Jak se to lisi pro ruzne cilove a pro ruzne zdrojove jazyky?}

\DM{kdyz se to agreguje, tak zjistime, ze vsechno ve vysledku koukalo skoro vsude.}

\section{Manual Analysis of Attention Matrices}
\label{sec:analysis}

On a small sample of 10 sentences and for each language pair, we created the heatmaps for all 16 attention heads of all 6 encoder layers.
Six heatmaps for one sentence from the en$\rightarrow$de encoder are shown in Figure~\ref{fig:heatmaps};
all 96 of them are enclosed in the Appendix.


\begin{figure*}
\includegraphics[width=0.5\textwidth]{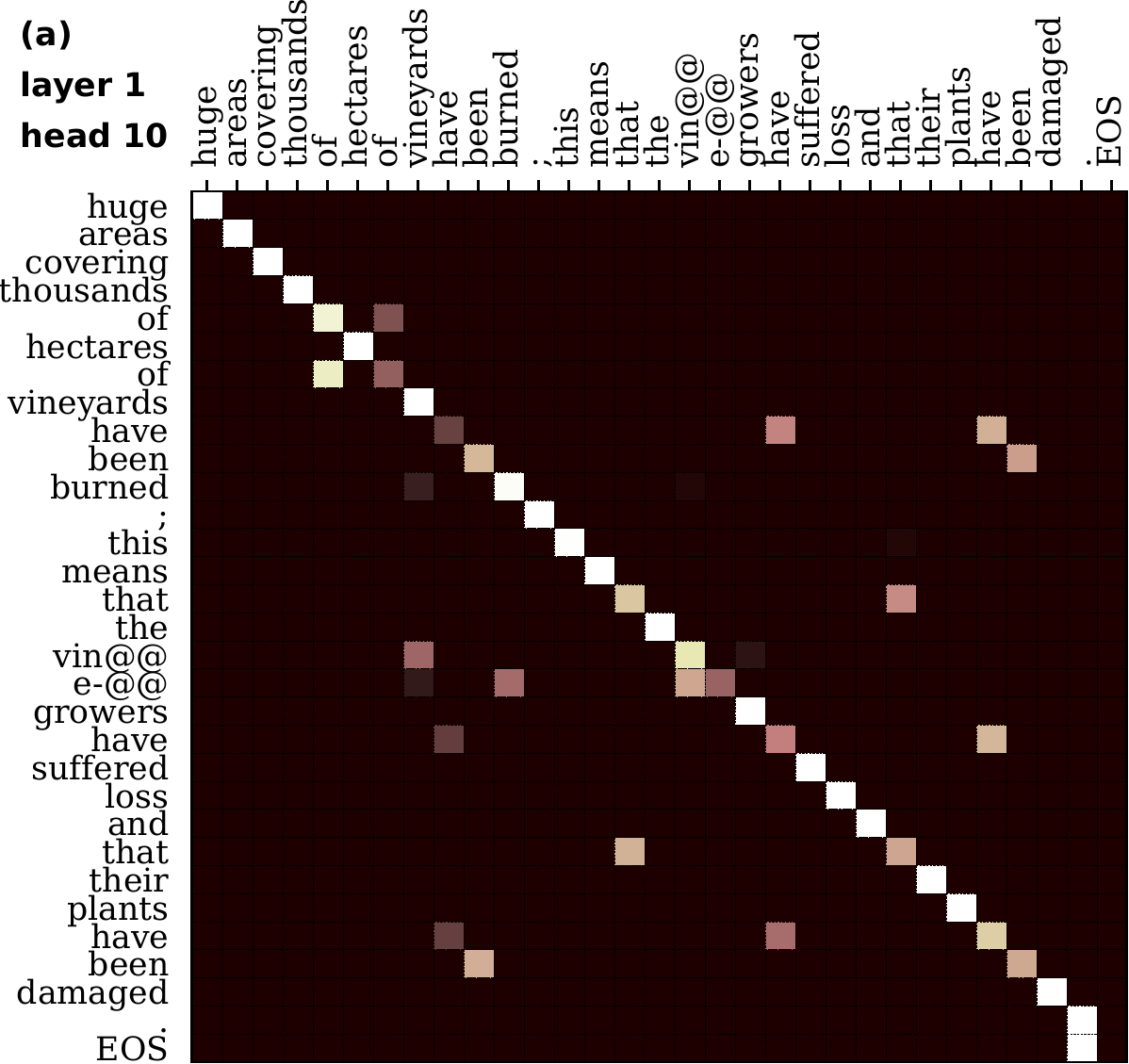}\hspace{2mm}
\includegraphics[width=0.5\textwidth]{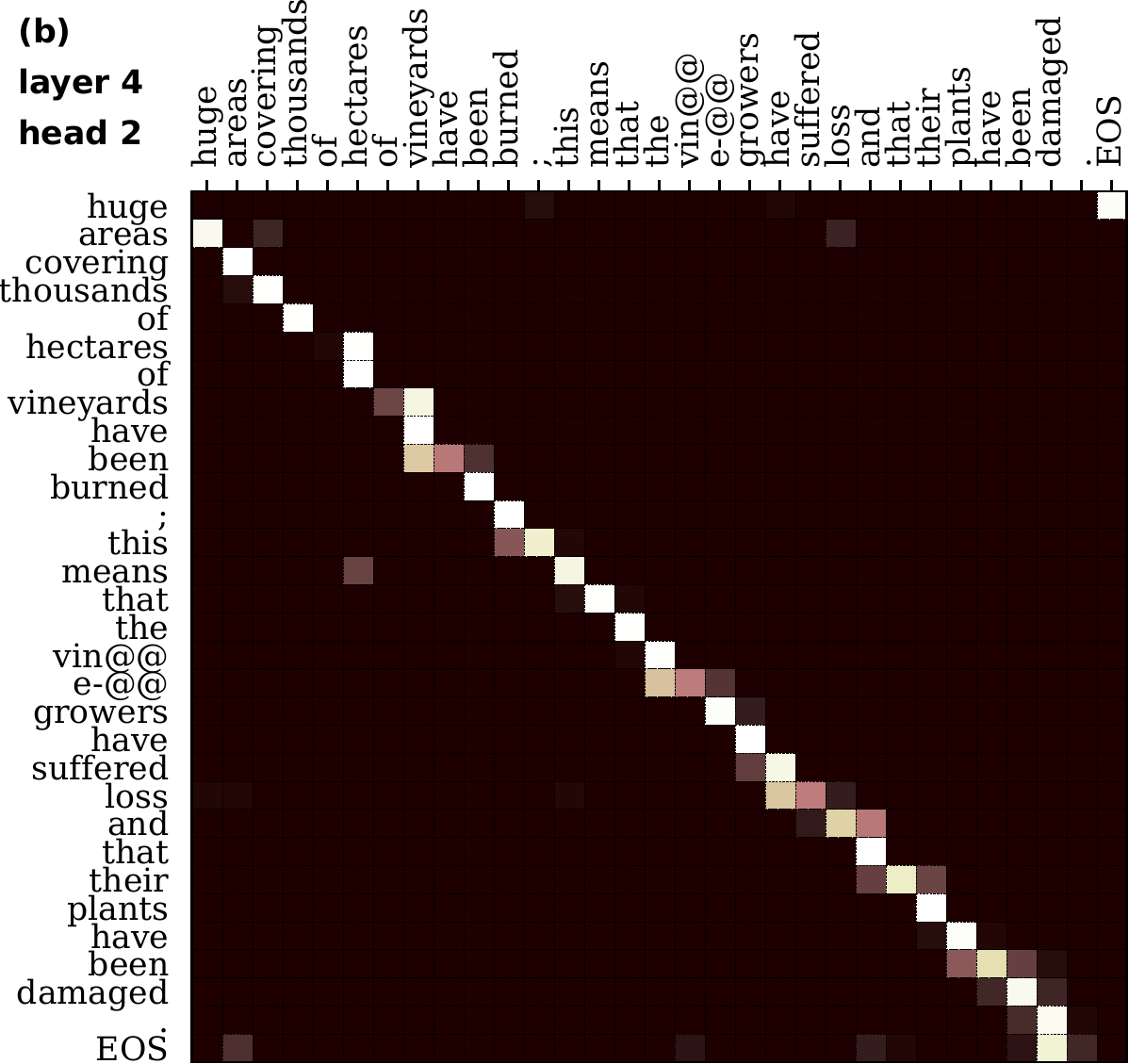}\\
\includegraphics[width=0.5\textwidth]{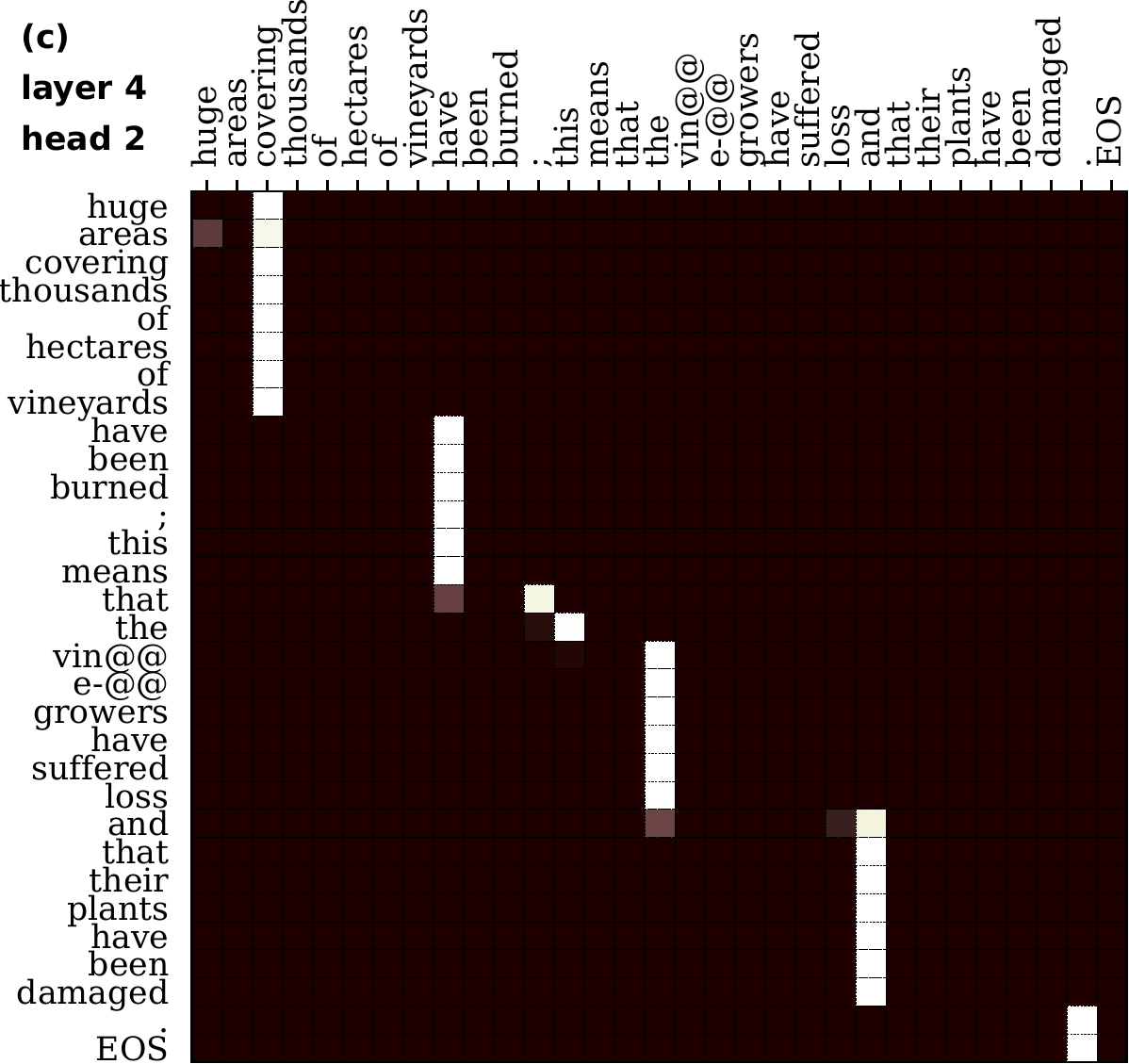}\hspace{2mm}
\includegraphics[width=0.5\textwidth]{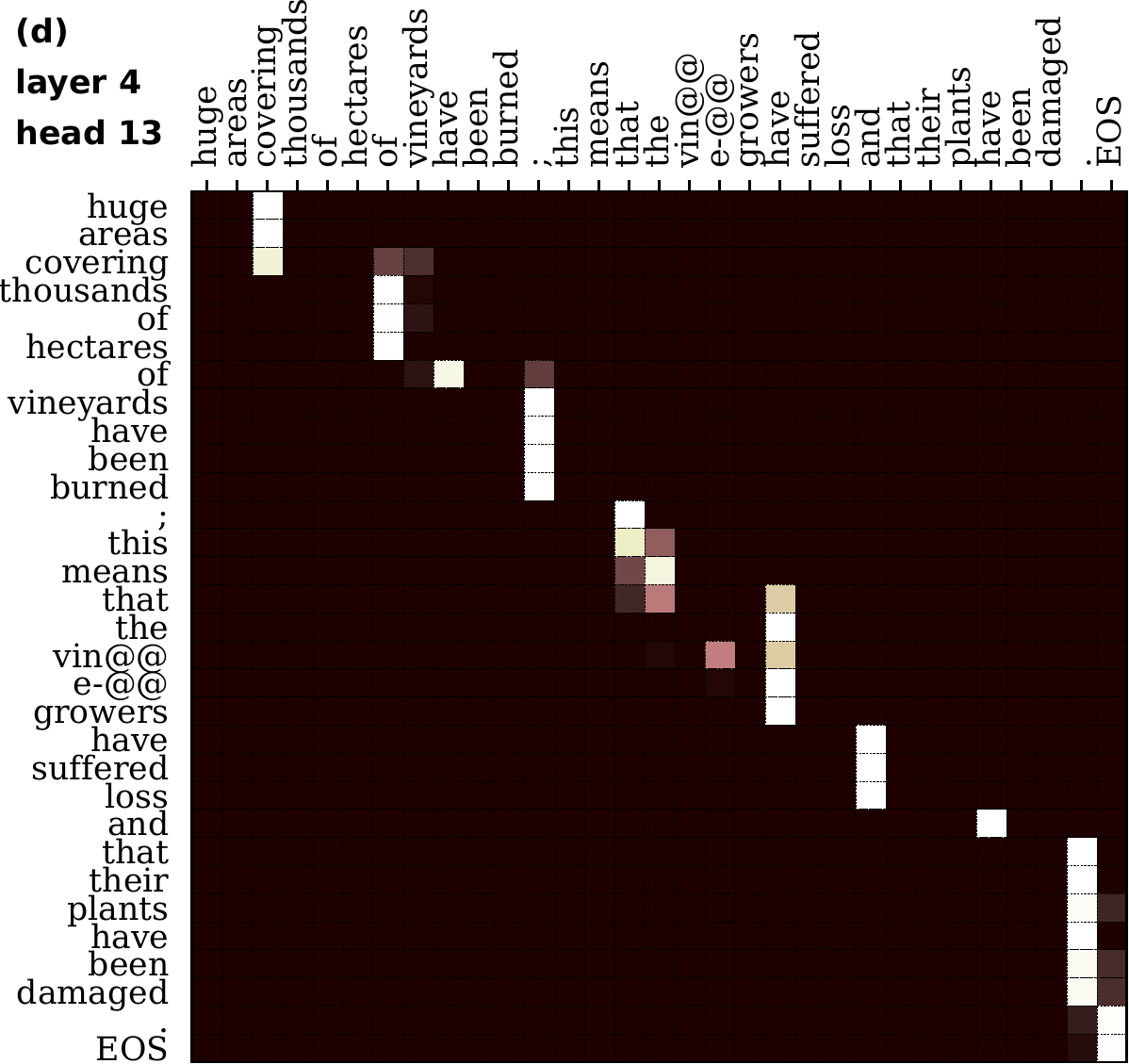}\\
\includegraphics[width=0.5\textwidth]{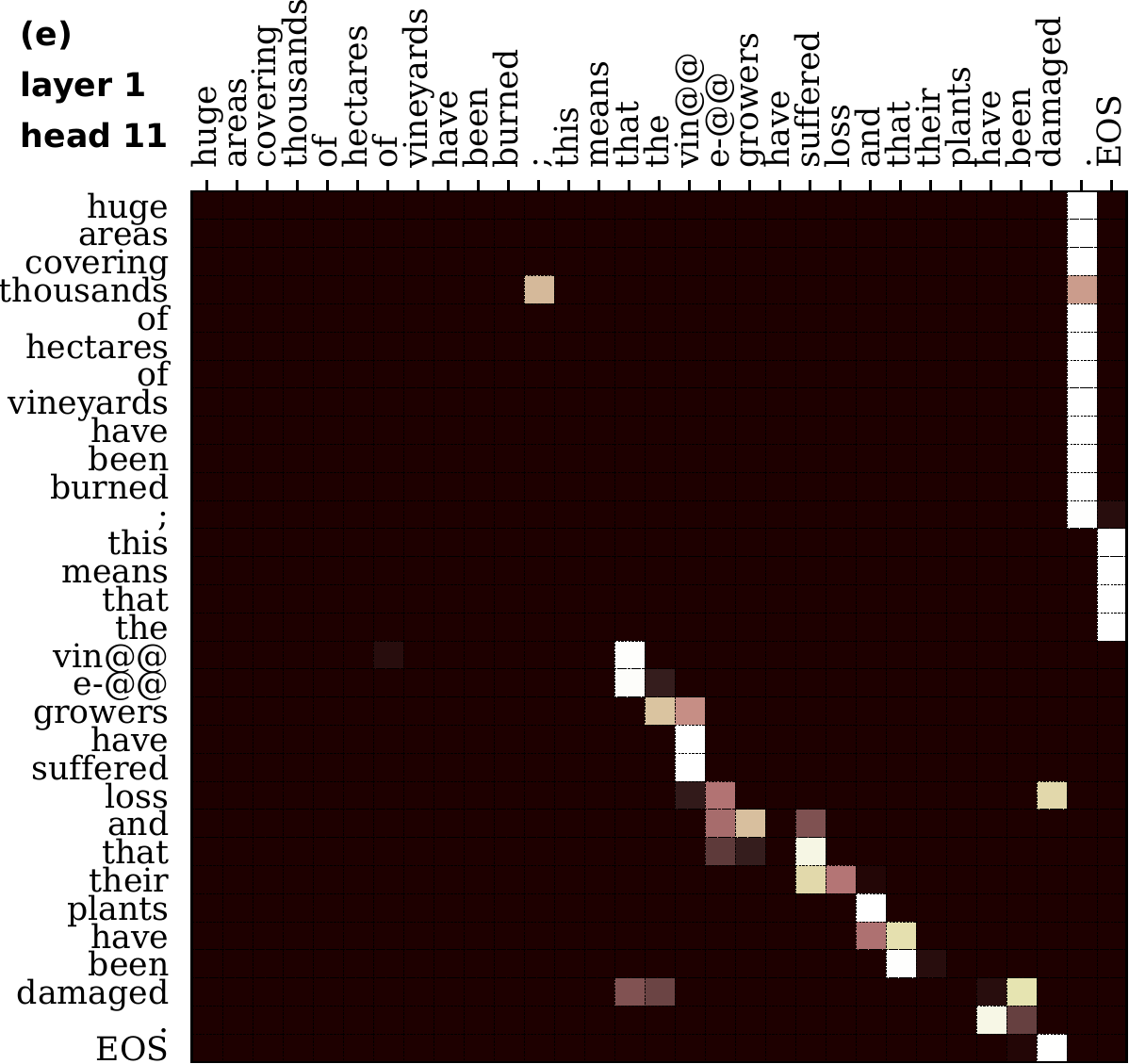}\hspace{2mm}
\includegraphics[width=0.5\textwidth]{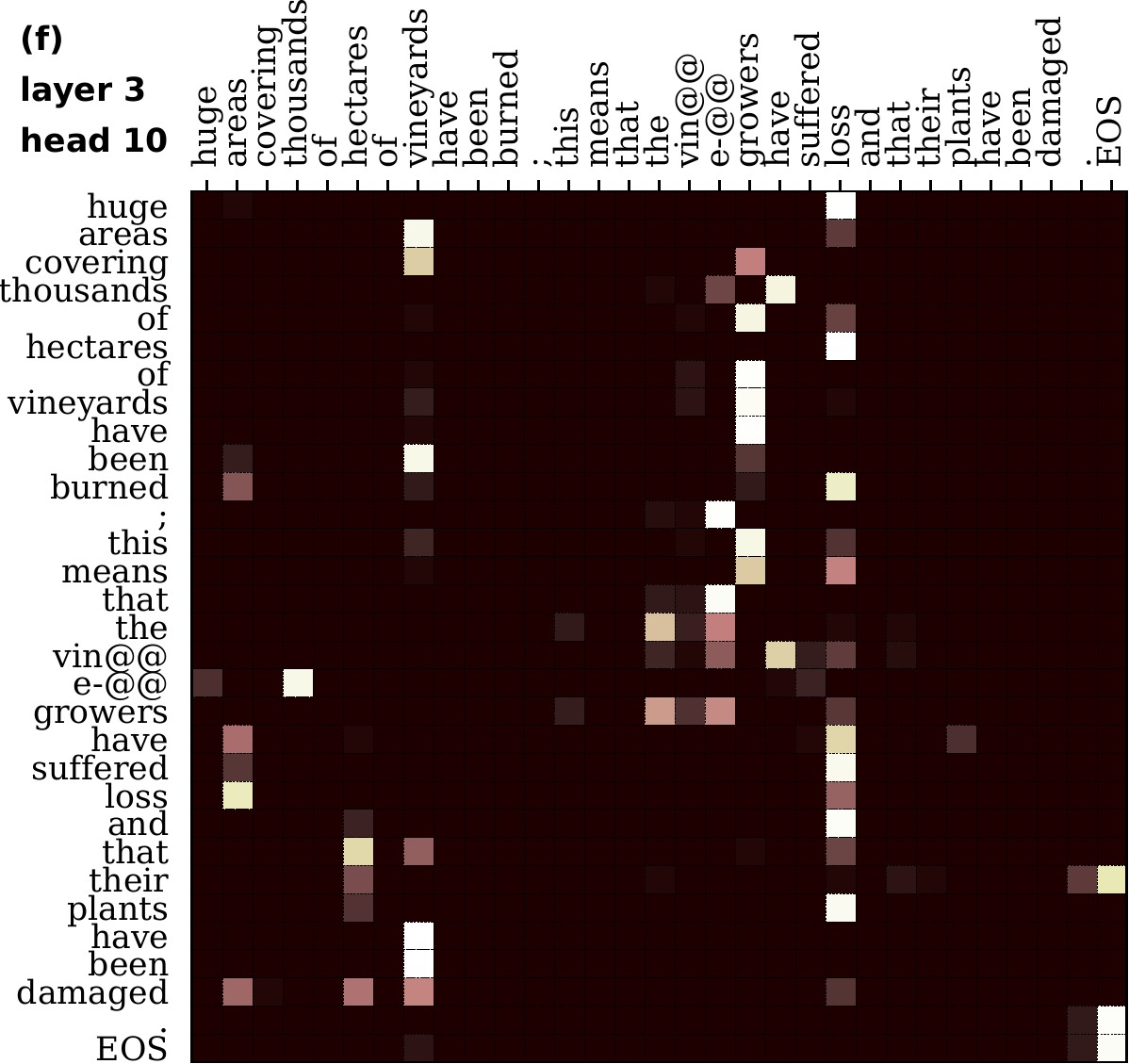}\\
\caption{Heatmaps of selected attention heads showing different patterns.
There are \emph{diagonal} patterns in (a) and (b), \emph{balustrades} in (c) and (d), a combination in (e), and rather scattered attention in (f).}
\label{fig:heatmaps}
\end{figure*}



A general observation is that the attentions are nearly always very peaked. Even though the attention mechanism was designed as soft, most attention heads concentrate nearly all of the attention at each \state onto just one \word.

In the following subsections, we list all of the distinctive patterns that we have identified.\footnote{%
We observe all patterns which \citet{raganato:2018} identified, i.e.\ diagonals and attending to the end of the sentence, but also other patterns which they did not observe.}
An important thing to note is that typically, a head behaves consistently across all sentences, i.e., for a given head on a given layer of a given trained Transformer encoder, we typically see the same attention patterns across all sentences.



\subsection{Diagonals}

Especially at the first encoder layer, there often appear various simple \emph{diagonal} heads.

Typically, each \state attends to the \word at the same position.
This may serve
to pass the subword information to the higher layers.

In some cases, most of the \states attend to the corresponding \words, but some of them attend elsewhere. The role of such \emph{partial diagonal} may be looking for a specific phenomenon that only occurs for some of the \states.

Often,
individual \states attend to preceding or following \words, forming a \emph{parallel diagonal} (Figure~\ref{fig:heatmaps}b).
Sometimes the heads attend further, \eg to the ``pre-previous'' \word.

\subsection{Balustrades}

The most frequent pattern, appearing in about 2/3 of the attention heads, are \emph{balustrades} -- a series of vertical bars, typically placed at the diagonal, which resemble the balusters of a staircase railing.
Examples of such balustrades are shown in Figure~\ref{fig:heatmaps}c,d,e.
The balustrades are often placed upwards or downwards from the main diagonal.
\DEL{Sometimes they are crossing it or floating somewhere else.}

We observe that different heads contain balusters of different lengths. 
For longer balusters, the \word that they attend to often corresponds to a punctuation or a conjunction; often there are also heads that attend exclusively or almost exclusively to the sentence-final punctuation.

We have noticed that in many cases, the sequence of subwords spanned by a baluster may be understood as a syntactic phrase (\eg a noun and its determiner, or a syntactic clause between two commas).
Furthermore, by looking at multiple attention heads at once, we can interpret the balusters of various lengths spanning the same subwords as shorter phrases nested within longer phrases.
This leads us to the idea of constructing a constituency tree from the nested phrases, and comparing it with classical syntactic constituency trees (see Section~\ref{sec:parsing}).

\subsection{Equal or Similar Subwords}
There is typically one or two heads where each \state attends to all instances of the same subword, usually with a more or less uniform distribution (see the subwords ``of'', ``have'' and ``that'' in Figure~\ref{fig:heatmaps}a).
\RR{Hypothesis: position-independent word identity (like a bag of words).}
We have also seen these heads to sometimes attend to very similar but not identical subwords (\eg singular and plural).

\DEL{\subsection{Not Peaked}
Quite rarely, we see heads that are not as peaked, distributing the attention over multiple \words, usually consecutive or close to each other.
We believe that this is simply a way of obtaining a representation of a whole phrase, but we do not have any explanation for why this happens in some cases.
We frequently observe such heads only in the German-to-French translation pair.
}


\subsection{The Rest}

Admittedly, for about 1/5 of the attention heads, we have not identified any clear pattern, and thus have no hypothesis as for the function of such heads.
Sometimes, the head shows some of the behaviours only for some of the \states; sometimes we do not see even such partial patterns (Figure~\ref{fig:heatmaps}f).

\DM{TODO: Quantitative evaluation: try to say how many patterns of type X are present for lang pair Y-Z
(and potentially also try to interpret this)}

\section{Extracting Constituency Trees}
\label{sec:parsing}


Our aim is to analyze whether syntactic structures seem to be captured by Transformer self-attentions, to what extent, and of what kind.
As explained in the previous section, we often observe balusters of various lengths in the attention heatmaps, which can be interpreted as nested syntactic phrases.
In this section, we try to measure
to which extent this interpretation seems to be valid.

For this purpose, we devise a linguistically uninformed transparent deterministic algorithm to extract binary constituency trees from the balusters (Section~\ref{sec:algorithm}).
We automatically evaluate the results by comparing them with classical syntactic trees, generated by a standard syntactic parser (Section~\ref{sec:eval}), to see whether the observed structures seem to capture syntax as we know it.
We discuss the results in Section~\ref{sec:results}.




\subsection{Tree Extraction Algorithm}
\label{sec:algorithm}




We now explain how we construct constituency trees from the balusters in the attention matrices.

Our goal is not to optimize our algorithm towards producing good syntactic trees. Rather, we try to keep our algorithm linguistically uninformed, to reveal only what really is captured by the self-attentions.
Therefore, we:
\begin{compactitem}
\item build binary constituency trees, as this is quite a
basic
way to represent nested phrases,
\item use information from all attention heads, not only those which seem to capture syntax,
\item keep the number of other hyperparameters minimal and set them to the most uninformed values, rather than tuning them,
\item do not train or tune the tree extraction in any way (unlike most related work).
\end{compactitem}

The first step is to identify the balusters.
We have previously described a baluster as a sequence of \states attending to a single \word.
The attentions are typically very peaked, with nearly all of the attention mass concentrated onto one \word.
However, as the attentions are soft, each of the \states in fact attends to all of the \words to some extent.
We thus ``harden'' the soft attention matrix $A'$ by only keeping the maximal attention weight on each row of the attention matrix, setting all the other weights to 0:
\DEL{(see also Figure~\ref{fig:line}):}
\begin{equation}
A_{o, i} = \begin{cases} A'_{o, i} & \text{if } A'_{o, i} = \max_{j \in [1,N]} A'_{o, j}\\
0 & \text{otherwise}\end{cases}
\end{equation}
where $i$ is the \word index, $o$ is the \state index,
and $N$ is the sentence length.

\DEL{
\begin{figure}
\begin{center}
\includegraphics[]{balustrades/line}
\end{center}
\caption{Hardening the attention on one row.}
\label{fig:line}
\end{figure}
}


Next, we extract candidate phrases from the balusters and weight them.
From each baluster, we extract only the candidate phrase corresponding to the full length of the baluster.
The weight of the phrase corresponds to the average attention that \states in the phrase give to the common \word they attend to (\ie the average brightness of the points in the baluster).
If the same phrase appears in multiple attention matrices,
their scores are summed together.
The weight of the phrase spanning the $a$-th to $b$-th subwords thus is:
\begin{equation}
w'_{a,b} = \sum_{h \in H_{a,b}} \frac{\sum_{o \in [a, b]} A^h_{o, i_h}}{b - a + 1}
\end{equation}
where $H_{a,b}$ is the set of attention heads containing a baluster
spanning the \states $a$ to $b$,
$A^h$ is the hardened attention matrix for head $h$, and $i_h$ is the \word attended by the baluster in head $h$.

The weights defined in this way are unbalanced, giving more importance to shorter phrases, as they are more frequent in the attention matrices.
We thus equalize the weights so that the average weight of all phrases of the same length equals 1:
\begin{equation}
w_{a, b} =  \frac{w'_{a, b} \cdot |P^{b-a+1}|}{\sum_{(c, d) \in P^{b-a+1}} w'_{c, d}}
\end{equation}
where $P^k$ is the index pair set of all extracted phrases of length $k$.



\DM{Example of visualized phrase weights is given in Figure
TODO: vizualizace vah kandidatskych frazi.}


To construct the constituency tree from the phrases,
we use the CKY dynamic programming algorithm~\cite{ney:1991}, which searches for the highest scoring constituency tree in $O(n^3)$.

For each tree spanning the $a$-th to $b$-th subword, we define its score $s_{a, b}$ recursively by
finding a separator $k$, $a \le k < b$, that maximizes the average of scores and weights of the two subtrees with spans $(a, k)$ and $(k+1,b)$:
%
\begin{equation}
s_{a,b} = \max\limits_{k} \frac{s_{a,k} + s_{k+1,b} + w_{a,k} + w_{k+1,b}}{4}.
\label{eq:value}
\end{equation}
%
%
The initial scores for single-subword subtrees are set to 1.
The averaging then keeps the scores equalized -- subtrees then have the same power regardless of the size of their spans.

The CKY algorithm works bottom up, starting with the trivial single-subword trees, and then iteratively computing the values of larger subtrees based on the values precomputed in previous steps. Together with the score of each tree, the algorithm also stores the $k$ from Equation~\ref{eq:value}, which defines the highest scoring pair of subtrees covering the same span.
Once the algorithm reaches the tree covering the whole sentence, it recursively returns the highest scoring tree based on the stored values of the highest scoring subtrees.

\RR{Tady je zakomentovanej původní popis, ten můj mi přijde lepší, ale asi taky neni dokonalej, a nejlepší by byl nějakej merge těch dvou.}


\subsection{Automatic Evaluation}
\label{sec:eval}

To evaluate the syntacticity of the Transformer self-attentive encoder,
we extract the constituency trees using our tree extraction algorithm for the 1,000 sentences of our evaluation set;
\DEL{, using all the 6 translation directions;} we will refer to these as \emph{extracted trees}.

We then induce 
syntactic trees for these sentences
with the Stanford Parser.
We use the factored lexicalized parsing models distributed together with the parser, which had been trained on standard constituency treebanks of the languages --
English Penn Treebank \citep{penntb},
German Negra Corpus \citep{negra},
and French Treebank \citep{ftb}.
We post-process the trees in the following way:
\begin{compactenum}
\item remove phrase labels
\item wrap each word into a single-word phrase
\item split words into subwords
\item flatten phrases containing only one immediate subphrase or only one subword
\end{compactenum}
We show an example of applying this procedure:
\begin{compactenum}
\setcounter{enumi}{-1}
\item (S (VP vinegrowers suffer) )
\item ( (vinegrowers suffer) )
\item ( ( (vinegrowers) (suffer) ) )
\item ( ( (vin- e- growers) (suffer) ) )
\item ( (vin- e- growers) suffer)
\end{compactenum}
We will refer to the resulting trees as \emph{parse trees}.

\RR{Ta ukázka neni moc hezky formátovaná. Ale myslim si že je vhodný ji tu mít.}


We compare the extracted trees with the parse trees, assuming that the more similar they are, the more syntactic the Transformer encoder is.




We calculate the \emph{precision} of the extracted tree as the proportion of its phrases that are ``correct'' in the sense that they are consistent with the parse tree,
not crossing any of its phrases.
\DEL{An extracted phrase is correct if, for each phrase in the parse tree, one of the phrases is contained in the other, or they are disjoint; \ie,}
(For the sake of this analysis, we only consider one possible way of capturing syntax, as defined in the respective treebanks; we
discuss that
in Section~\ref{sec:results}.)

Let $P$ be the parse tree, an extracted phrase $e$ is correct if and only if:
\begin{equation}
\forall p \in P: (p \cap e = \emptyset) \vee (p \subseteq e) \vee (e \subseteq p).
\end{equation}

\emph{Recall} is computed inversely, as the proportion of phrases in the parse tree that are consistent with the extracted tree.
We compute the total precision and recall as an average over all extracted phrases in all the trees\DEL{ (\ie not as a macro average over the sentences)},
and also report their harmonic mean (\emph{F1}).

\begin{figure}[t]
\begin{center}
\begin{tikzpicture}[scale=0.7, grow=left]
\tikzset{every leaf node/.style={font=\Large, align=right, anchor=east, text width=55pt}}
\tikzset{level distance=23pt}
\tikzset{edge from parent/.style=
{draw,
edge from parent path={(\tikzparentnode.west)
-- +(-5pt, 0)
|- (\tikzchildnode.east)}}}
\Tree
  [.X
    [.X
      [.X Their plants ]
      [.X have been ]
    ]
  [.X damaged ] ]
\end{tikzpicture}
\begin{tikzpicture}[scale=0.7, grow=left]
\tikzset{every leaf node/.style={font=\Large, align=right, anchor=east, text width=55pt}}
\tikzset{level distance=23pt}
\tikzset{edge from parent/.style=
{draw,
edge from parent path={(\tikzparentnode.west)
-- +(-5pt, 0)
|- (\tikzchildnode.east)}}}
\Tree
  [.X Their
    [.X
      [.X plants have ]
      [.X been damaged ]
    ] ]
\end{tikzpicture}
\end{center}
\caption{Left (\emph{lbal}) and right (\emph{rbal}) balanced binary tree baselines.}
\label{fig:baselines}
\end{figure}
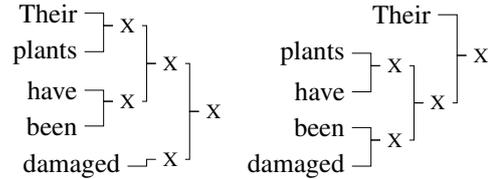

The results of the evaluations for all three source languages are shown in Table~\ref{tab:results}.
To put them into perspective, we also report scores for several uninformed parsing baselines:
\begin{compactenum}
\item \emph{rbal}: balanced binary tree aligned right
\item \emph{lbal}: balanced binary tree aligned left
\item \emph{rand.init}: our proposed algorithm using randomly initialized Transformer weights
\end{compactenum}
Examples of the \emph{lbal} and \emph{rbal} baselines are shown in Figure~\ref{fig:baselines}.

\subsection{Discussion of Results}
\label{sec:results}

\begin{table}[t]
\centering
\emph{English}
\begin{tabular}{l|rrr}
system & precision & recall & F1 score \\
\hline
rbal     & 30.1\% & 24.3\%	& 26.8\% \\
lbal     & 27.8\% &	20.8\%	& 23.8\% \\
rand.init& 25.1\% &	20.0\%	&22.3\% \\
\hline
en $\rightarrow$ de & 35.4\% & 30.6\% & 32.8\% \\
en $\rightarrow$ fr & 35.4\% & 30.2\% & 32.6\% \\
\end{tabular}
\medskip

\emph{German}
\begin{tabular}{l|rrr}
system & precision & recall & F1 score \\
\hline
rbal     & 39.1\% & 31.3\% & 34.8\% \\
lbal     & 38.1\% & 27.6\% & 32.0\% \\
rand.init& 33.7\% &	25.9\% & 29.3\% \\
\hline
de $\rightarrow$ en & 46.1\% & 39.6\% & 42.6\% \\
de $\rightarrow$ fr & 46.7\% & 40.9\% & 43.6\% \\
\end{tabular}
\medskip

\emph{French}
\begin{tabular}{l|rrr}
system & precision & recall & F1 score \\
\hline
rbal     & 34.3\% & 28.7\%	& 31.3\% \\
lbal     & 32.5\% & 25.4\%	& 28.5\% \\
rand.init& 26.1\% & 24.4\%	& 25.3\% \\
\hline
fr $\rightarrow$ en & 44.4\% & 39.7\% & 41.9\% \\
fr $\rightarrow$ de & 46.9\% & 41.7\% & 44.2\% \\
\end{tabular}
\caption{Scores of baseline trees and our extracted trees using all attention heads, evaluated against standard syntactic parse trees.}
\label{tab:results}
\end{table}

The F1 scores of the trees extracted from the attention matrices are 6 to 13 percentage points higher than the best baselines, showing that some syntax
is
indeed captured by the Transformer encoder.

For English, the scores are notably lower than for the other languages.
Manual inspection has shown that this is mostly due to the English parse trees being strongly right-branching, while the other treebanks use flatter, more balanced trees, mainly due to different annotation styles of the treebanks.
The trees extracted from the attention matrices are similar for all of the languages, and resemble the German or French parse trees more than the English ones.
However, a part of the score differences may also be due to a differing syntacticity of the individual encoders, as can be seen from the differing scores for fr$\rightarrow$en and fr$\rightarrow$de.





\begin{figure}[t]
\begin{tikzpicture}[scale=0.7, grow=left]
\tikzset{every leaf node/.style={font=\Large, align=right, anchor=east, text width=55pt}}
\tikzset{level distance=23pt}
\tikzset{edge from parent/.style=
{draw,
edge from parent path={(\tikzparentnode.west)
-- +(-5pt, 0)
|- (\tikzchildnode.east)}}}
\Tree
[.X
  [.X
    [.X
      [.X huge areas ]
      [.X covering [.X [.X thousands [.X of hectares ] ] [.X of vineyards ] ] ] ]
    [.X
      [.X [.X have been ] burned ]
      [.X
        [.X
          [.X [.X [.X ; this ] means ] that ]
          [.X
            [.X [.X the [.X vin- e- ] ] growers ]
            [.X [.X [.X have [.X suffered loss ] ] and ] that ] ] ]
        [.X their plants ] ] ] ]
  [.X [.X [.X have been ] damaged ] [.X . EOS ] ] ]
\end{tikzpicture}
\caption{A constituency tree generated by our tree extraction algorithm from the attention matrices of the en-de encoder for the 4th sentence of the evaluation set.
}
\label{fig:tree}
\end{figure}
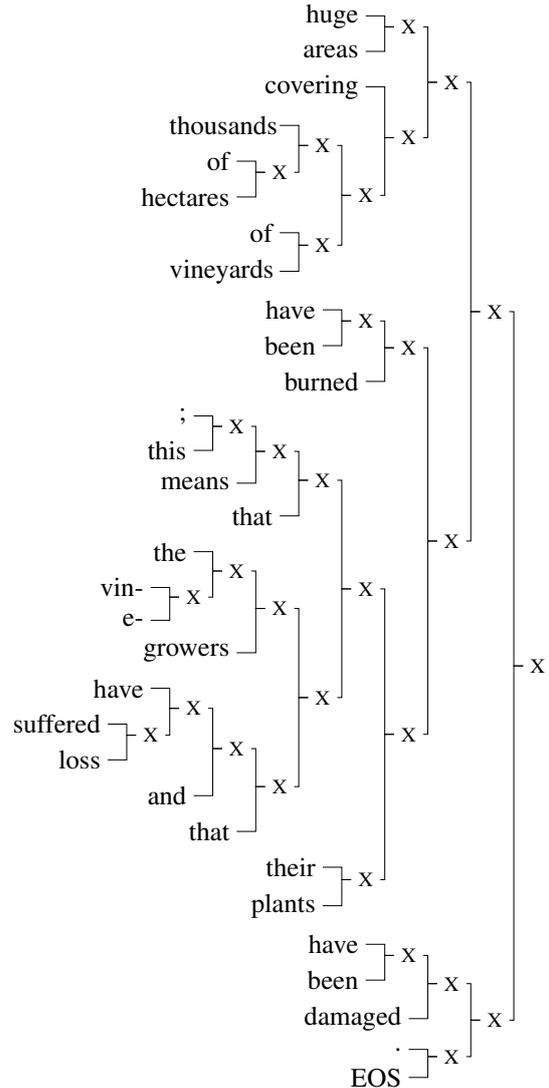

Figure~\ref{fig:tree} shows an example of a tree extracted from the en$\rightarrow$de encoder (the sentence is the same as in Figure~\ref{fig:heatmaps}).
We can see that many of the subtrees seem to make sense syntactically, both smaller ones, such as ``[have been] damaged'',
as well as larger ones, such as the tree spanning ``huge\ldots vineyards''.
Some are questionable, but not necessarily wrong, \eg ``[the vine-] growers''.

A clear limitation of our automatic evaluation method is that it only evaluates whether the structures match those of the syntactic formalism of the standard treebank, but it cannot appreciate alternative structures that also make sense syntactically.
However, this issue is hard to solve without a significant amount of manual work.

Nevertheless, some structures clearly do not correspond to the syntactic structure of the sentence, regardless of the syntactic formalism that we adhere to. E.g.\ 
the phrases ``their plants'' and ``have been damaged'' belong together, but they are separated in the extracted tree all the way to the root.
The reason we find
these incorrect structures in the extracted trees
may be
that we are using all the encoder attention matrices in the extraction algorithm, even though not all of the attention heads seem to behave syntactically; we investigate this to some extent in the next section.
However, it is also quite likely that the encoder only captures some parts of the syntactic structure of the sentence, not a full syntactic tree
-- especially given the fact that the model is trained to do machine translation, and may thus have no reason to capture structures irrelevant for this task.
Moreover, classical syntactic trees are by far not the only possible way of capturing syntax, and it is quite likely that the syntax captured by the self-attentive encoder should be understood differently.\footnote{For example, the syntactic structure could be quite flat, with shorter phrases or treelets joined into a linked list, rather than a complex tree structure with long-distance relations.
Also, we have noted that connectors, such as punctuation and conjunctions, often seem to be part of both of their neighbouring phrases, which could lead to a formalism using partially overlapping phrases.
We intend to investigate this in future.}

\section{Selecting Syntactic Heads}
\label{sec:head-selection}

As we have discussed in Section~\ref{sec:analysis}, there is a range of different types of attention heads.
In our interpretation, some of them, especially the \emph{balustrades}, seem to capture syntactic structures, while others seem not to do so.
A logical step thus is to try to identify the syntactic heads, and only use those for the tree extraction.\footnote{
However, once we start subselecting only some of the heads, we are clearly introducing our expectations about the syntactic structures to be found into the process -- we are now contaminating the so far linguistically uninformed approach with our notion of ``good'' or ``syntactic'' phrases.
}
\DEL{However, there is a big threat hidden in this step.
When using all of the attention heads in the parsing, we truly evaluate what is captured by the heads, without imposing any strong assumptions about the syntax that we want to find in the attentions.
But once we start subselecting only some of the heads, we are clearly introducing our expectations about the syntactic structures to be found into the process -- we are now contaminating the so far linguistically uninformed approach with our notion of ``good'' or ``syntactic'' phrases.
This may be wishful thinking, \ie we might be trying so hard to get syntactic trees from the self-attentions as to find stuff that is not really there. There are 96 heads, so it is likely that there is a way of subselecting some of them to increase the parsing accuracy, but we should be careful with the interpretation.
2 ways: supervised (evaluate against standard trees), heuristic (define a balustradeness measure and select the most balustrady heads)
\subsection{Supervised}
}

We propose to use the automatic evaluation \DEL{of our parse trees against the standard parse trees} as the criterion for selecting the ``syntactic'' heads. We suggest two greedy approaches: \emph{head addition}, and \emph{head ablation}.

In the \emph{head addition} approach, we start with an empty set of heads and then iteratively add the heads one by one, maximizing the precision of the extracted trees in each step, until we have the set of all heads. We then identify the highest scoring head combination that we encountered.

The \emph{head ablation} approach is the logical inverse; we start with all the heads and iteratively remove them until we end up with only one head. 

\DEL{the standard variant of our parsing algorithm, using all of the attention heads to extract phrase scores.
We then try to remove each head, construct the parse trees, and evaluate them against the standard parse trees.
From the evaluated setups, we select the one which achieved the highest score, and iterate, until we end up with only one head.
In each step, we remove the head whose removal leads to the highest increase or lowest decrease of the crossing brackets precision of the resulting tree.
The \emph{head addition} approach is the logical inverse of the head ablation approach: we start with no heads, and 
}

\begin{table}[t]
\centering
\begin{tabular}{l|rrr}
& \multicolumn{3}{c}{improvement in} \\
system & precision & recall & F1 score \\
\hline
en $\rightarrow$ de & +9.48\% & +7.01\% & +8.10\% \\
en $\rightarrow$ fr & +8.43\% & +6.23\% & +7.19\% \\
\hline
de $\rightarrow$ en & +4.60\% & +2.06\% & +3.13\% \\
de $\rightarrow$ fr & +5.96\% & +1.76\% & +3.52\% \\
\hline
fr $\rightarrow$ en & +11.58\% & +8.54\% & +9.91\% \\
fr $\rightarrow$ de & +12.16\% & +8.63\% & +10.20\% \\
\end{tabular}
\caption{Evaluation of syntactic heads subselection. Score gains over the base tree extraction as reported in Table~\ref{tab:results}, in percentage points.}
\label{tab:addhead}
\end{table}

We ran the selection algorithms using only the first 100 sentences.
The setups selected as best by the algorithm were then evaluated on the full evaluation set.
As the \emph{head addition} consistently outperformed \emph{head ablation} by approximately 2 percentage points, we only report the evaluation of the \emph{head addition} in Table~\ref{tab:addhead}.

We can see improvements in F1 ranging from 3 to 10 percentage points,
\DEL{Moreover, the improvements on the 100 development sentences (not shown here) are higher by several percentage points than on the full set.}
showing
that better syntactic trees can be extracted by subselecting the heads.
However, we are perhaps overtuning the setup, and the reported numbers are thus probably somewhat inflated.
Therefore, we are reluctant to draw any strong conclusions from the results.

Nevertheless, the meta-analysis of the heads selected as syntactic is of interest.
For each of the language pairs, between 18 and 32 heads of the total 96 were selected.
However, these are not evenly distributed across the layers.
As we show in Table~\ref{tab:headlayers}, on average, one third of the selected heads come from the first layer, which mostly contains diagonals and short balusters; the last two layers, which contain a lot of balusters of varied lengths, each contributes one fifth of the heads.

\begin{table}[t]
\centering
\begin{tabular}{l|cccccc}
L & 1 & 2 & 3 & 4 & 5 & 6 \\
\hline
P & 36\% & 3\% & 10\% & 10\% & 19\% & 21\% \\
\end{tabular}
\caption{Average proportion of attention head layers in the best subselection setups for all language pairs. $L$ is the number of the layer, $P$ is the proportion of the selected heads that come from the given layer.}
\label{tab:headlayers}
\end{table}



\DEL{
\subsection{Heuristic}
To compute balustradeness: Go over a column, multiply neighbouring squares, sum up, sum up over all columns, divide by N+1 (N=number of tokens) and multiply by 4 (to get into the 0-1 interval).
Separates some of the patterns but not all of them.
Highest are parallel diagonals, ten balustrades or watching the end, then rubbish, then diagonal.}

\section{Conclusion}

We analyzed the Transformer encoder self-attention, identifying baluster structures resembling syntactic phrases.
We devised a transparent linguistically uninformed algorithm for extracting constituency trees from the balusters, compared the resulting trees with standard syntactic parse trees,
and showed that syntax is indeed captured.




\section*{Acknowledgments}
This work has been supported by the grant 18-02196S of the Czech
Science Foundation.

\bibliographystyle{acl_natbib}
\bibliography{ms}

\begin{thebibliography}{25}
\expandafter\ifx\csname natexlab\endcsname\relax\def\natexlab#1{#1}\fi

\bibitem[{Abeill{\'e} et~al.(2003)Abeill{\'e}, Cl{\'e}ment, and
  Toussenel}]{ftb}
Anne Abeill{\'e}, Lionel Cl{\'e}ment, and Fran{\c{c}}ois Toussenel. 2003.
\newblock Building a treebank for french.
\newblock In \emph{Treebanks}, pages 165--187. Springer.

\bibitem[{Adi et~al.(2016)Adi, Kermany, Belinkov, Lavi, and
  Goldberg}]{adi:2016}
Yossi Adi, Einat Kermany, Yonatan Belinkov, Ofer Lavi, and Yoav Goldberg. 2016.
\newblock Fine-grained analysis of sentence embeddings using auxiliary
  prediction tasks.
\newblock \emph{CoRR}, abs/1608.04207.

\bibitem[{Belinkov and Glass(2018)}]{belinkov:survey}
Yonatan Belinkov and James Glass. 2018.
\newblock \href {http://arxiv.org/abs/1812.08951} {Analysis methods in neural
  language processing: {A} survey}.
\newblock \emph{CoRR}, abs/1812.08951.

\bibitem[{Blevins et~al.(2018)Blevins, Levy, and
  Zettlemoyer}]{blevins-levy-zettlemoyer:2018}
Terra Blevins, Omer Levy, and Luke Zettlemoyer. 2018.
\newblock \href {http://www.aclweb.org/anthology/P18-2003} {Deep {RNN}s encode
  soft hierarchical syntax}.
\newblock In \emph{Proceedings of the 56th Annual Meeting of the Association
  for Computational Linguistics (Volume 2: Short Papers)}, pages 14--19,
  Melbourne, Australia. Association for Computational Linguistics.

\bibitem[{Devlin et~al.(2018)Devlin, Chang, Lee, and Toutanova}]{bert}
Jacob Devlin, Ming{-}Wei Chang, Kenton Lee, and Kristina Toutanova. 2018.
\newblock \href {http://arxiv.org/abs/1810.04805} {{BERT:} pre-training of deep
  bidirectional transformers for language understanding}.
\newblock \emph{CoRR}, abs/1810.04805.

\bibitem[{Forcada et~al.(2011)Forcada, Ginest{\'i}-Rosell, Nordfalk, O'Regan,
  Ortiz-Rojas, P{\'e}rez-Ortiz, S{\'a}nchez-Mart{\'i}nez,
  Ram{\'i}rez-S{\'a}nchez, and Tyers}]{apertium:2011}
Mikel~L. Forcada, Mireia Ginest{\'i}-Rosell, Jacob Nordfalk, Jim O'Regan,
  Sergio Ortiz-Rojas, Juan~Antonio P{\'e}rez-Ortiz, Felipe
  S{\'a}nchez-Mart{\'i}nez, Gema Ram{\'i}rez-S{\'a}nchez, and Francis~M. Tyers.
  2011.
\newblock \href {https://doi.org/10.1007/s10590-011-9090-0} {Apertium: a
  free/open-source platform for rule-based machine translation}.
\newblock \emph{Machine Translation}, 25(2):127--144.

\bibitem[{Helcl et~al.(2018)Helcl, Libovick{\'{y}}, Kocmi, Musil, C{\'{i}}fka,
  Vari{\v{s}}, and Bojar}]{neuralmonkey:2018}
Jind{\v{r}}ich Helcl, Jind{\v{r}}ich Libovick{\'{y}}, Tom Kocmi,
  Tom{\'{a}}{\v{s}} Musil, Ond{\v{r}}ej C{\'{i}}fka, Du{\v{s}}an Vari{\v{s}},
  and Ond{\v{r}}ej Bojar. 2018.
\newblock Neural monkey: The current state and beyond.
\newblock In \emph{The 13th Conference of The Association for Machine
  Translation in the Americas, Vol. 1: {MT} Researchers’ Track}, pages
  168--176, Stroudsburg, {PA}, {USA}. The Association for Machine Translation
  in the Americas, The Association for Machine Translation in the Americas.

\bibitem[{Hewitt and Manning(2019)}]{hewitt:2019}
John Hewitt and Christopher~D. Manning. 2019.
\newblock {Structural Probe for Finding Syntax in Word Representations}.
\newblock In \emph{Proceedings of NAACL 2019}.

\bibitem[{Klein and Manning(2003)}]{stanfordparser}
Dan Klein and Christopher~D Manning. 2003.
\newblock Fast exact inference with a factored model for natural language
  parsing.
\newblock In \emph{Advances in neural information processing systems}, pages
  3--10.

\bibitem[{Koehn(2005)}]{europarl}
Philipp Koehn. 2005.
\newblock {Europarl: A Parallel Corpus for Statistical Machine Translation}.
\newblock In \emph{{Conference Proceedings: the tenth Machine Translation
  Summit}}, pages 79--86, Phuket, Thailand. AAMT, AAMT.

\bibitem[{Linzen et~al.(2016)Linzen, Dupoux, and
  Goldberg}]{linzen2016assessing}
Tal Linzen, Emmanuel Dupoux, and Yoav Goldberg. 2016.
\newblock Assessing the ability of {LSTMs} to learn syntax-sensitive
  dependencies.
\newblock \emph{Transactions of the Association for Computational Linguistics},
  4:521--535.

\bibitem[{Liu et~al.(2019)Liu, Gardner, Belinkov, Peters, and
  Smith}]{belinkov:2019}
Nelson~F. Liu, Matt Gardner, Yonatan Belinkov, Matthew Peters, and Noah~A.
  Smith. 2019.
\newblock \href {http://arxiv.org/abs/1903.08855} {Linguistic knowledge and
  transferability of contextual representations}.
\newblock \emph{CoRR}, abs/1903.08855.

\bibitem[{Manning et~al.(2014)Manning, Surdeanu, Bauer, Finkel, Bethard, and
  McClosky}]{manning:2014}
Christopher~D. Manning, Mihai Surdeanu, John Bauer, Jenny Finkel, Steven~J.
  Bethard, and David McClosky. 2014.
\newblock The {Stanford} {CoreNLP} natural language processing toolkit.
\newblock In \emph{Association for Computational Linguistics (ACL) System
  Demonstrations}, pages 55--60.

\bibitem[{Marcus et~al.(1993)Marcus, Marcinkiewicz, and Santorini}]{penntb}
Mitchell~P Marcus, Mary~Ann Marcinkiewicz, and Beatrice Santorini. 1993.
\newblock Building a large annotated corpus of english: The penn treebank.
\newblock \emph{Computational linguistics}, 19(2):313--330.

\bibitem[{Mare{\v{c}}ek and Rosa(2018)}]{marecek:blackbox2018}
David Mare{\v{c}}ek and Rudolf Rosa. 2018.
\newblock Extracting syntactic trees from transformer encoder self-attentions.
\newblock In \emph{Proceedings of the First Workshop on Analyzing and
  Interpreting Neural Networks for {NLP}}, pages 347--349, Stroudsburg, {PA},
  {USA}. The Assotiation of Computational Linguistics.

\bibitem[{Ney(1991)}]{ney:1991}
Hermann Ney. 1991.
\newblock \href {https://doi.org/10.1109/78.80816} {Dynamic programming parsing
  for context-free grammars in continuous speech recognition}.
\newblock \emph{Trans. Sig. Proc.}, 39(2):336--340.

\bibitem[{Peters et~al.(2018)Peters, Neumann, Iyyer, Gardner, Clark, Lee, and
  Zettlemoyer}]{elmo}
Matthew~E Peters, Mark Neumann, Mohit Iyyer, Matt Gardner, Christopher Clark,
  Kenton Lee, and Luke Zettlemoyer. 2018.
\newblock Deep contextualized word representations.
\newblock \emph{arXiv preprint arXiv:1802.05365}.

\bibitem[{Popel and {\v{Z}}abokrtsk{\'{y}}(2010)}]{tectomt:2010}
Martin Popel and Zden{\v{e}}k {\v{Z}}abokrtsk{\'{y}}. 2010.
\newblock Tecto{MT}: Modular {NLP} framework.
\newblock In \emph{Lecture Notes in Artificial Intelligence, Proceedings of the
  7th International Conference on Advances in Natural Language Processing
  (Ice{TAL} 2010)}, volume 6233 of \emph{Lecture Notes in Computer Science},
  pages 293--304, Berlin / Heidelberg. Iceland Centre for Language Technology
  ({ICLT}), Springer.

\bibitem[{Raganato and Tiedemann(2018)}]{raganato:2018}
Alessandro Raganato and J{\"o}rg Tiedemann. 2018.
\newblock \href {http://aclweb.org/anthology/W18-5431} {An analysis of encoder
  representations in transformer-based machine translation}.
\newblock In \emph{Proceedings of the 2018 EMNLP Workshop BlackboxNLP:
  Analyzing and Interpreting Neural Networks for NLP}, pages 287--297.
  Association for Computational Linguistics.

\bibitem[{Sennrich et~al.(2016)Sennrich, Haddow, and Birch}]{bpe}
Rico Sennrich, Barry Haddow, and Alexandra Birch. 2016.
\newblock \href {https://doi.org/10.18653/v1/P16-1162} {Neural machine
  translation of rare words with subword units}.
\newblock In \emph{Proceedings of the 54th Annual Meeting of the Association
  for Computational Linguistics (Volume 1: Long Papers)}, pages 1715--1725.
  Association for Computational Linguistics.

\bibitem[{Shi et~al.(2016)Shi, Padhi, and Knight}]{shi:padhi:knight:2016}
Xing Shi, Inkit Padhi, and Kevin Knight. 2016.
\newblock Does string-based neural {MT} learn source syntax?
\newblock In \emph{EMNLP}, pages 1526--1534.

\bibitem[{Skut et~al.(1999)Skut, Uszkoreit, and Brants}]{negra}
Wojciech Skut, Hans Uszkoreit, and Thorsten Brants. 1999.
\newblock Syntactic annotation of a german newspaper corpus.
\newblock In \emph{ATALA sur le Corpus Annotés pour la Syntaxe Treebanks, June
  18-19}, pages 69--76, Paris, France. o.A.

\bibitem[{Tang et~al.(2018)Tang, M{\"u}ller, Rios, and Sennrich}]{tang:2018}
Gongbo Tang, Mathias M{\"u}ller, Annette Rios, and Rico Sennrich. 2018.
\newblock \href {http://aclweb.org/anthology/D18-1458} {Why self-attention? a
  targeted evaluation of neural machine translation architectures}.
\newblock In \emph{Proceedings of the 2018 Conference on Empirical Methods in
  Natural Language Processing}, pages 4263--4272. Association for Computational
  Linguistics.

\bibitem[{Vaswani et~al.(2017)Vaswani, Shazeer, Parmar, Uszkoreit, Jones,
  Gomez, Kaiser, and Polosukhin}]{vaswani2017attention}
Ashish Vaswani, Noam Shazeer, Niki Parmar, Jakob Uszkoreit, Llion Jones,
  Aidan~N Gomez, Lukasz Kaiser, and Illia Polosukhin. 2017.
\newblock \href
  {http://papers.nips.cc/paper/7181-attention-is-all-you-need.pdf} {Attention
  is all you need}.
\newblock In I.~Guyon, U.~V. Luxburg, S.~Bengio, H.~Wallach, R.~Fergus,
  S.~Vishwanathan, and R.~Garnett, editors, \emph{Advances in Neural
  Information Processing Systems 30}, pages 6000--6010. Curran Associates, Inc.

\bibitem[{Zhang and Bowman(2018)}]{zhang:2018}
Kelly~W. Zhang and Samuel~R. Bowman. 2018.
\newblock Language modeling teaches you more syntax than translation does:
  Lessons learned through auxiliary task analysis.
\newblock \emph{CoRR}, abs/1809.10040.

\end{thebibliography}

\newpage
\appendix
\clearpage

\section*{Appendix: Visualization of all attention heads}
We provide visualisations of encoder's self-attention heads for English source sentence
\emph{``Huge areas covering thousands of hectares of vineyards have been burned; this means that the vin@@ e-@@ growers have suffered loss and that their plants have been damaged.''},
when translating into German.


\begin{figure}[b!]
\begin{center}
\includegraphics[width=0.15\textwidth]{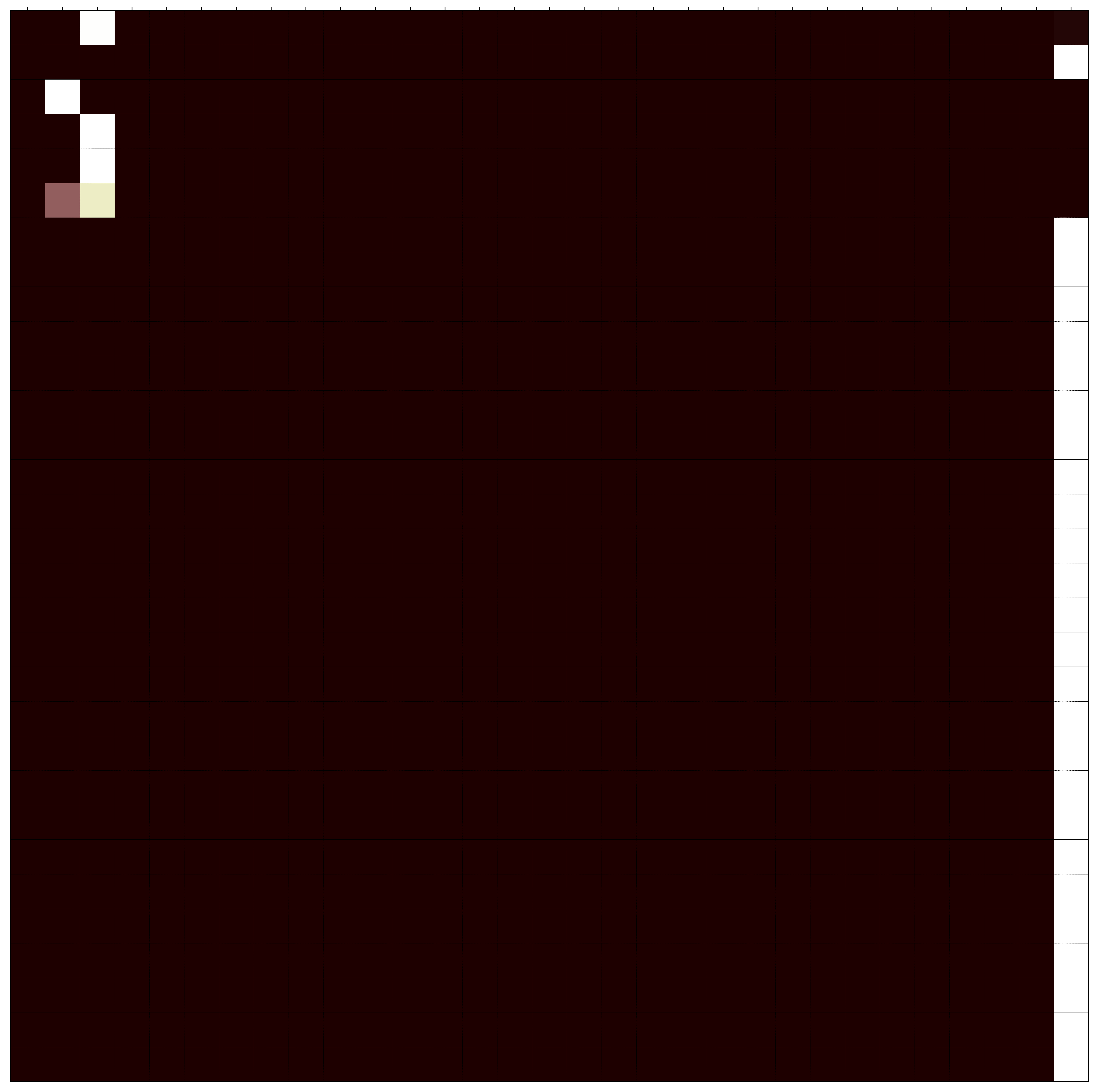}
\includegraphics[width=0.15\textwidth]{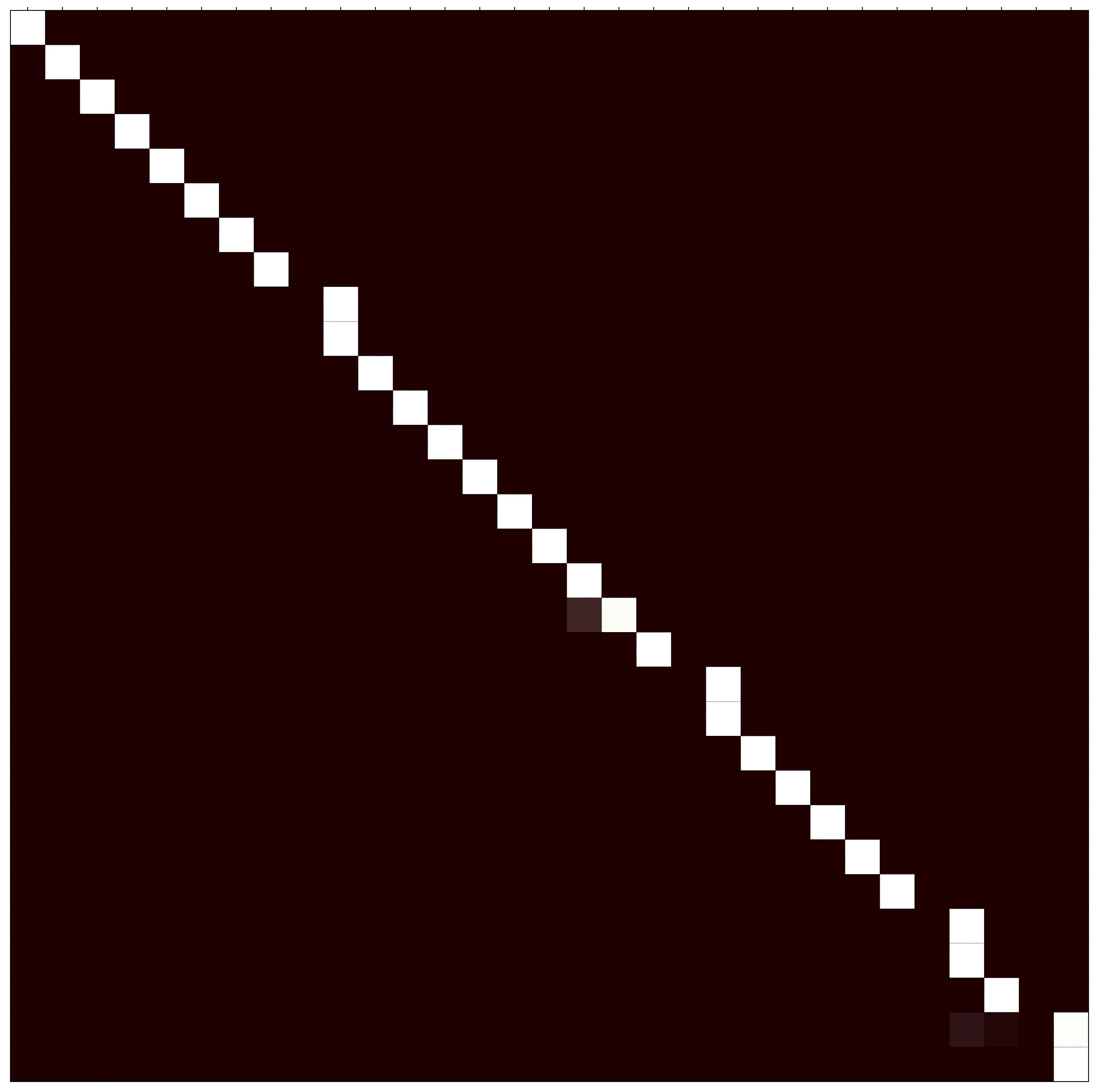}
\includegraphics[width=0.15\textwidth]{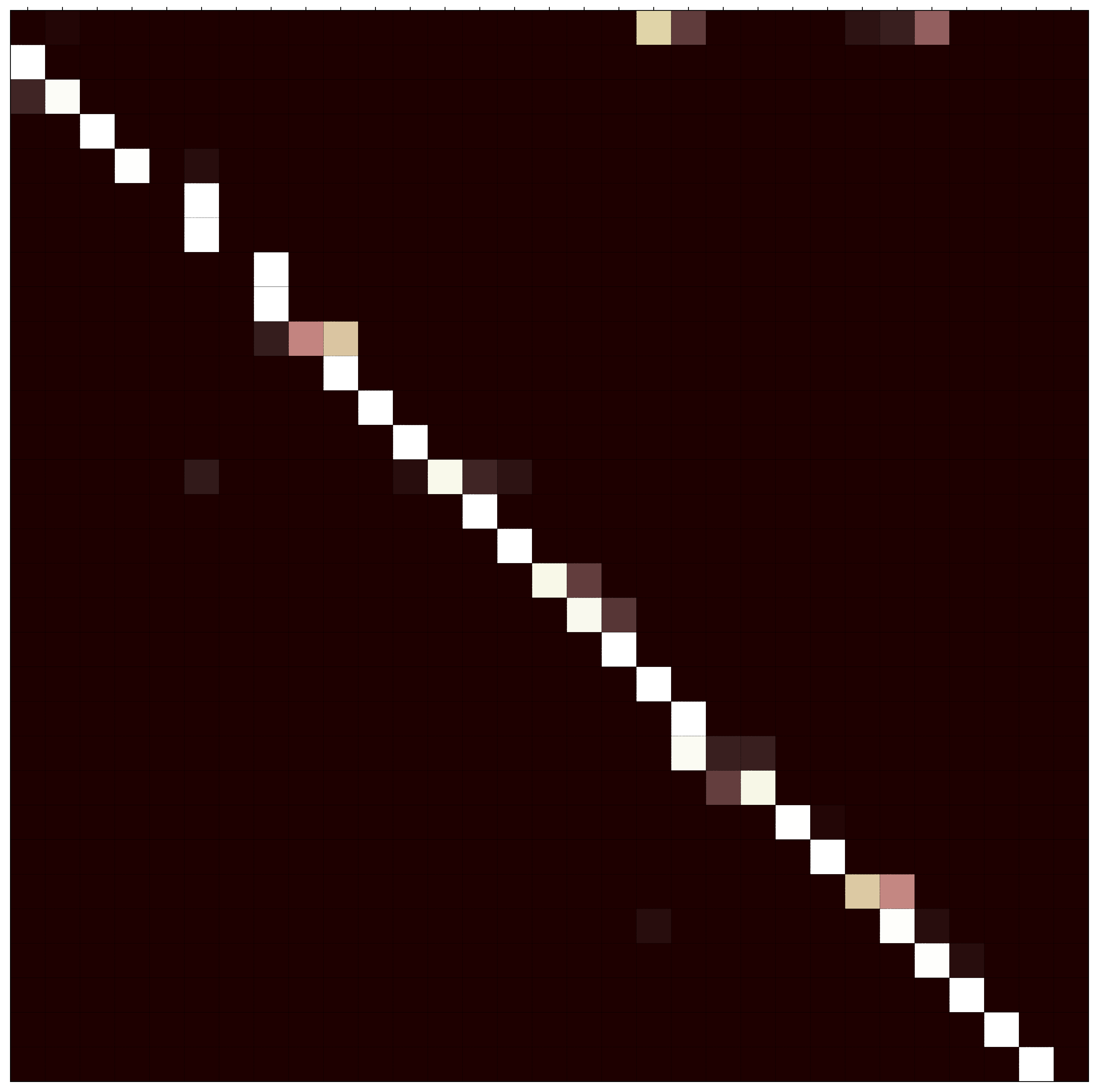}
\includegraphics[width=0.15\textwidth]{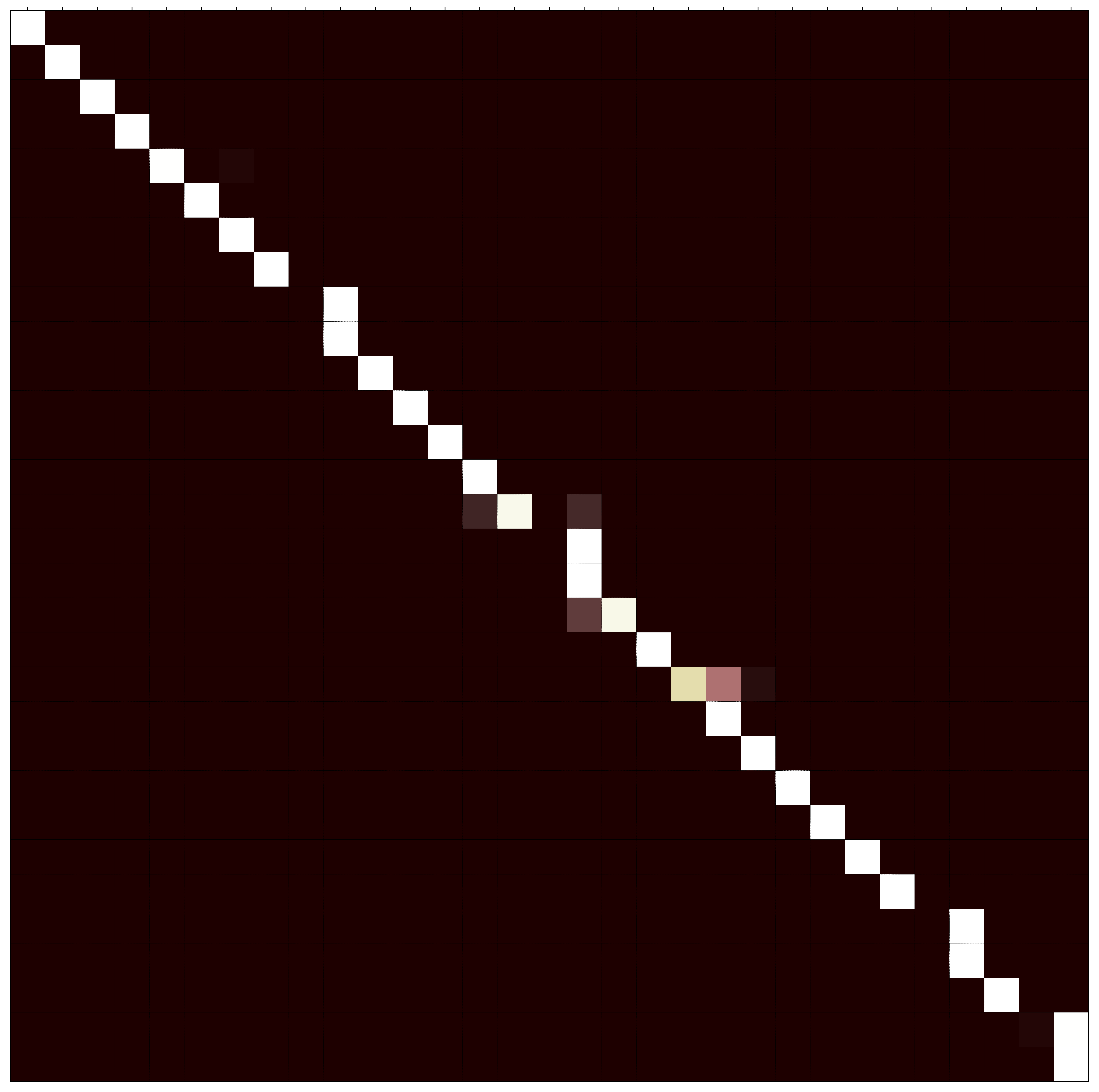}
\includegraphics[width=0.15\textwidth]{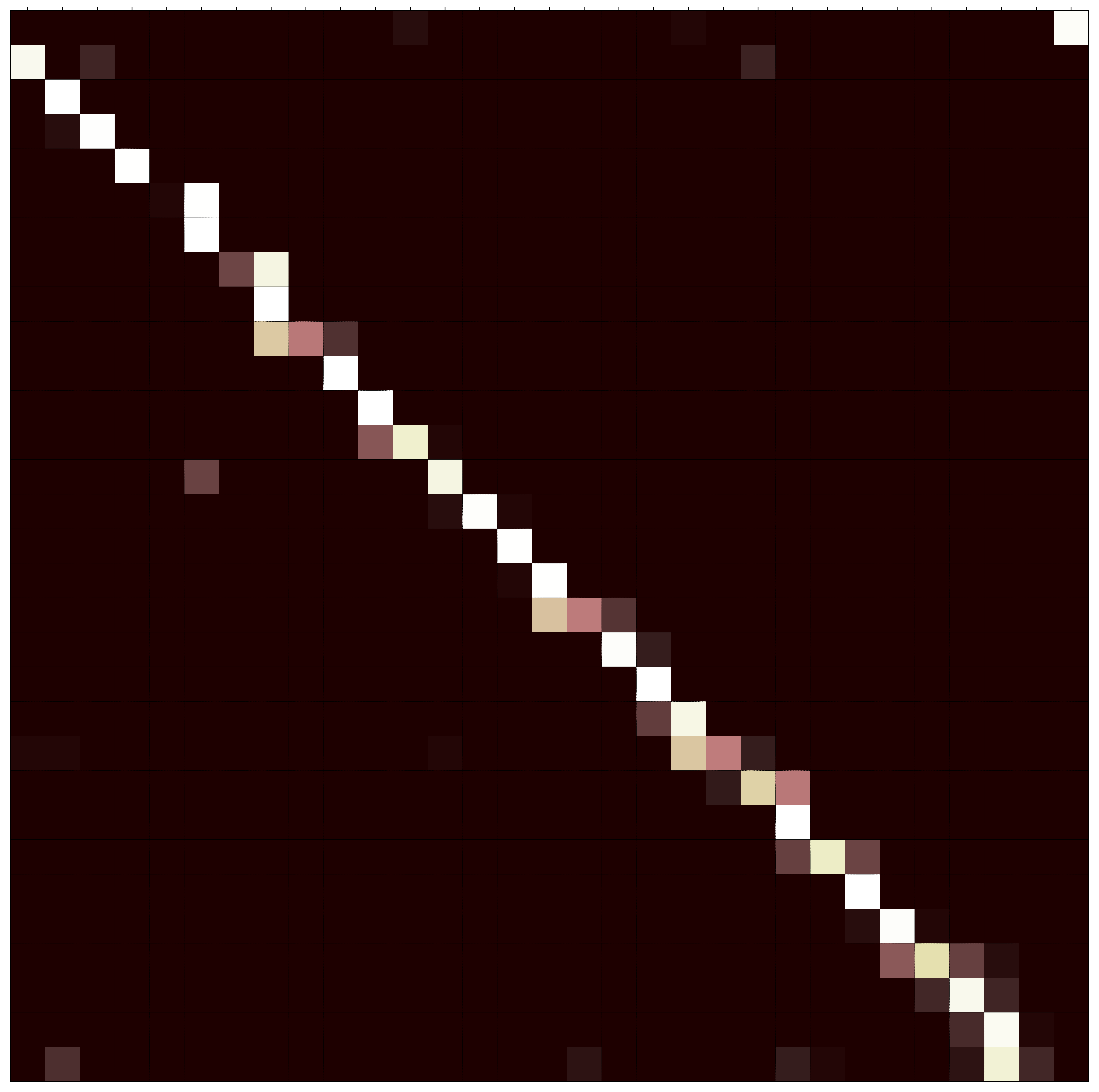}
\includegraphics[width=0.15\textwidth]{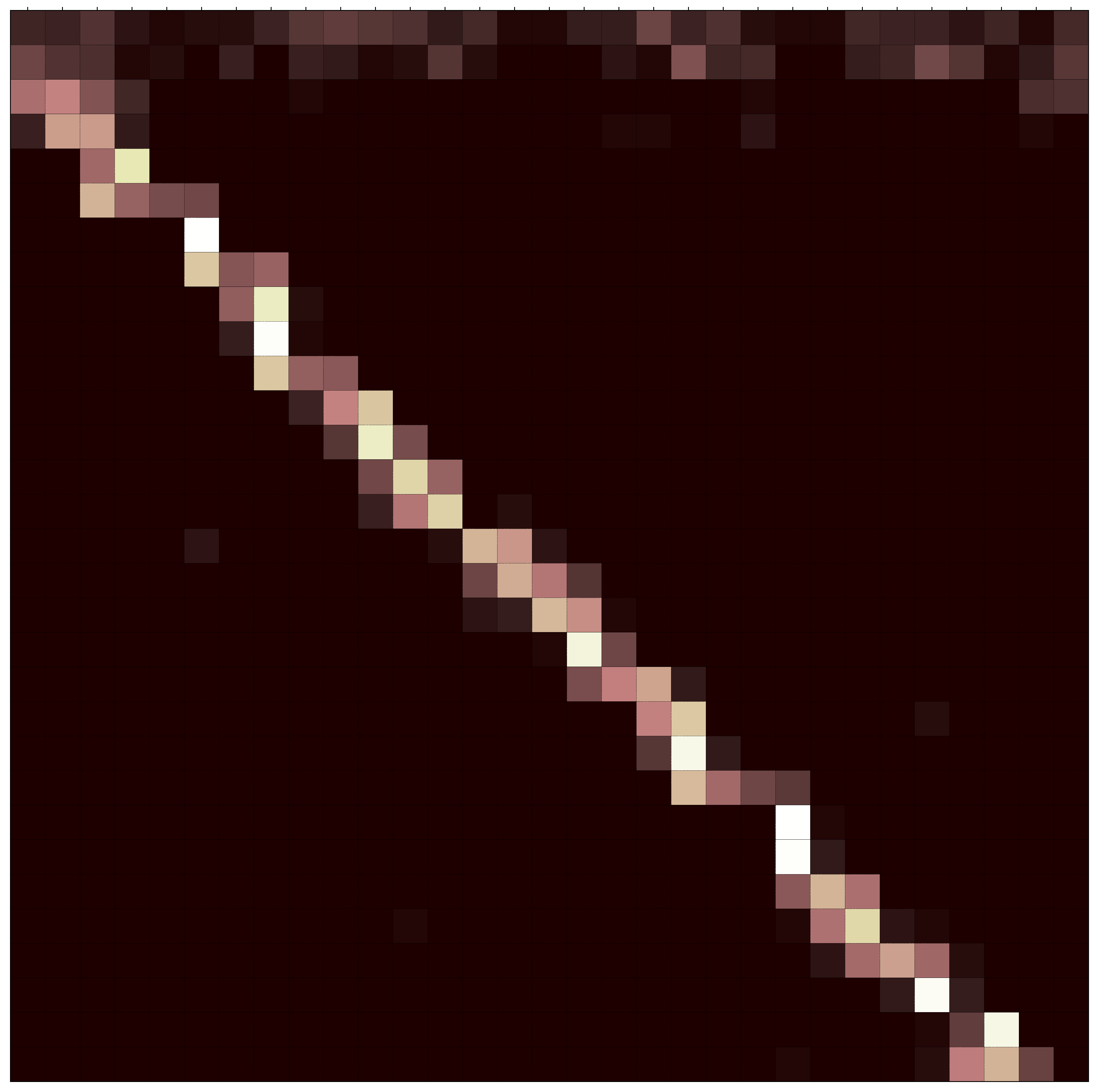}
\includegraphics[width=0.15\textwidth]{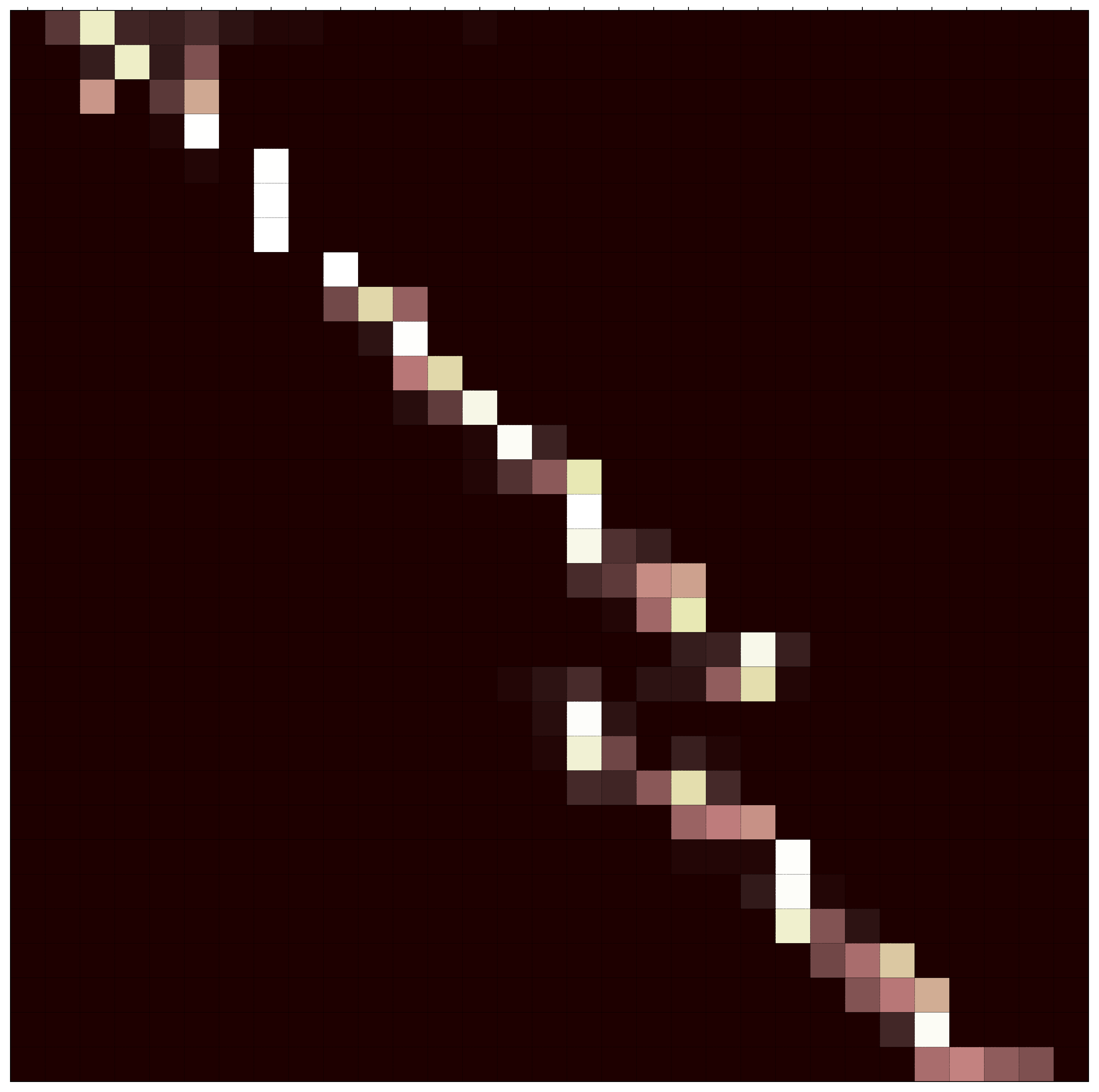}
\includegraphics[width=0.15\textwidth]{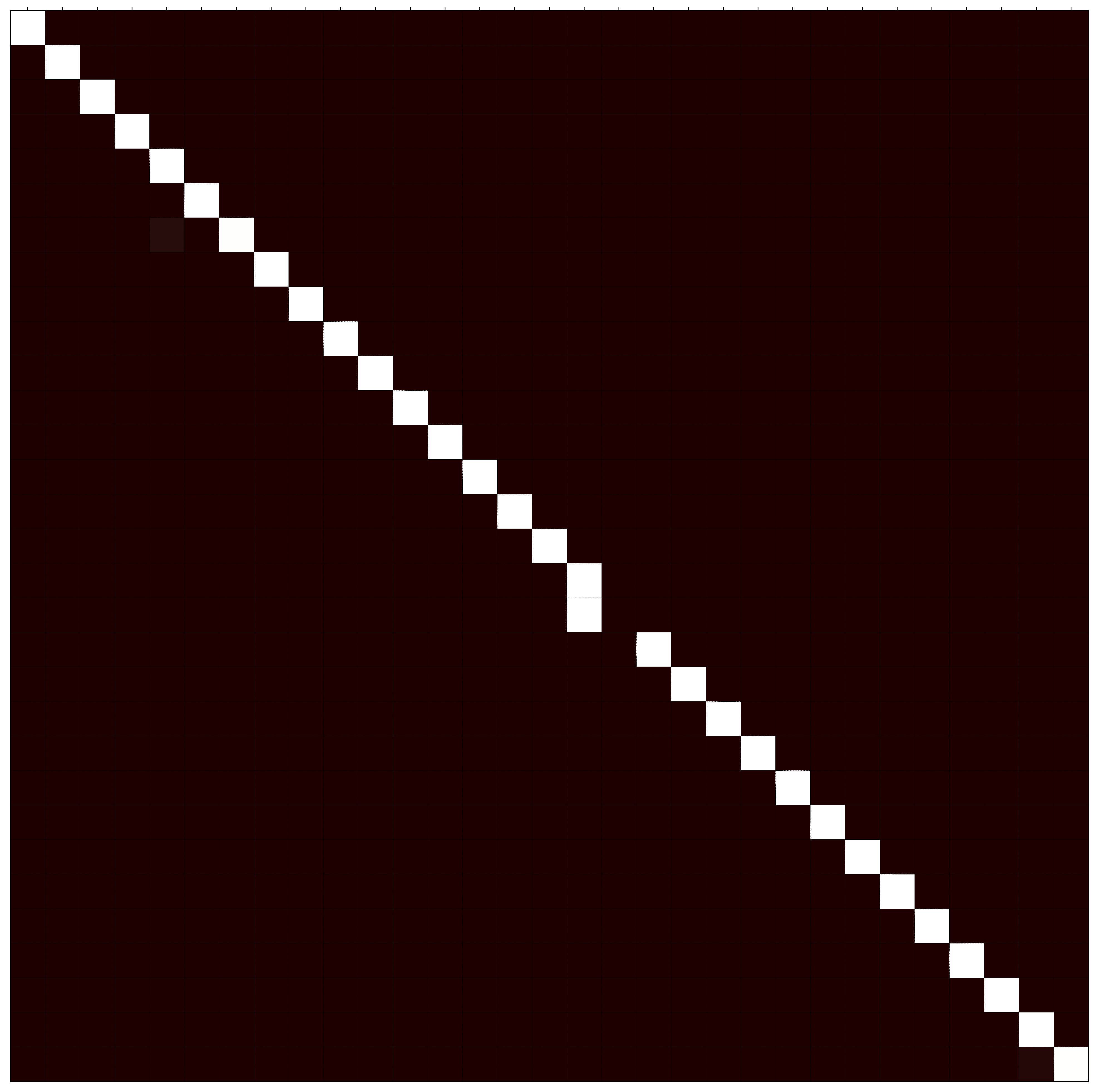}
\includegraphics[width=0.15\textwidth]{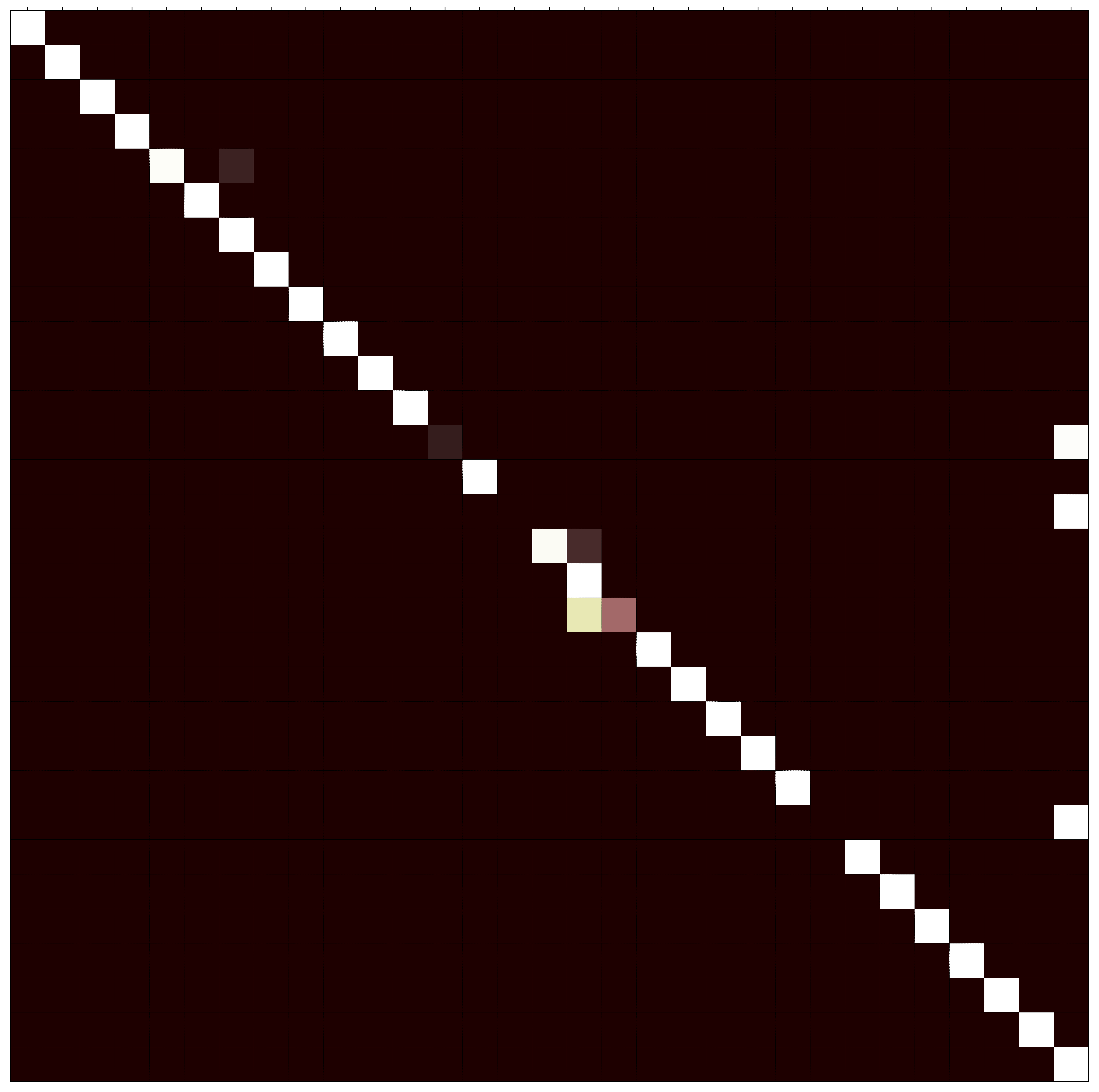}
\includegraphics[width=0.15\textwidth]{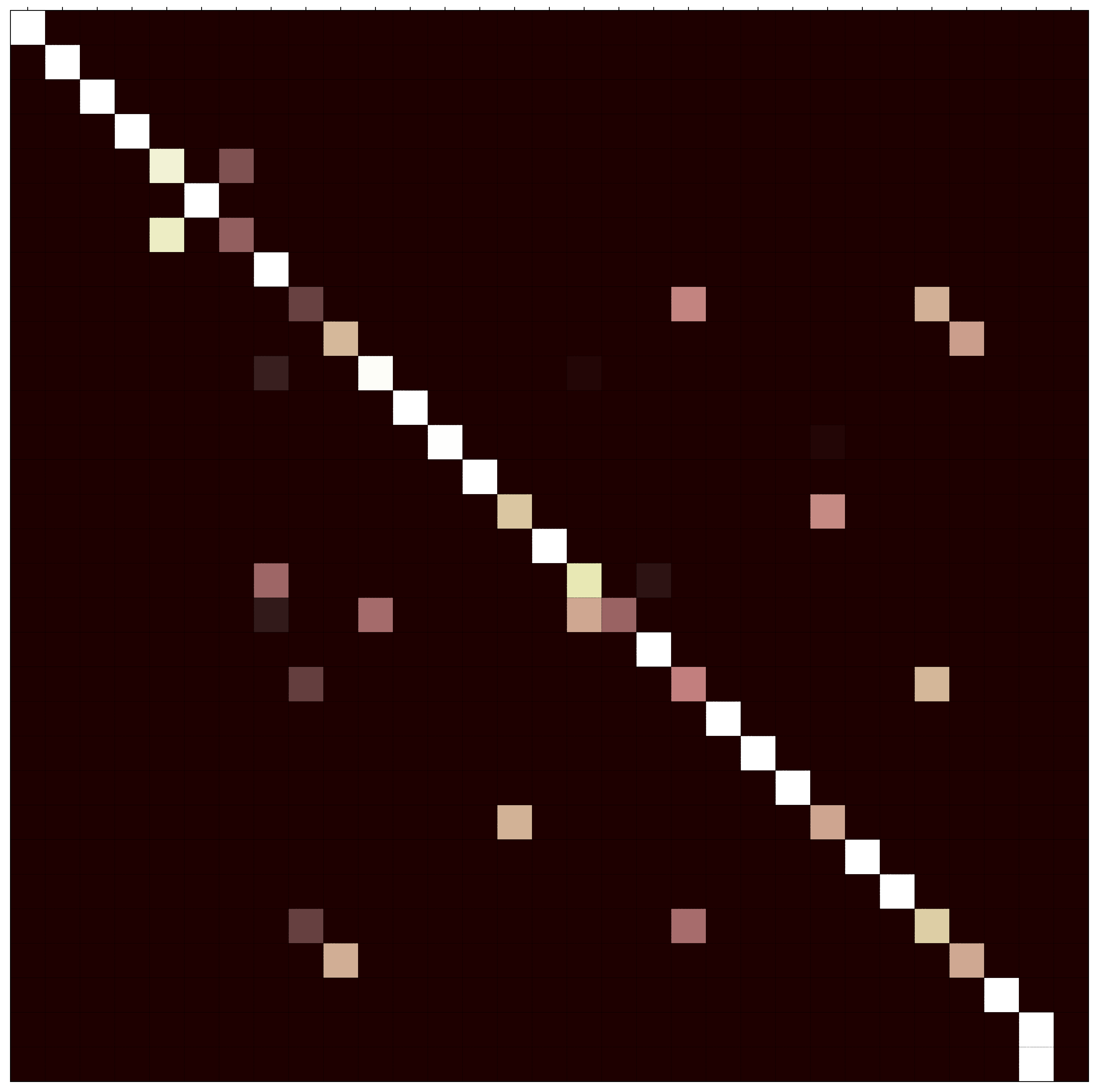}
\includegraphics[width=0.15\textwidth]{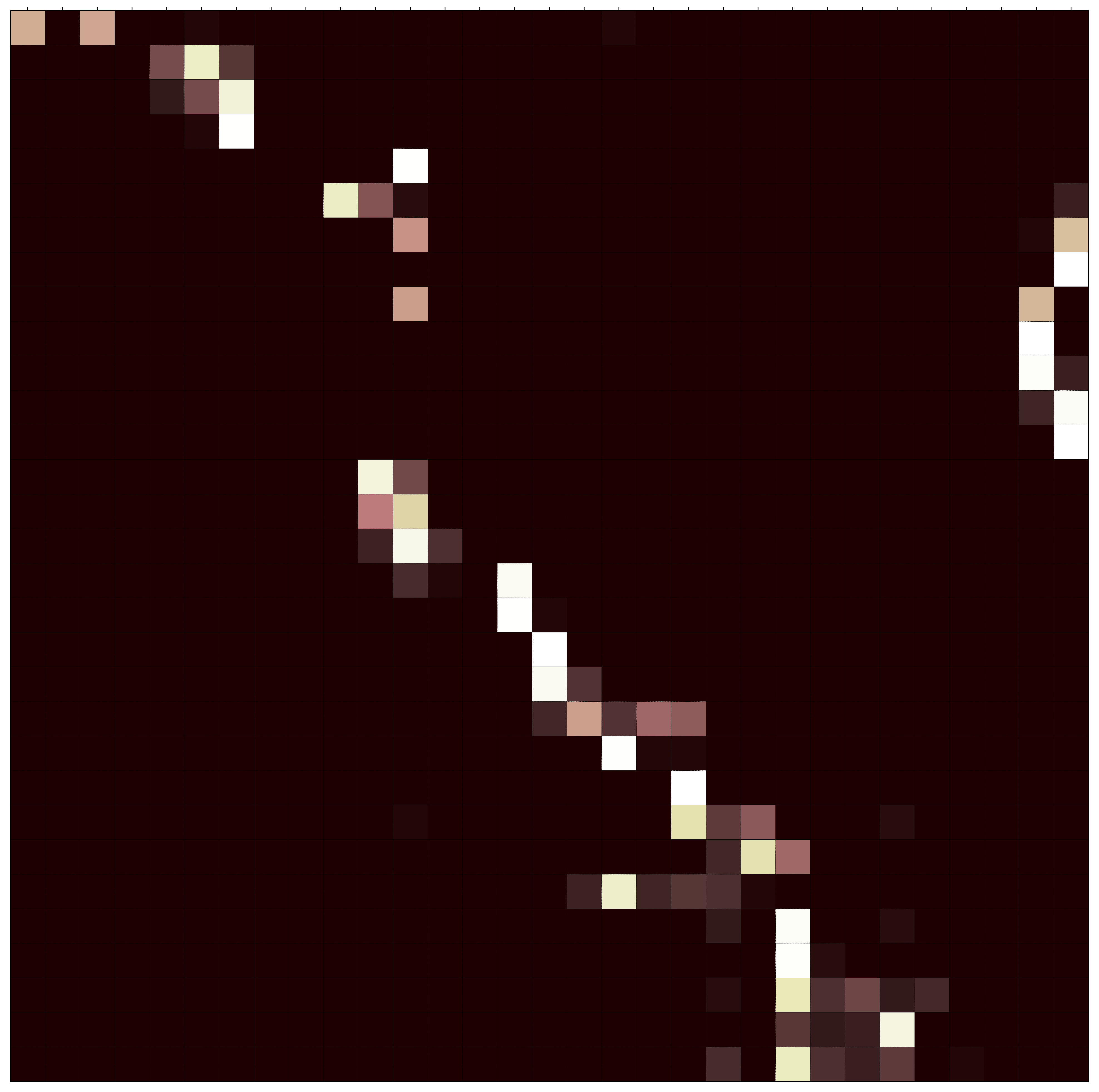}
\includegraphics[width=0.15\textwidth]{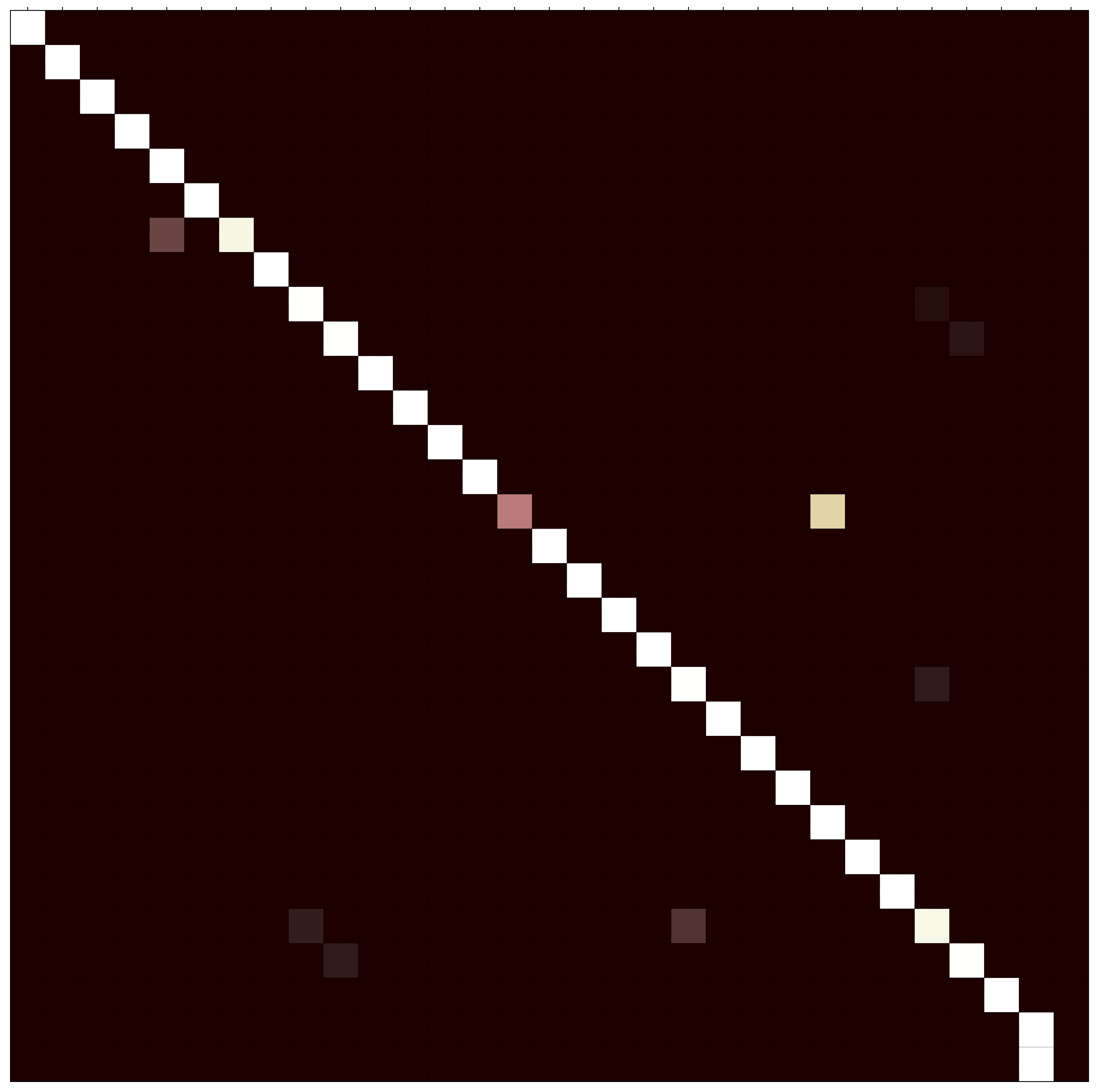}
\includegraphics[width=0.15\textwidth]{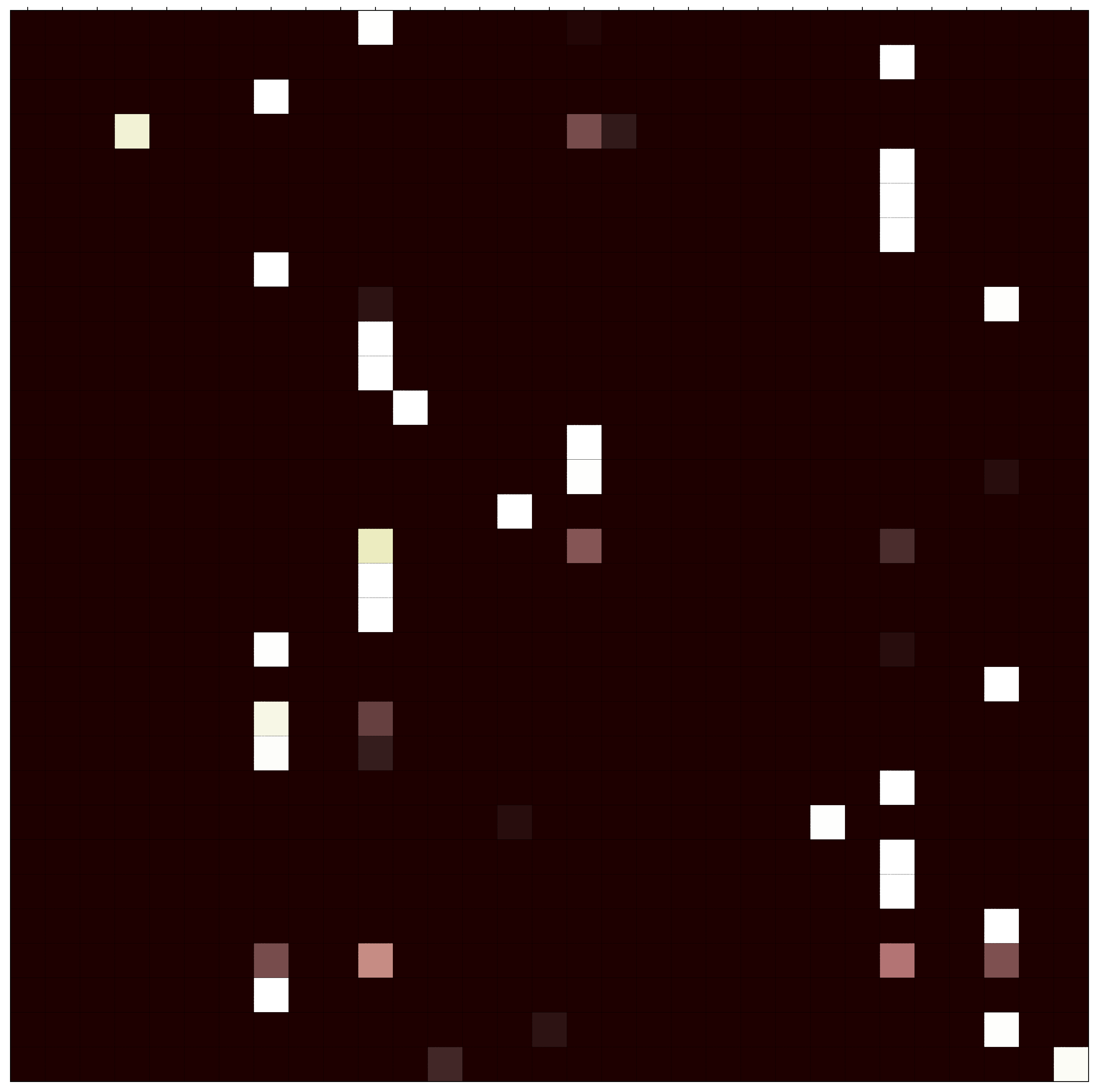}
\includegraphics[width=0.15\textwidth]{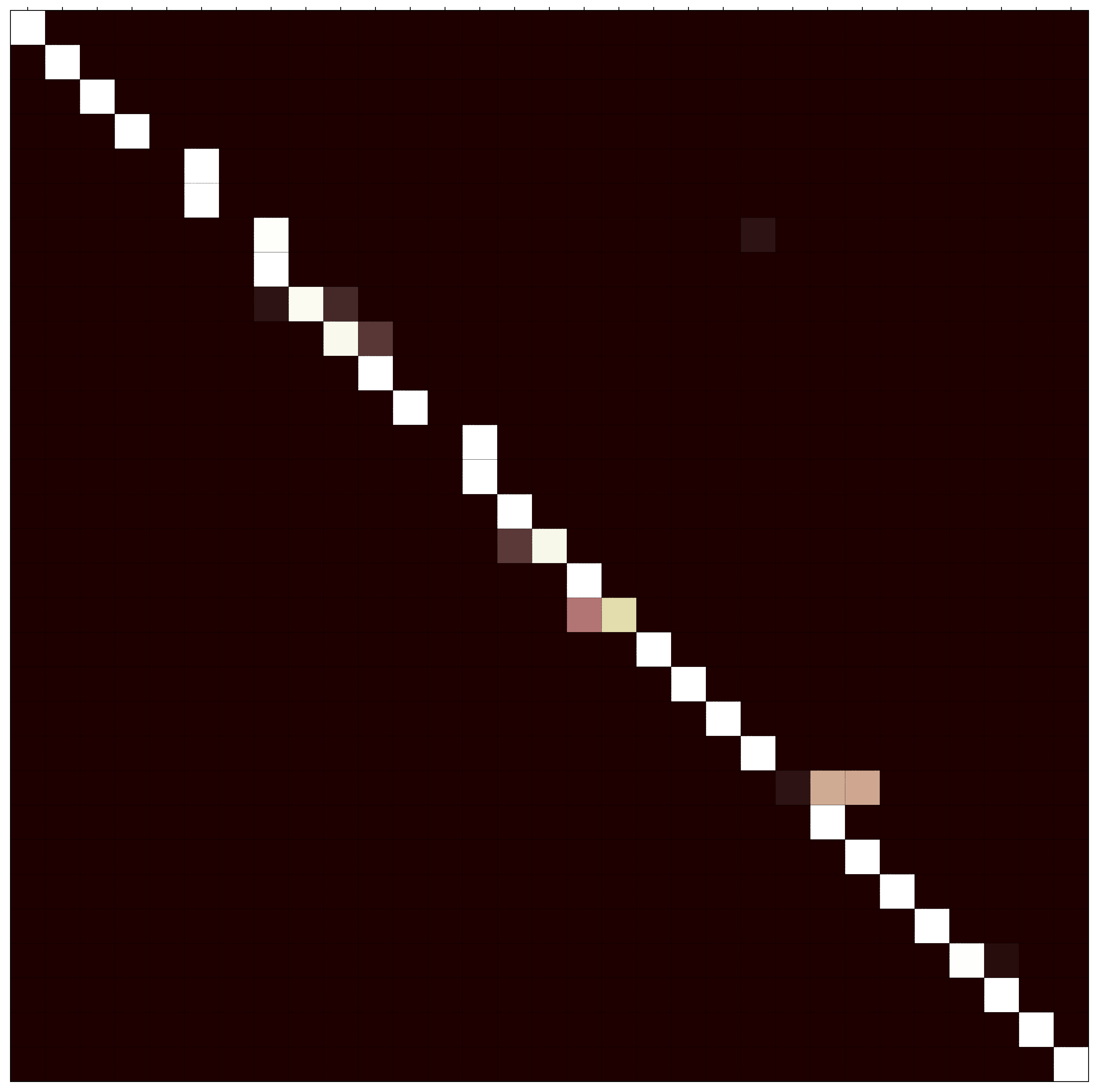}
\includegraphics[width=0.15\textwidth]{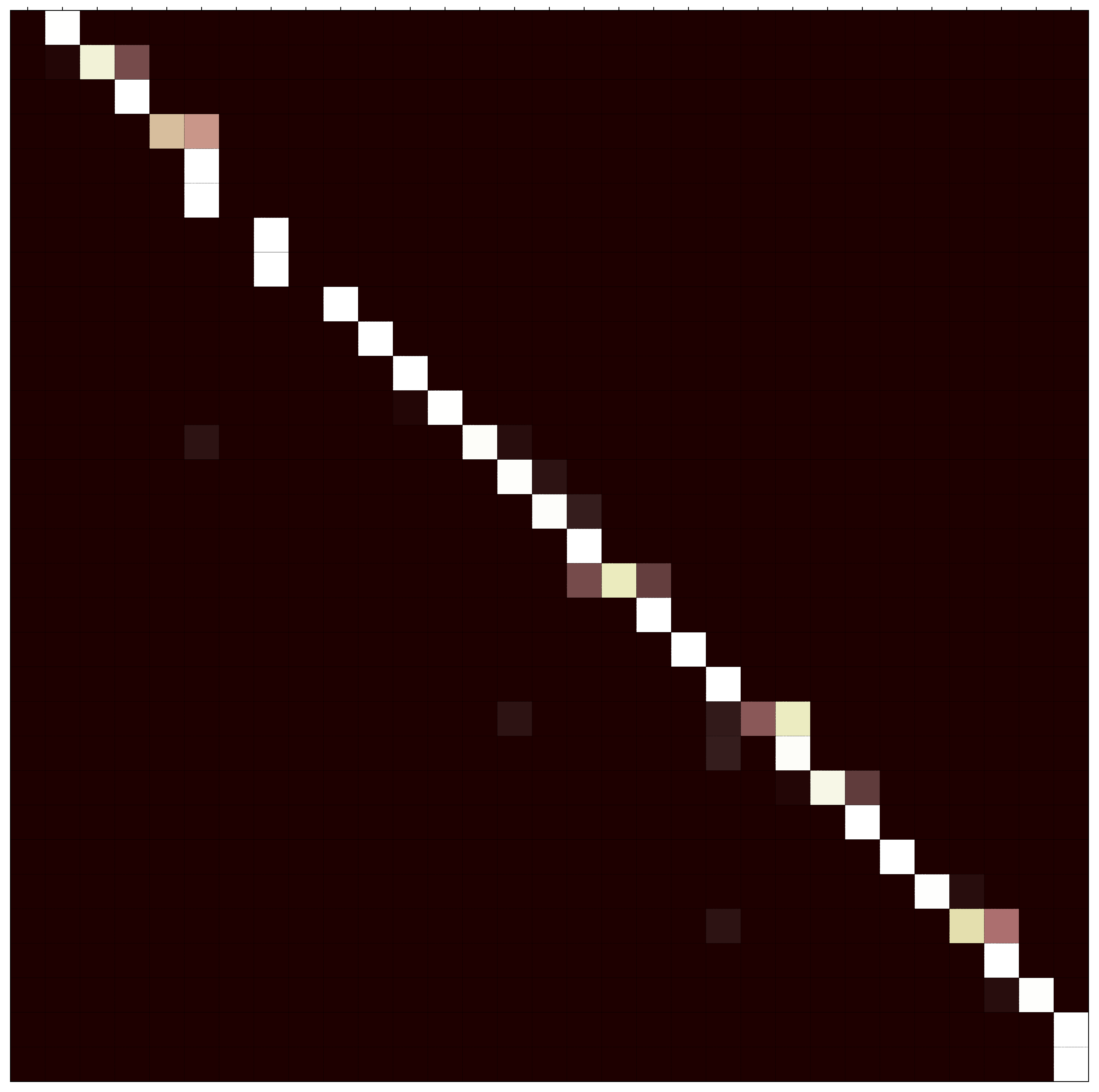}
\includegraphics[width=0.15\textwidth]{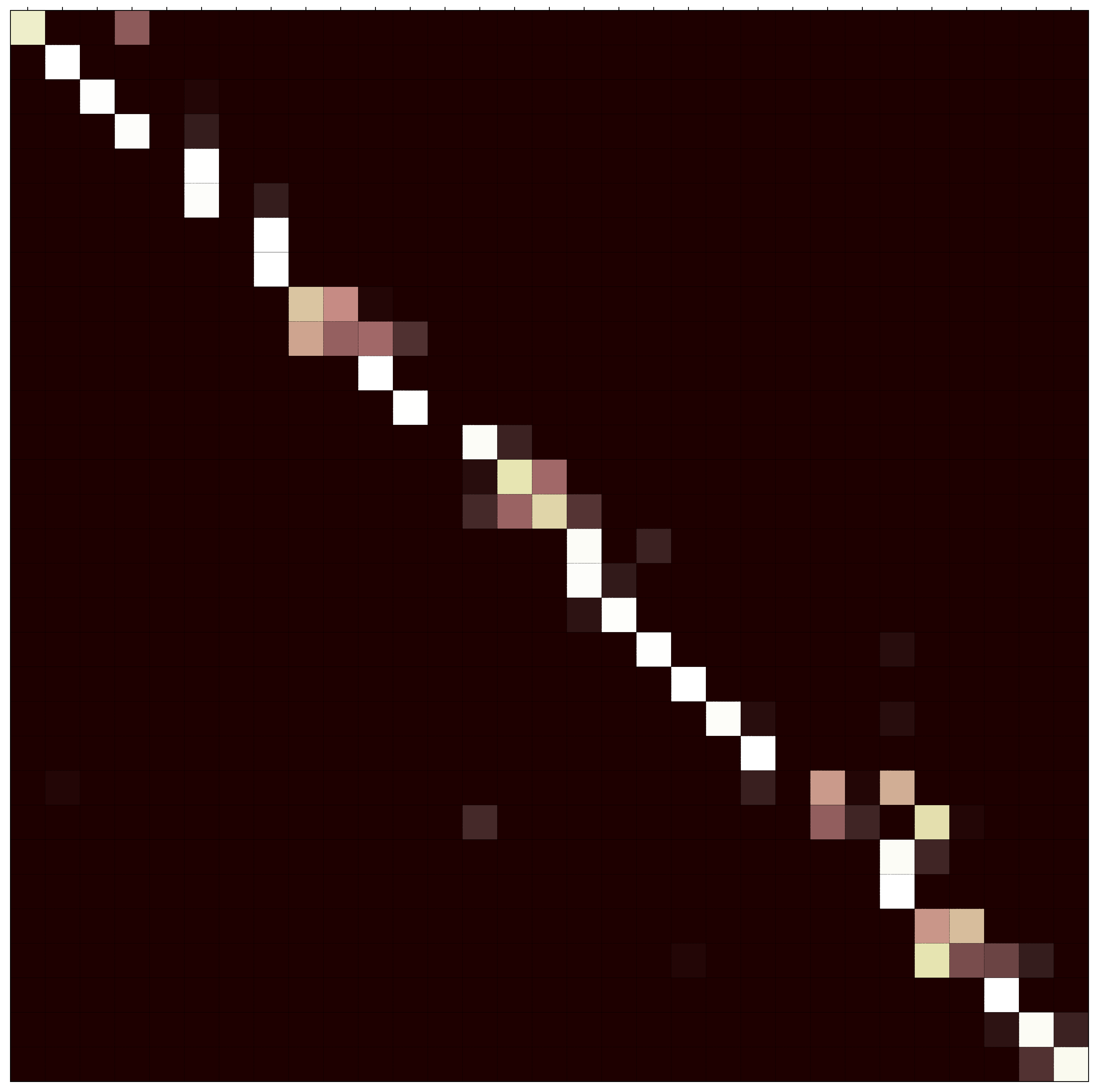}
\end{center}
\caption{Layer 1}
\end{figure}

\begin{figure}[b!]
\vspace{93mm}
\begin{center}
\includegraphics[width=0.15\textwidth]{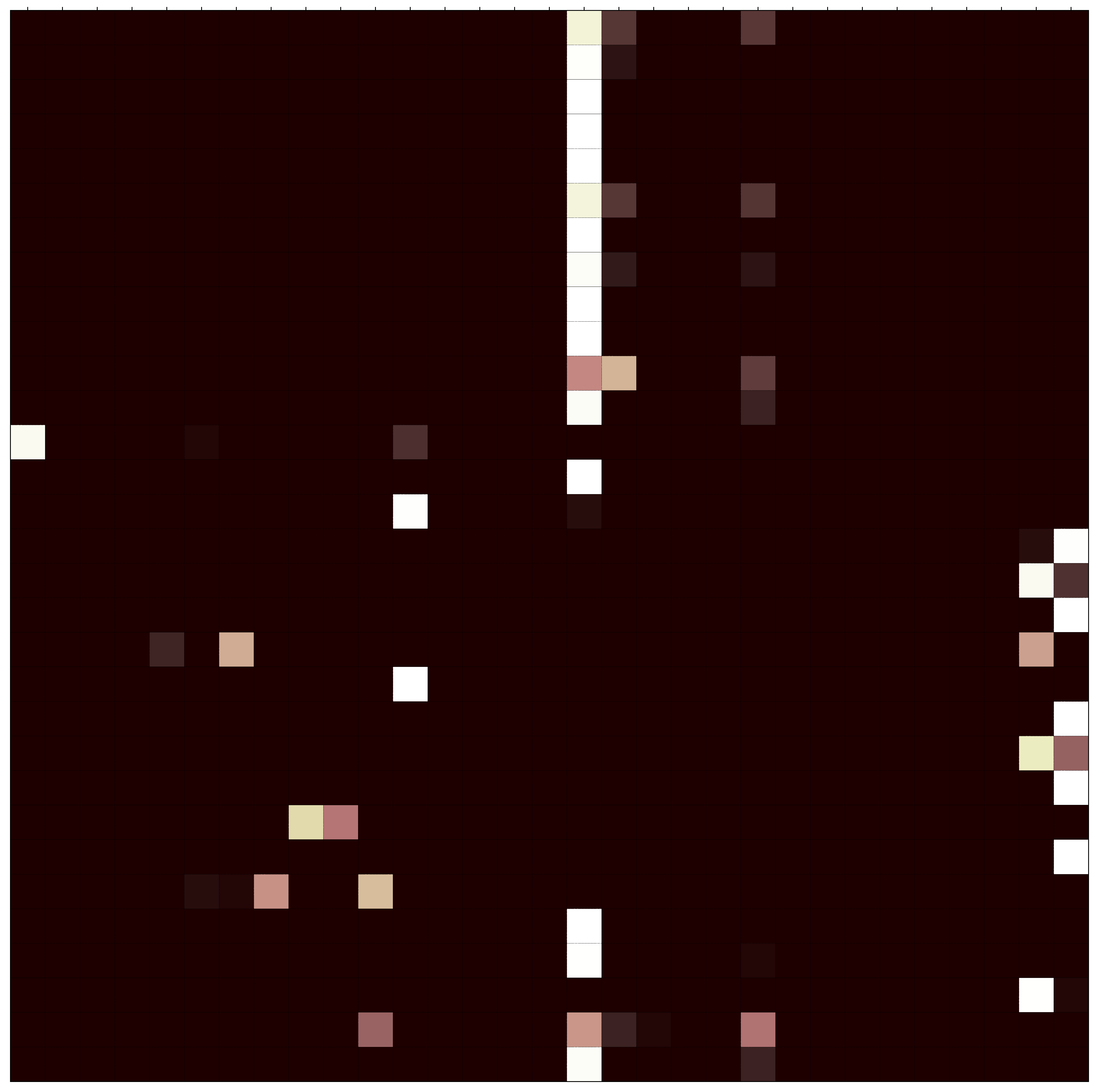}
\includegraphics[width=0.15\textwidth]{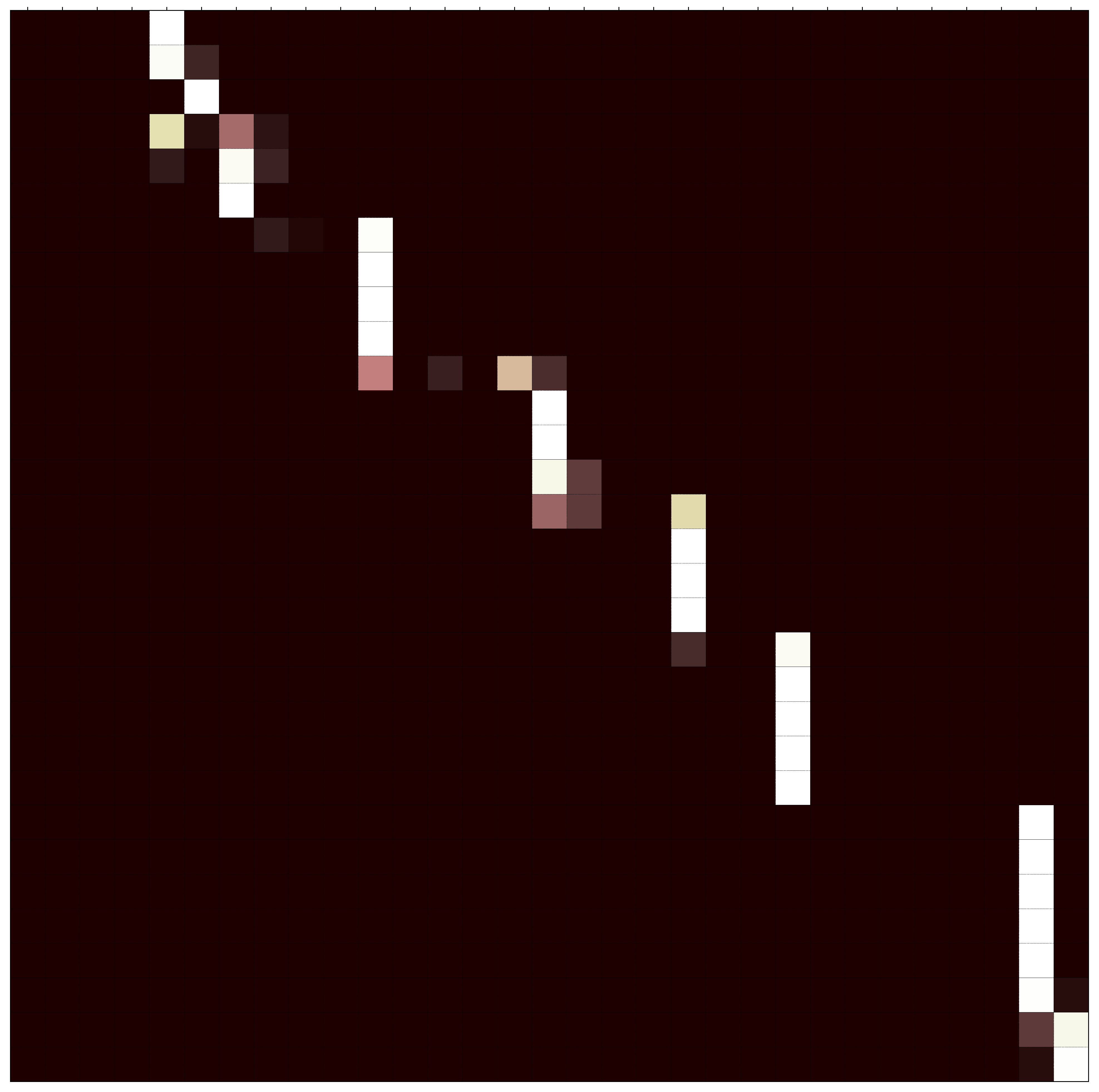}
\includegraphics[width=0.15\textwidth]{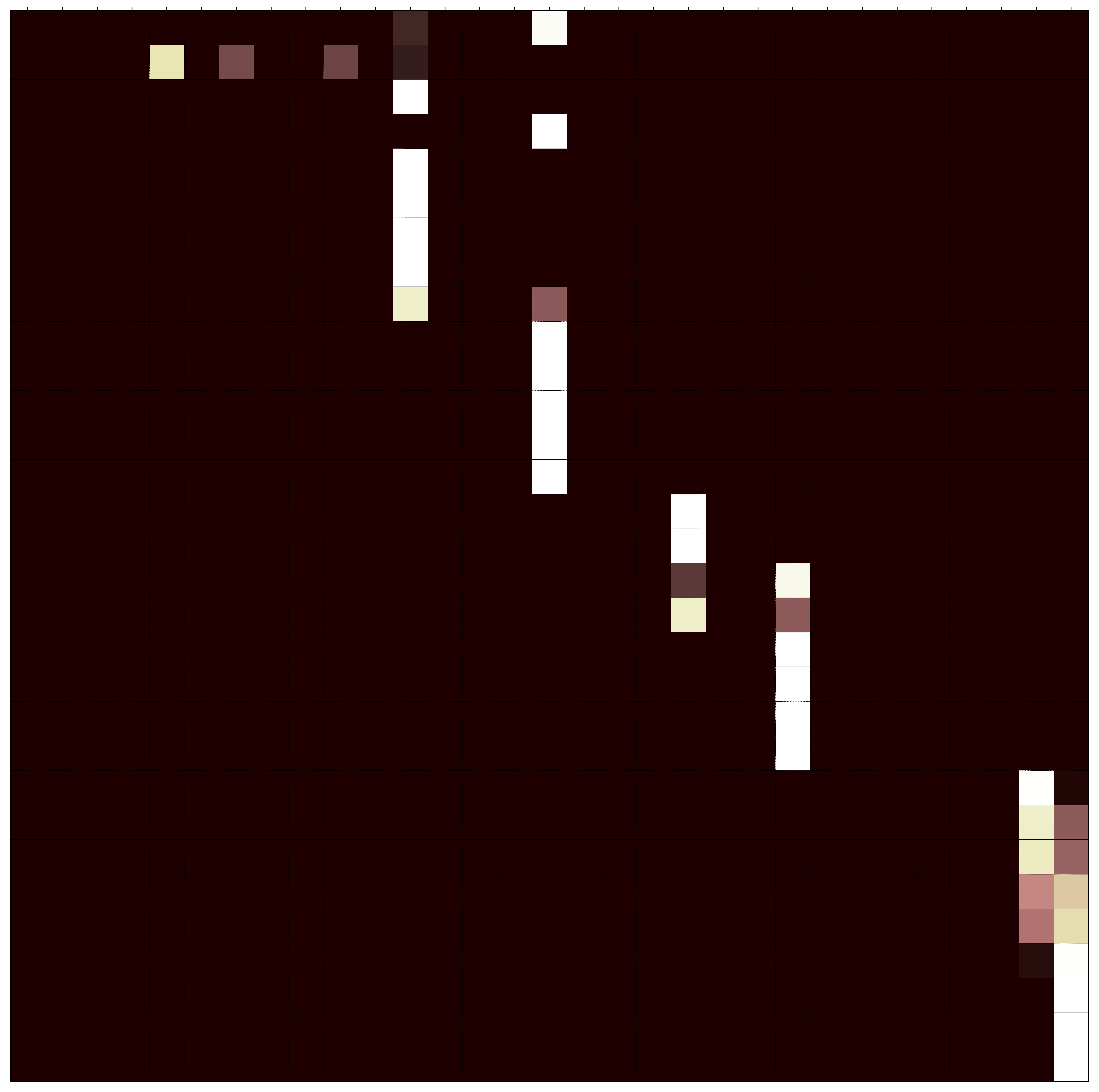}
\includegraphics[width=0.15\textwidth]{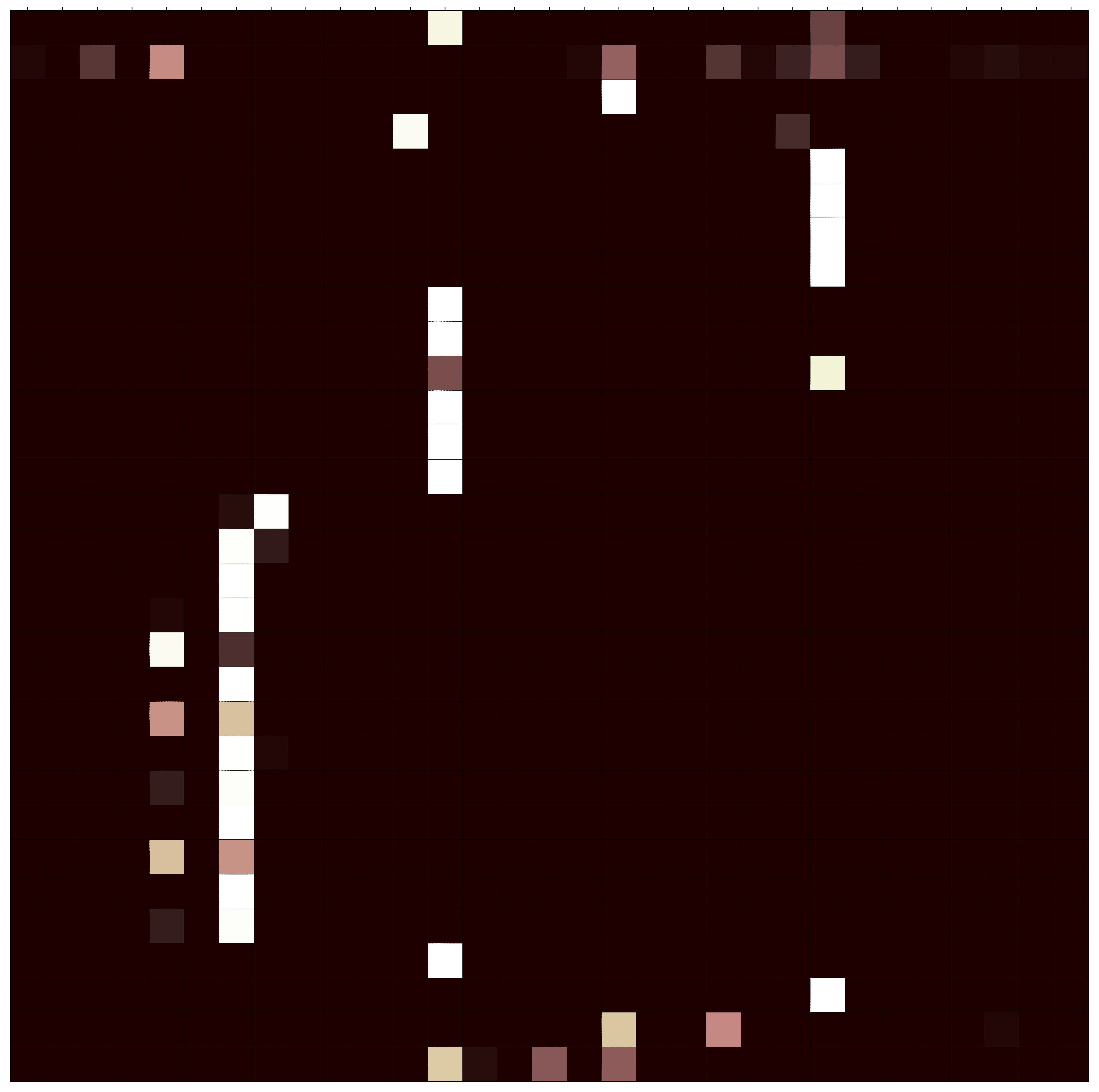}
\includegraphics[width=0.15\textwidth]{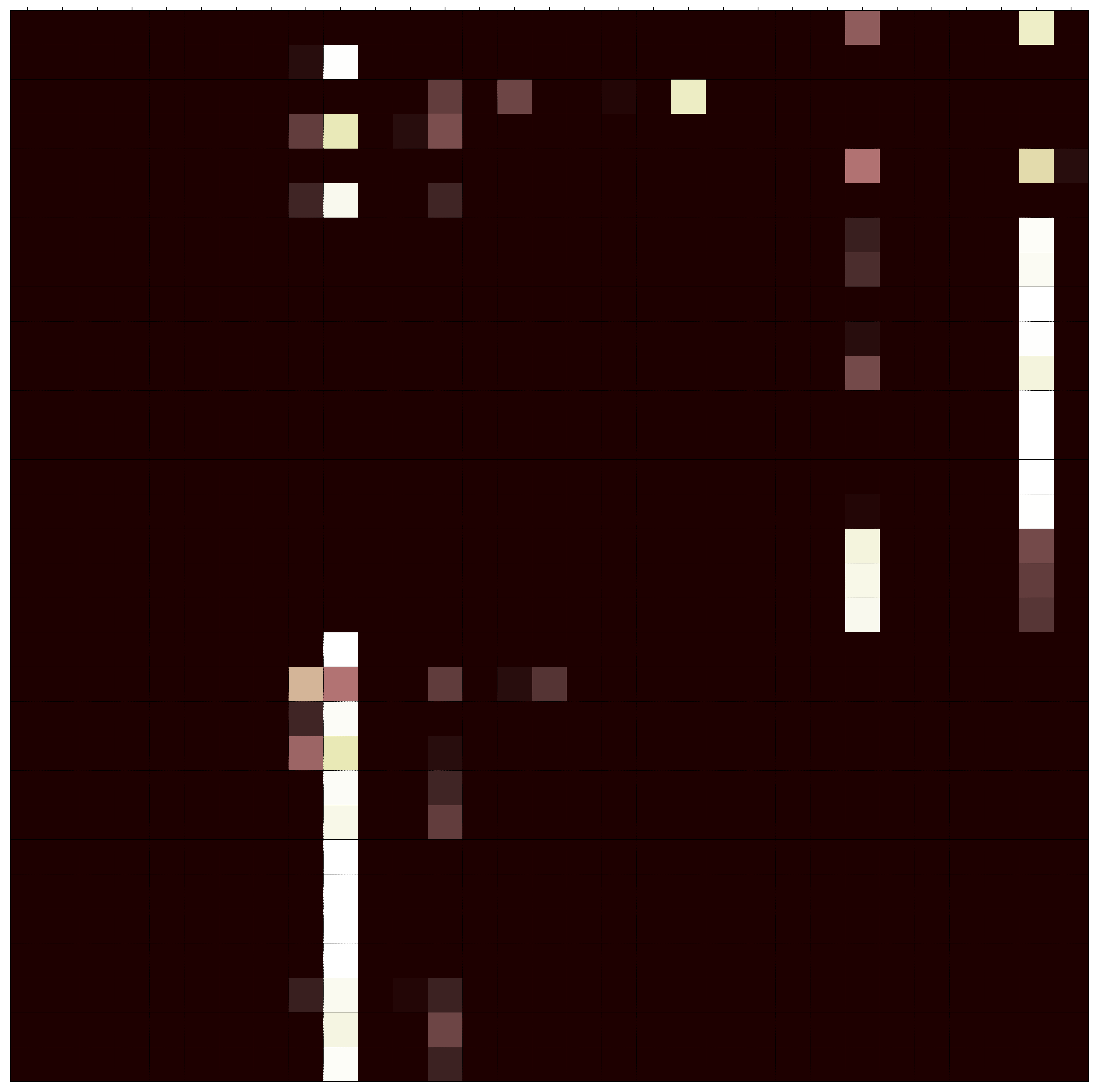}
\includegraphics[width=0.15\textwidth]{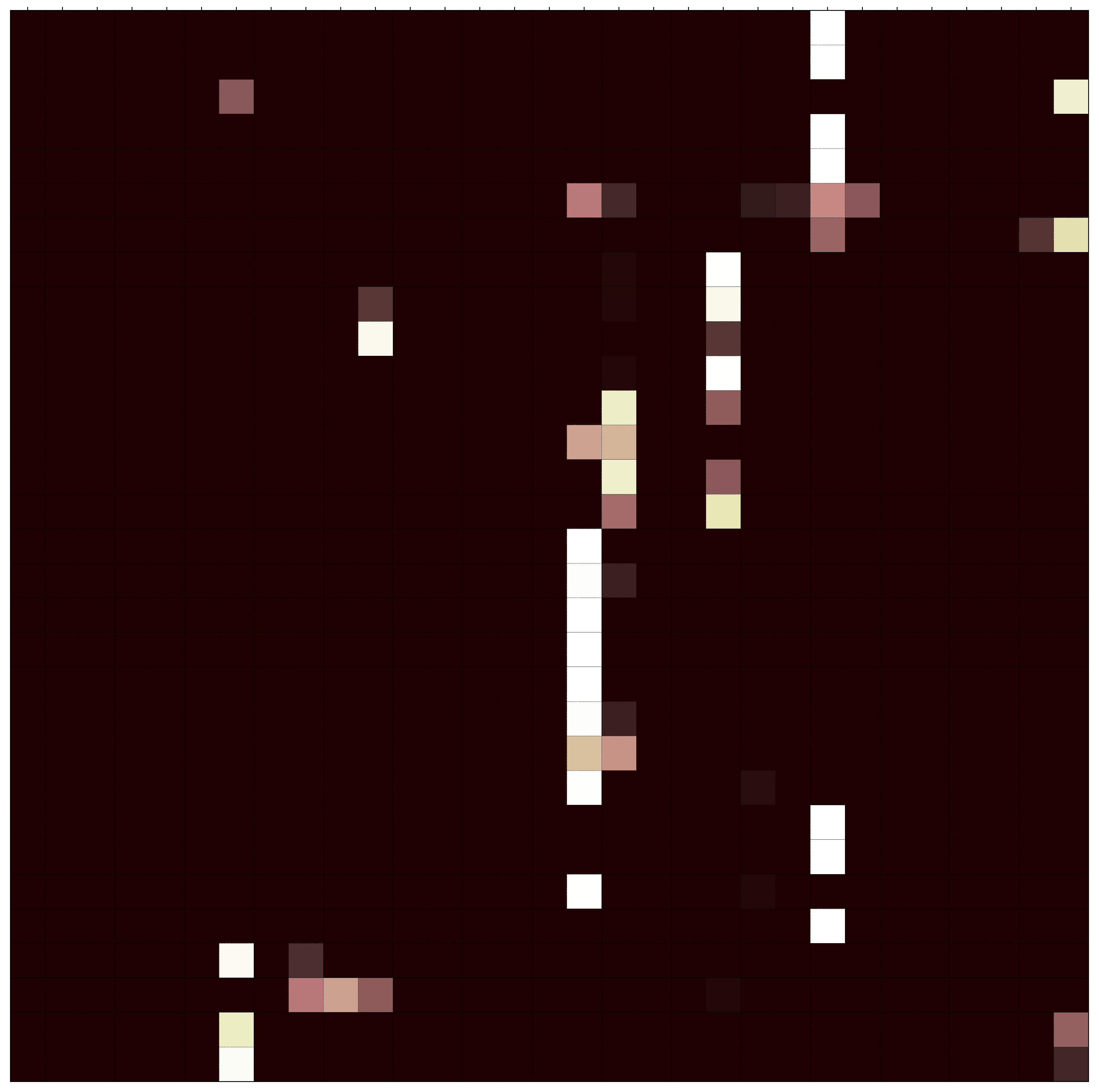}
\includegraphics[width=0.15\textwidth]{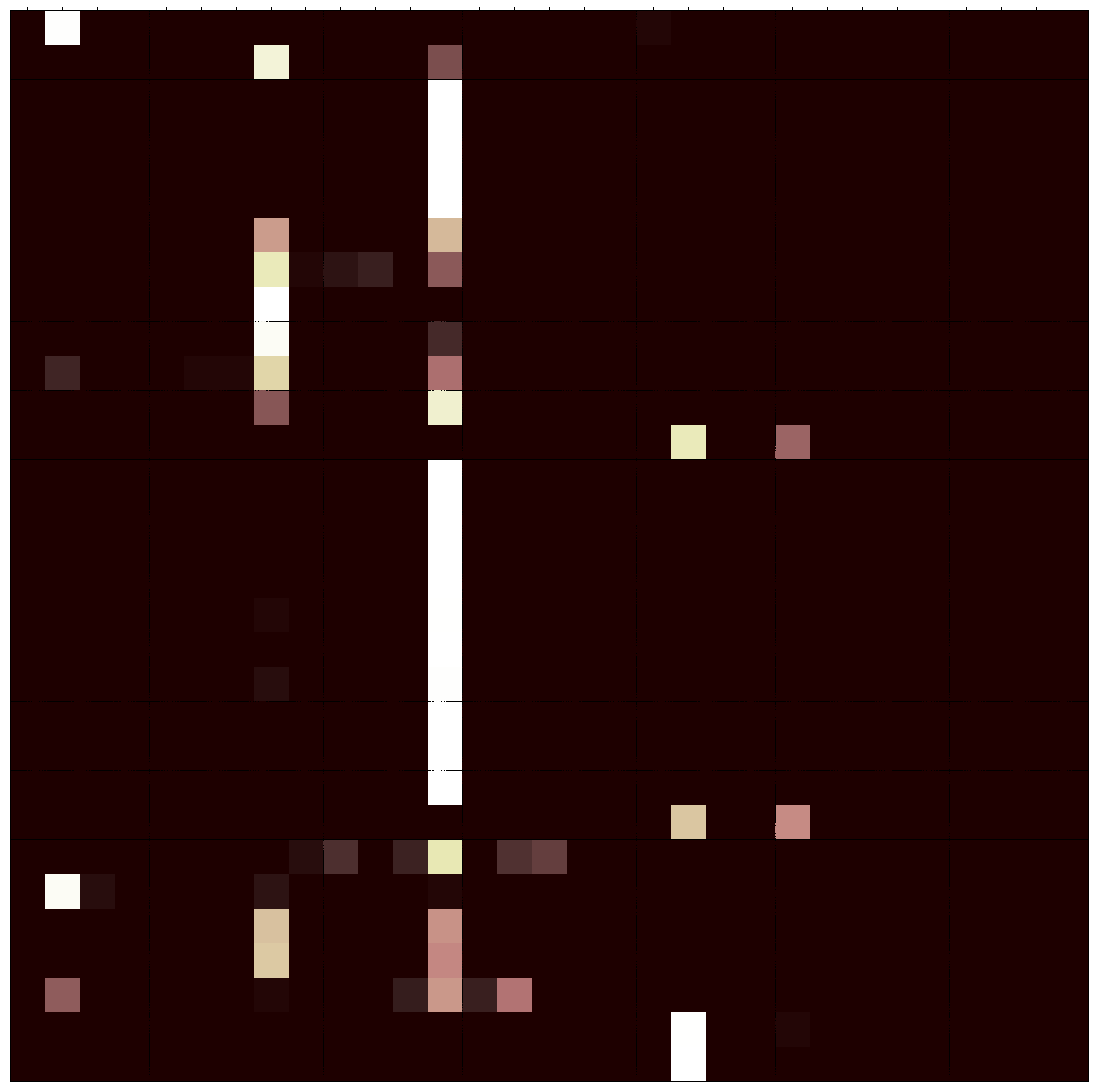}
\includegraphics[width=0.15\textwidth]{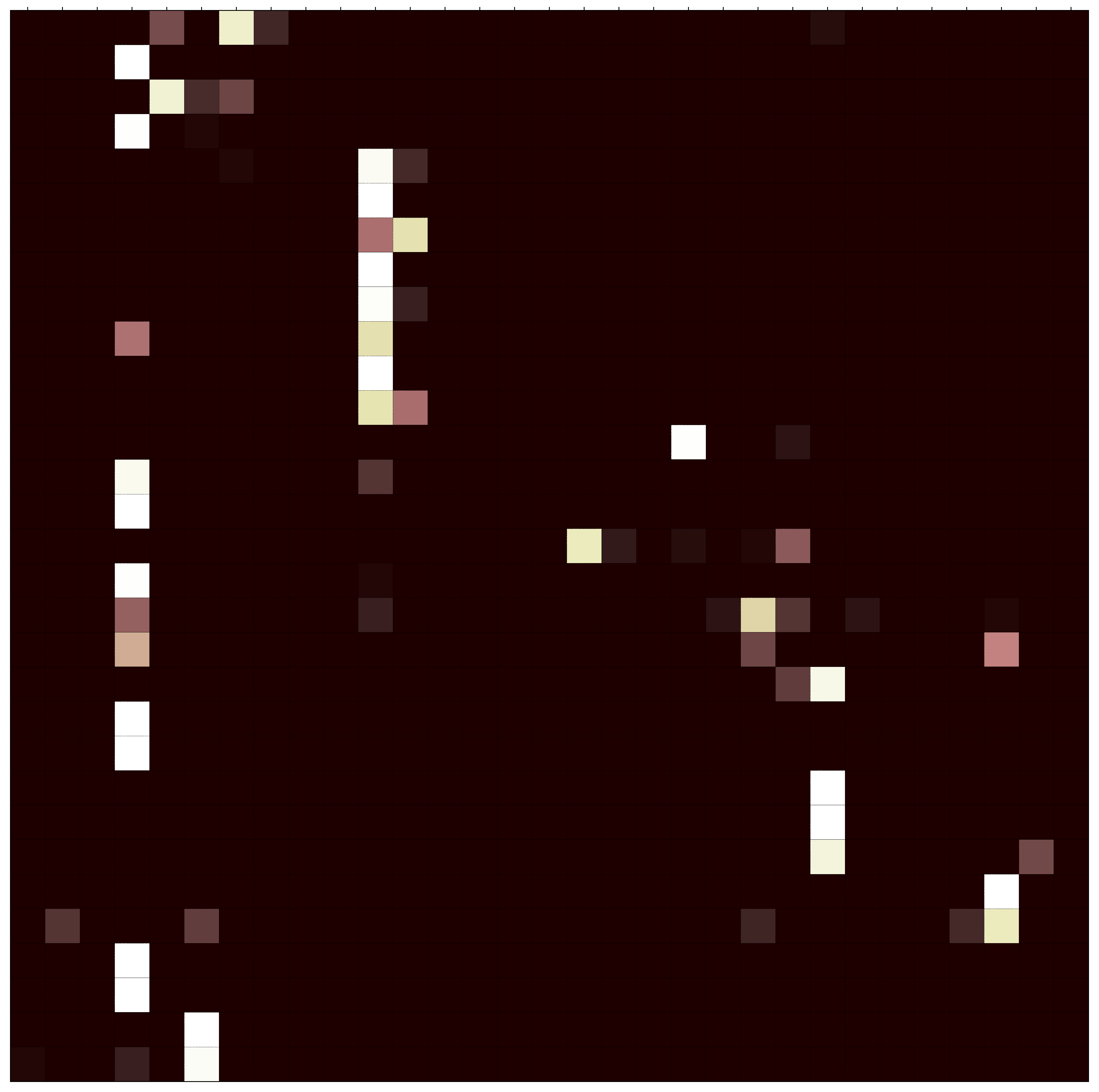}
\includegraphics[width=0.15\textwidth]{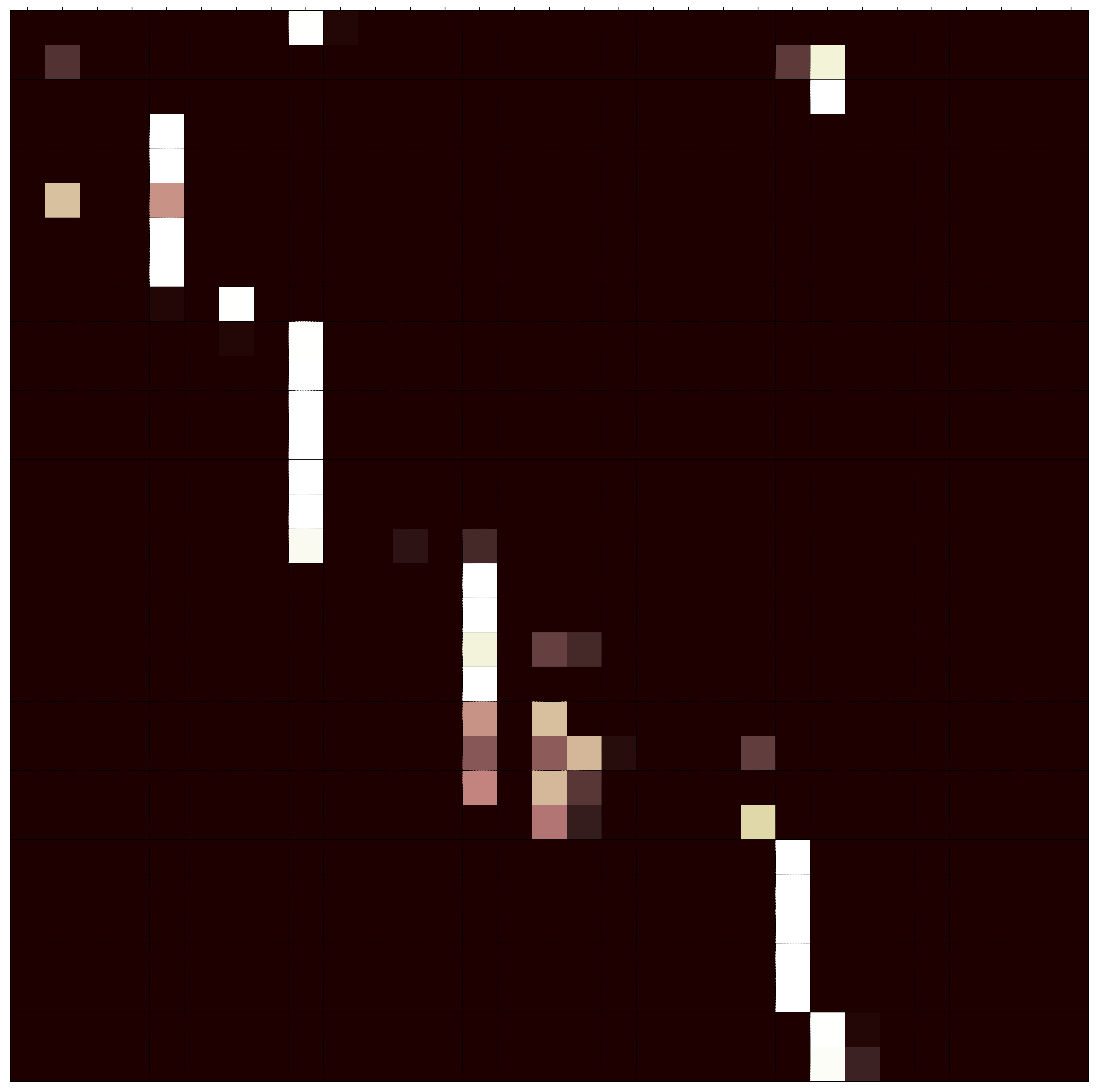}
\includegraphics[width=0.15\textwidth]{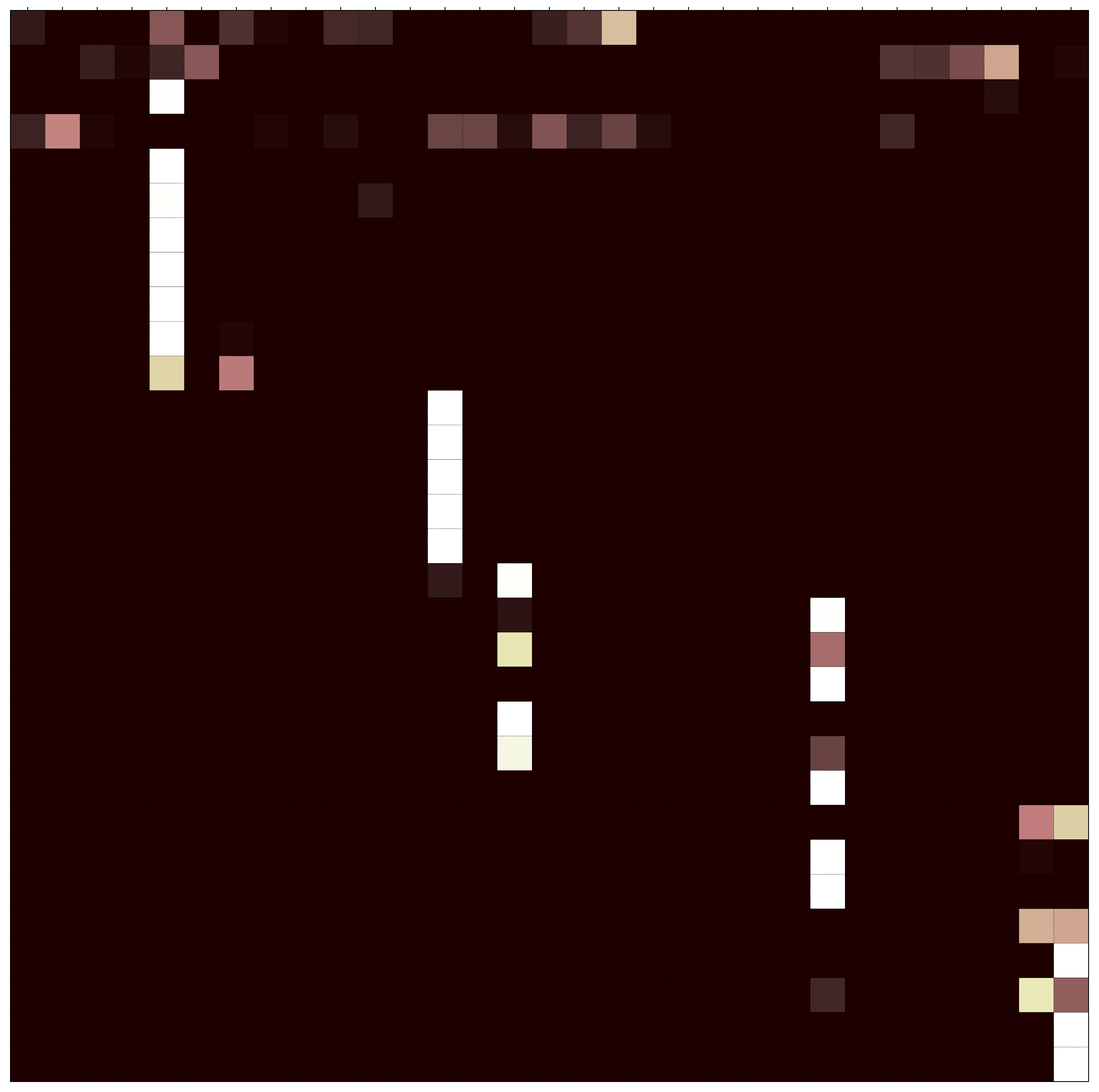}
\includegraphics[width=0.15\textwidth]{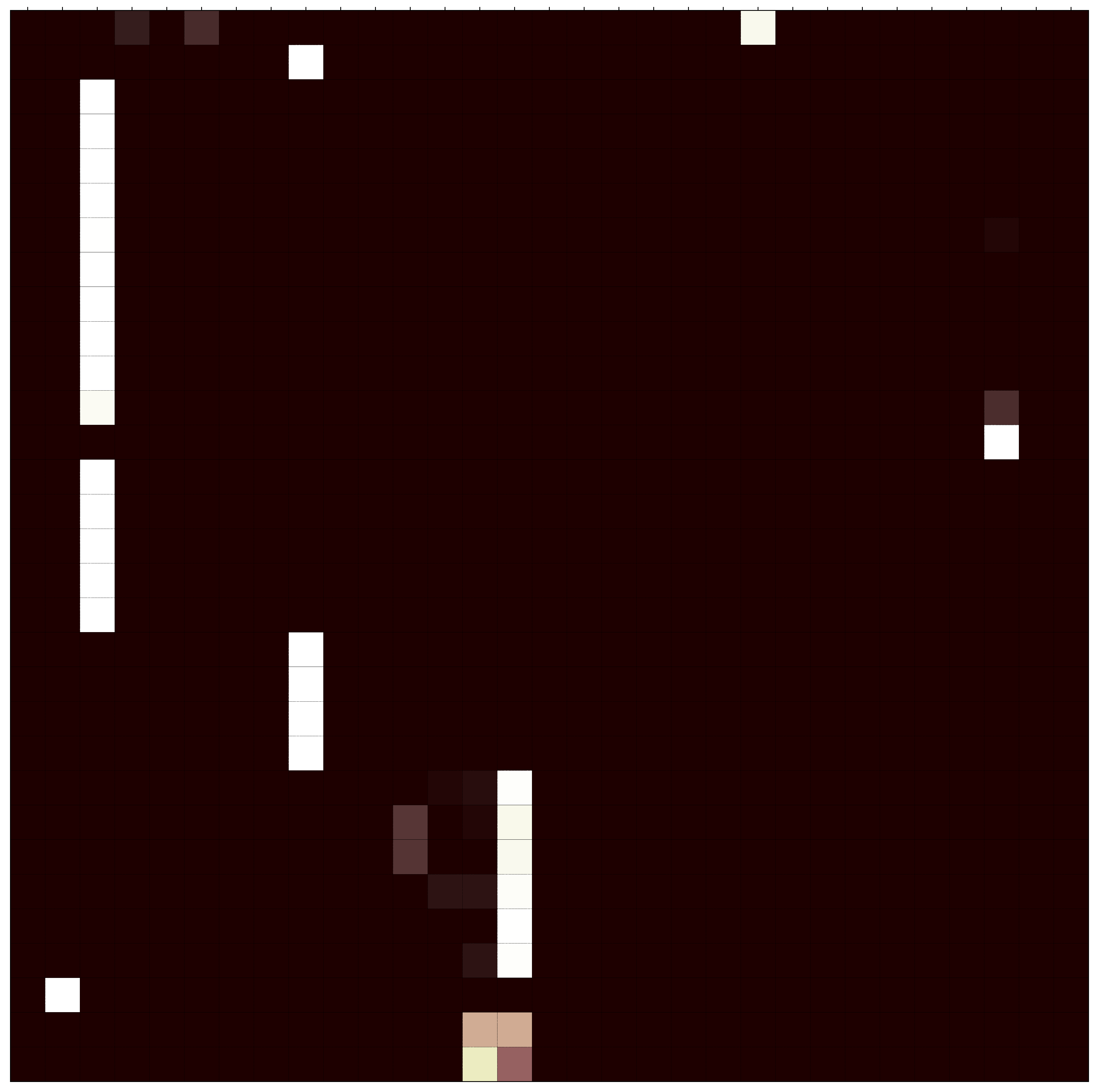}
\includegraphics[width=0.15\textwidth]{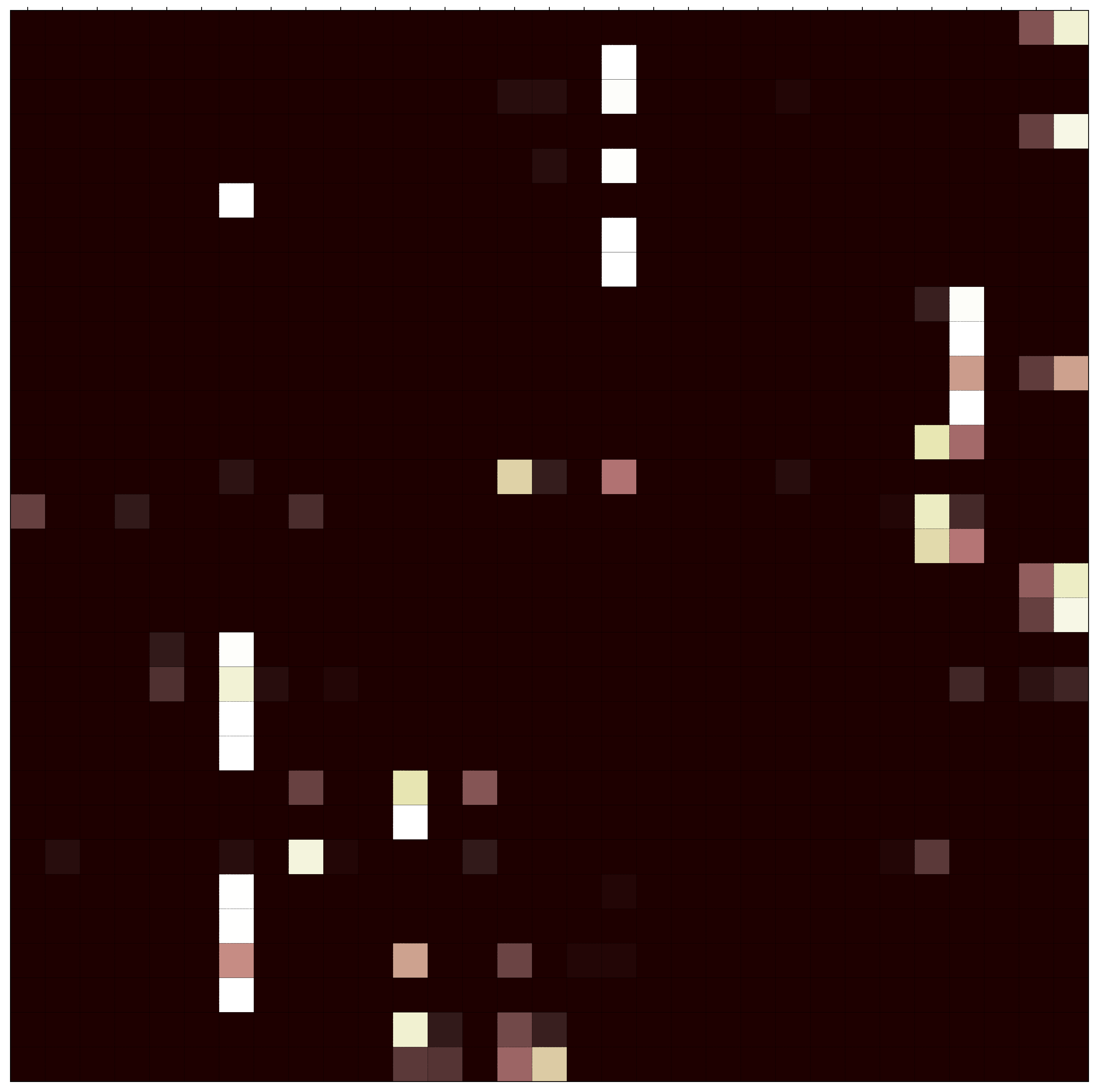}
\includegraphics[width=0.15\textwidth]{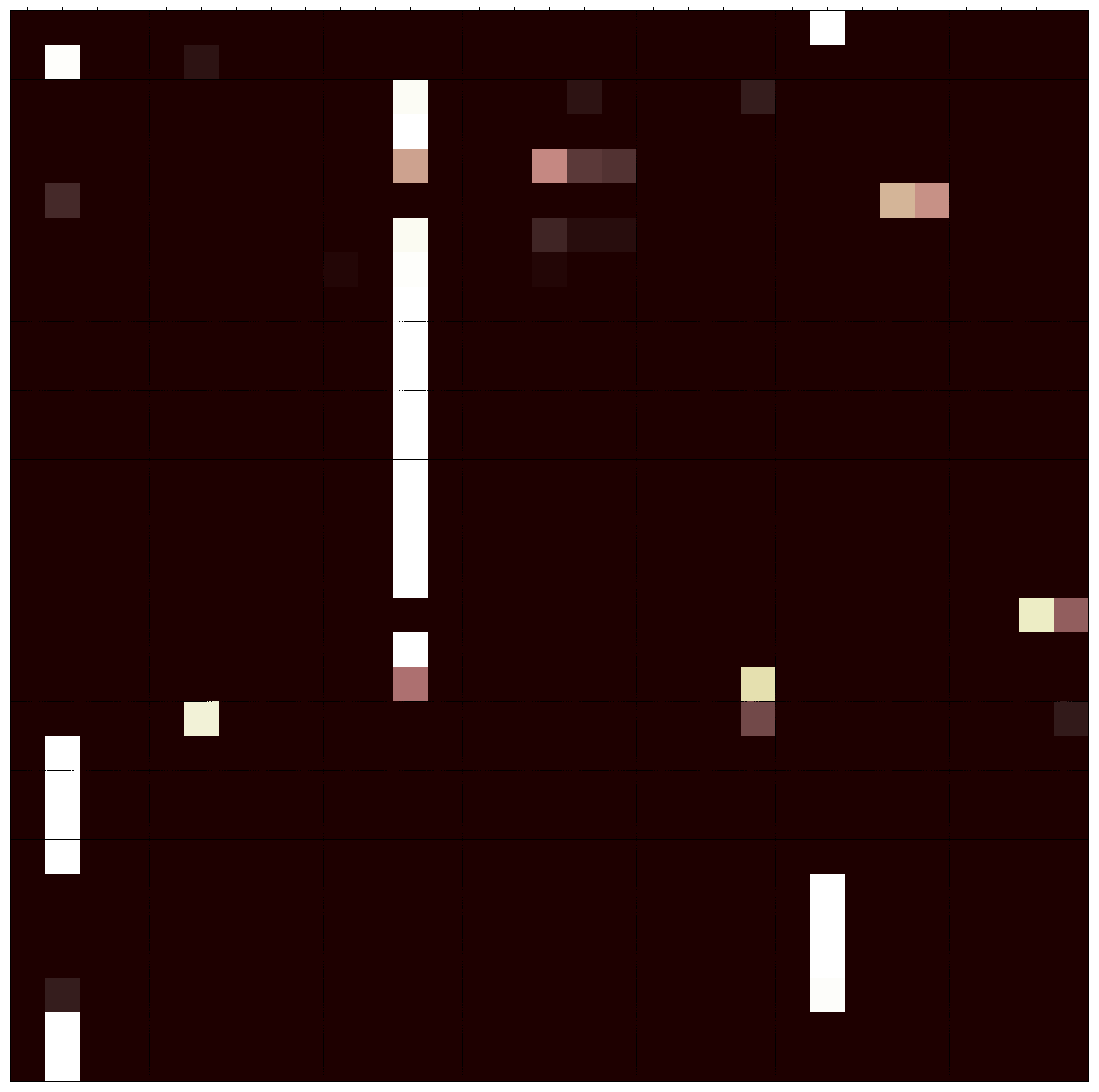}
\includegraphics[width=0.15\textwidth]{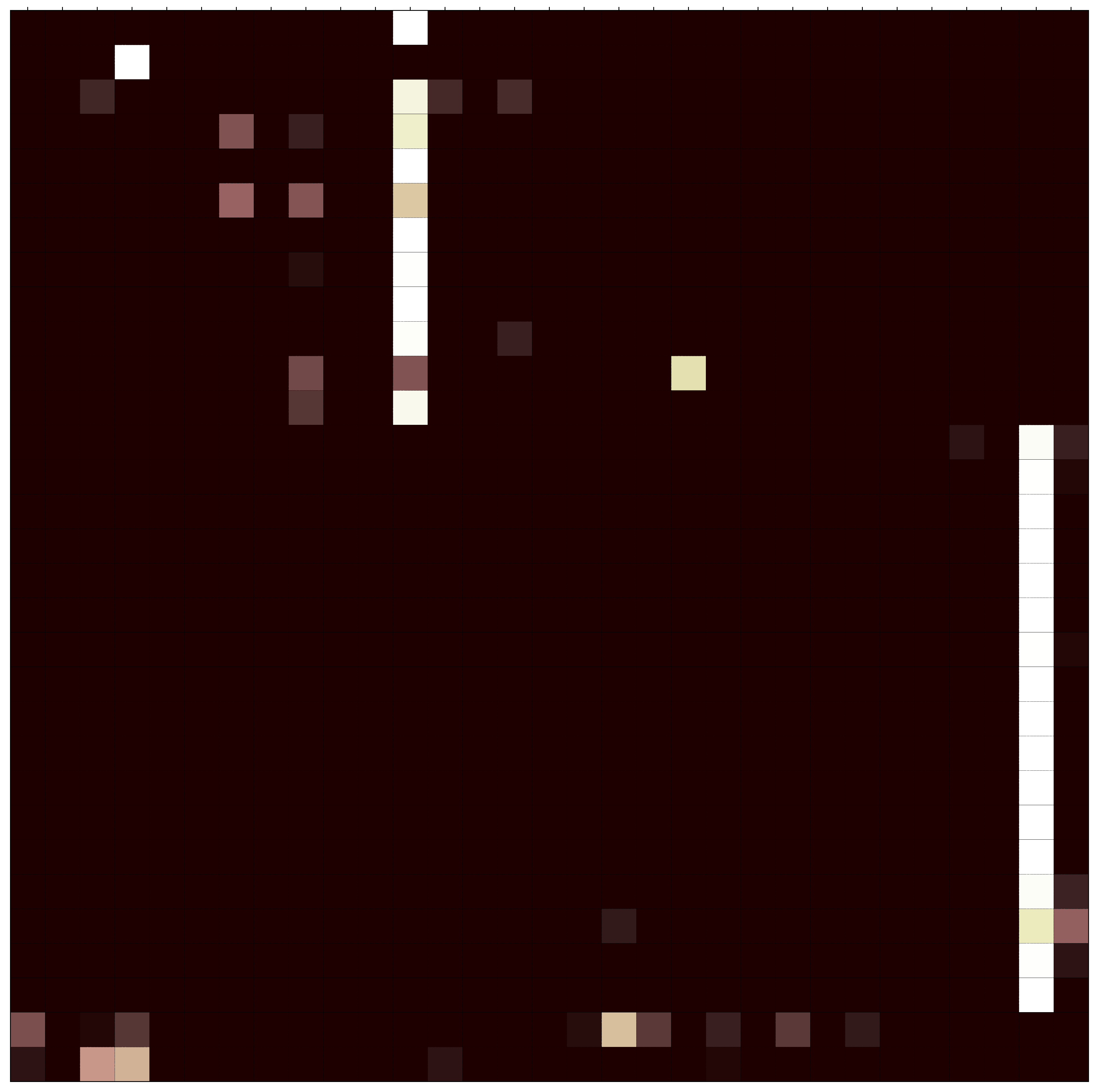}
\includegraphics[width=0.15\textwidth]{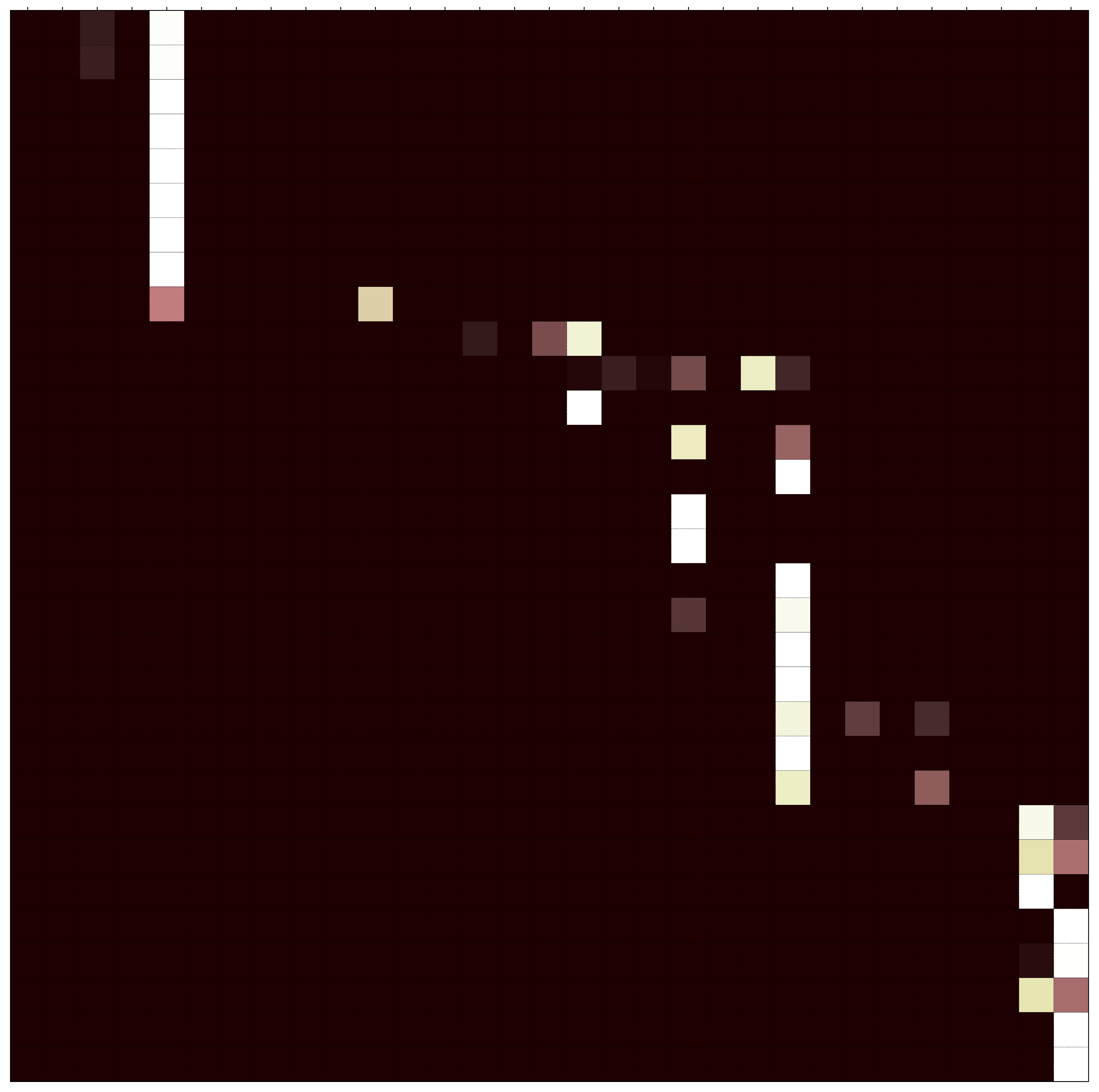}
\includegraphics[width=0.15\textwidth]{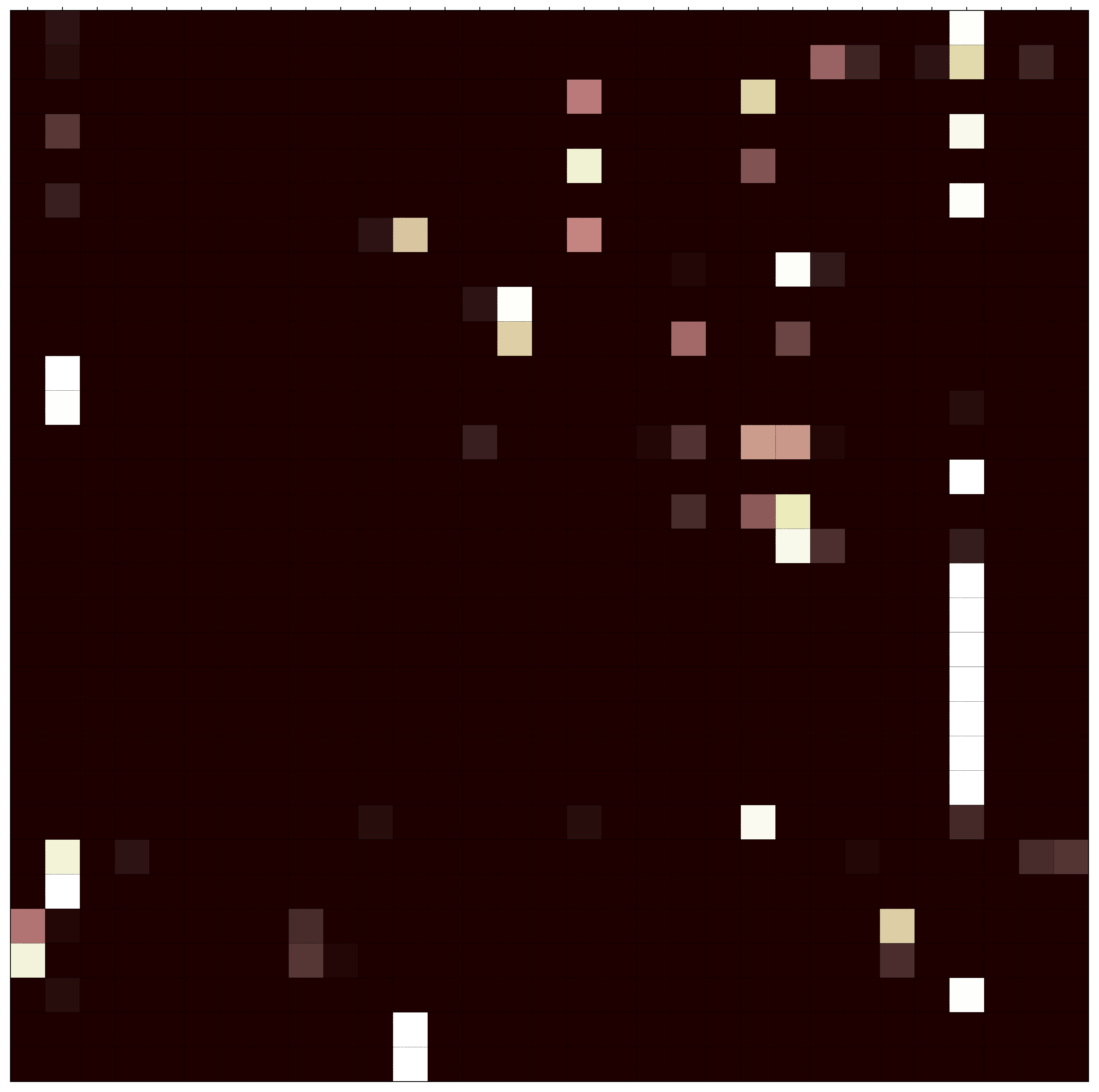}
\end{center}
\caption{Layer 2}
\end{figure}

\begin{figure}
\begin{center}
\includegraphics[width=0.15\textwidth]{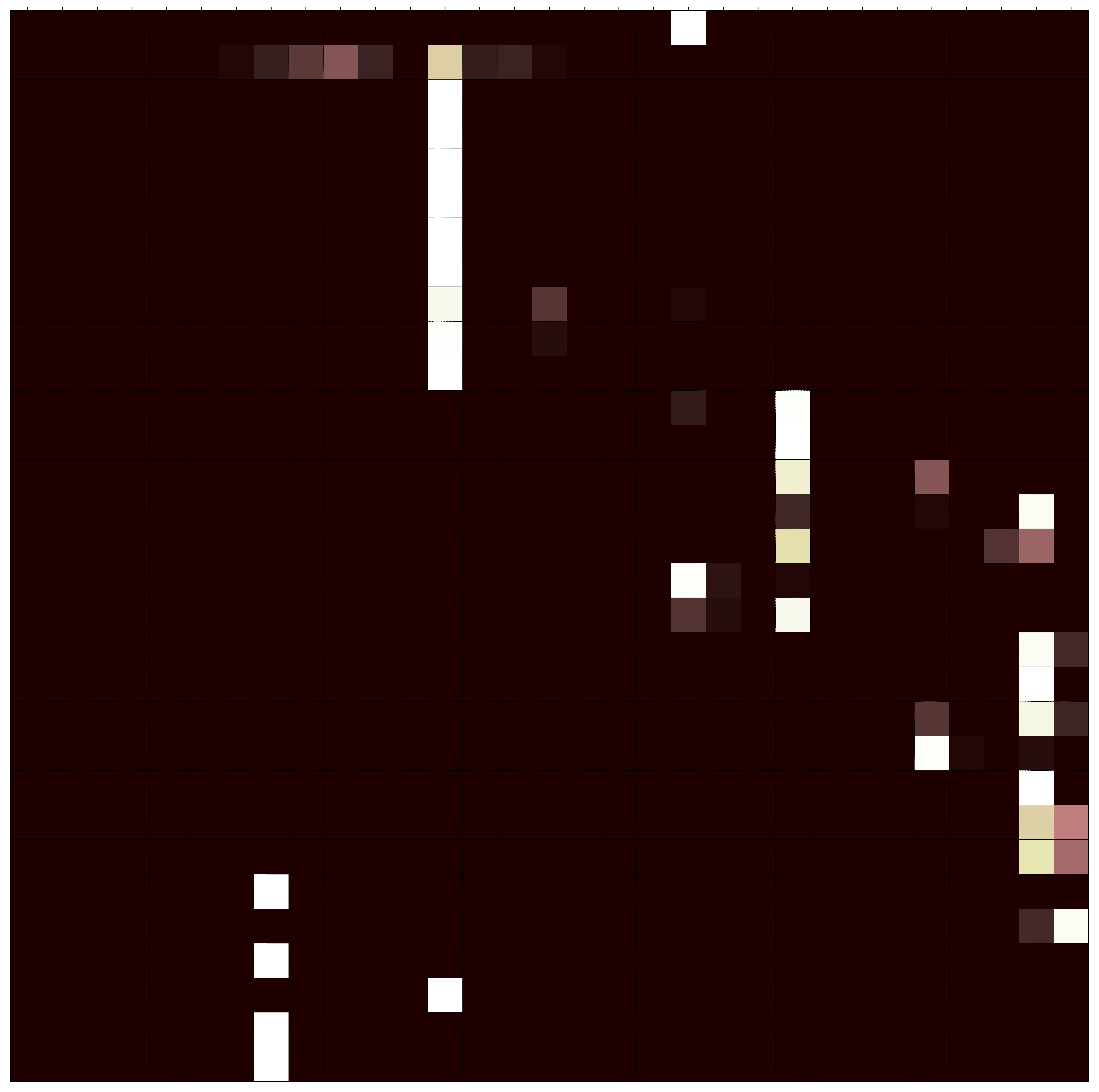}
\includegraphics[width=0.15\textwidth]{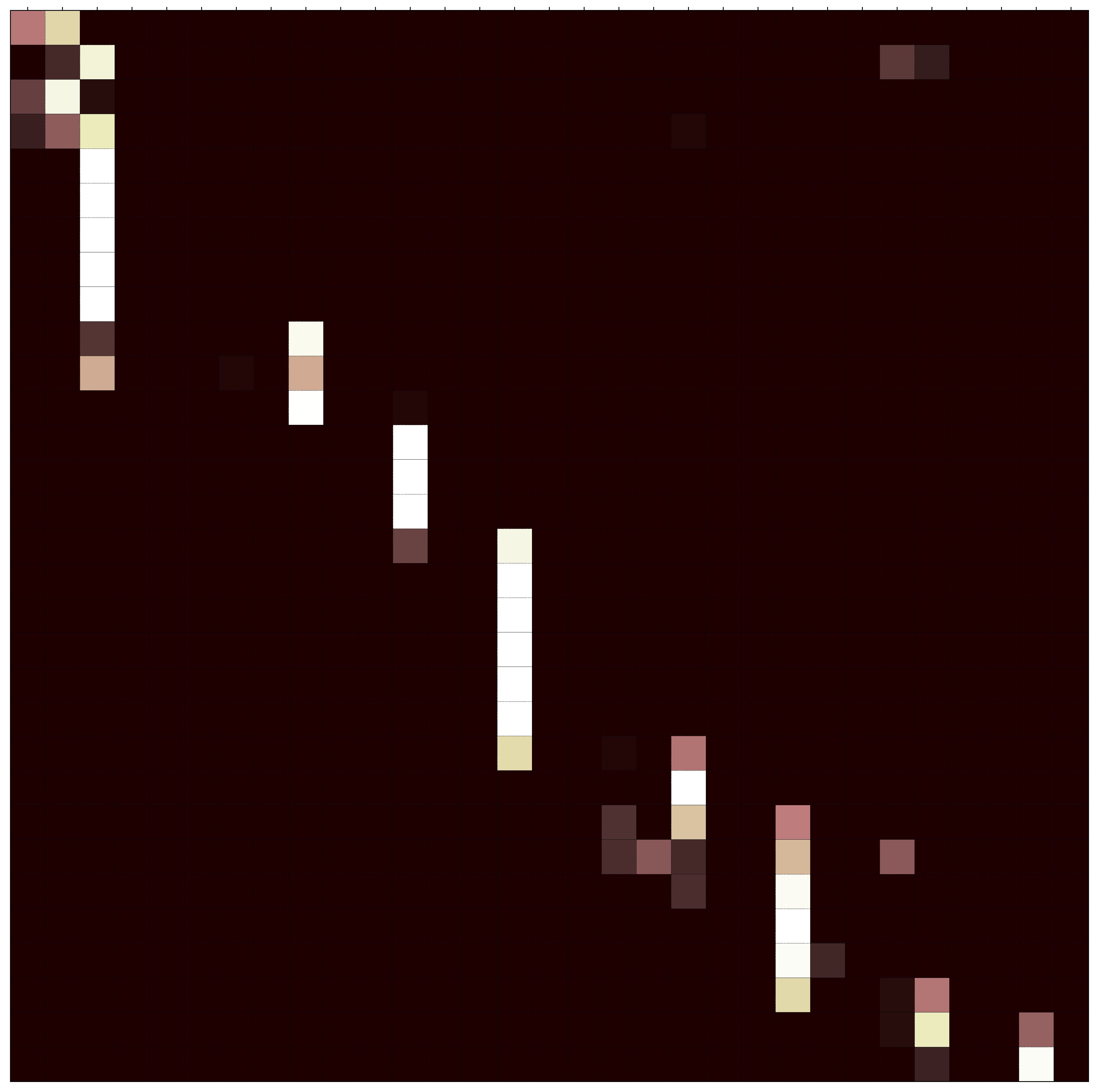}
\includegraphics[width=0.15\textwidth]{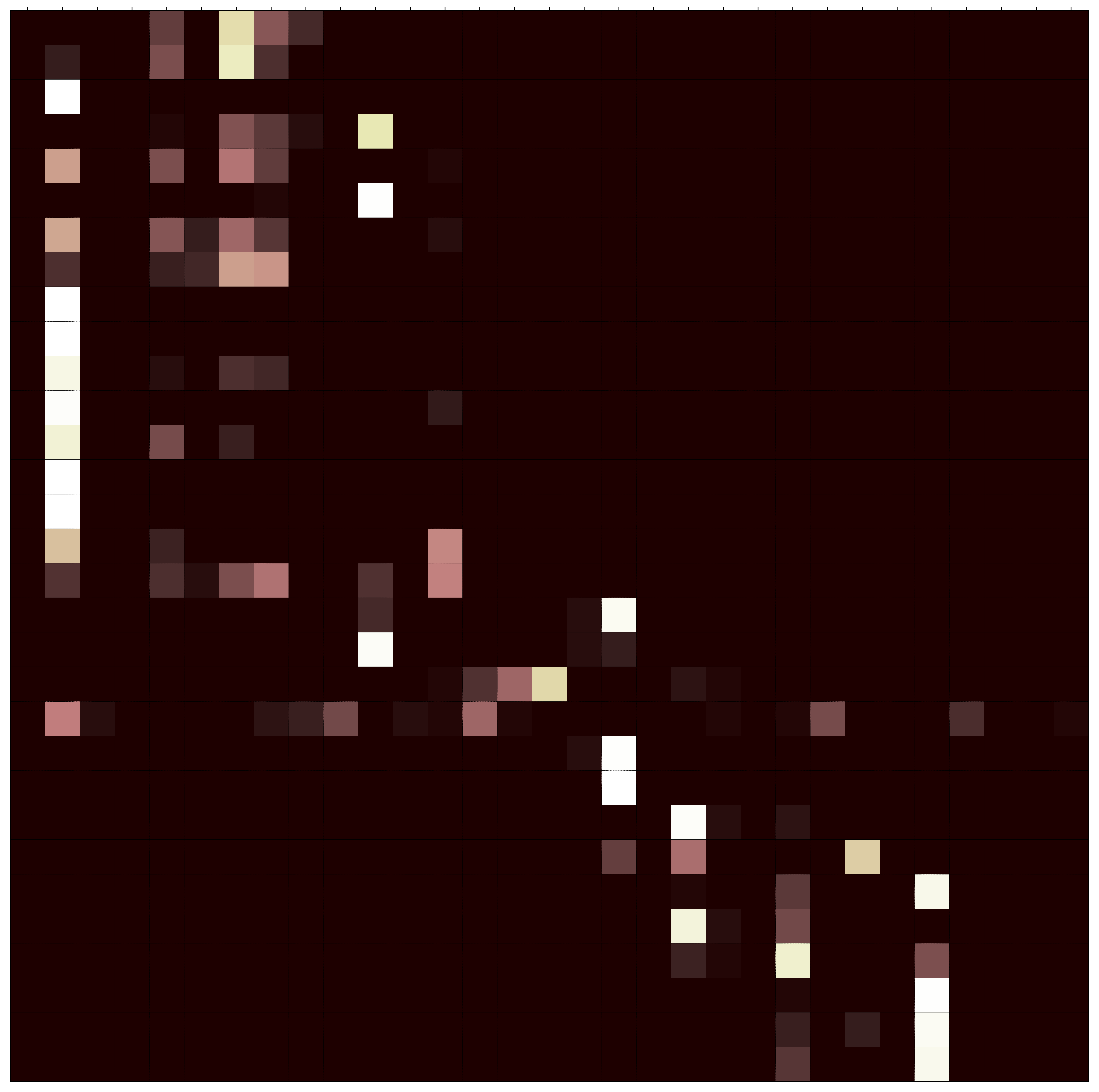}
\includegraphics[width=0.15\textwidth]{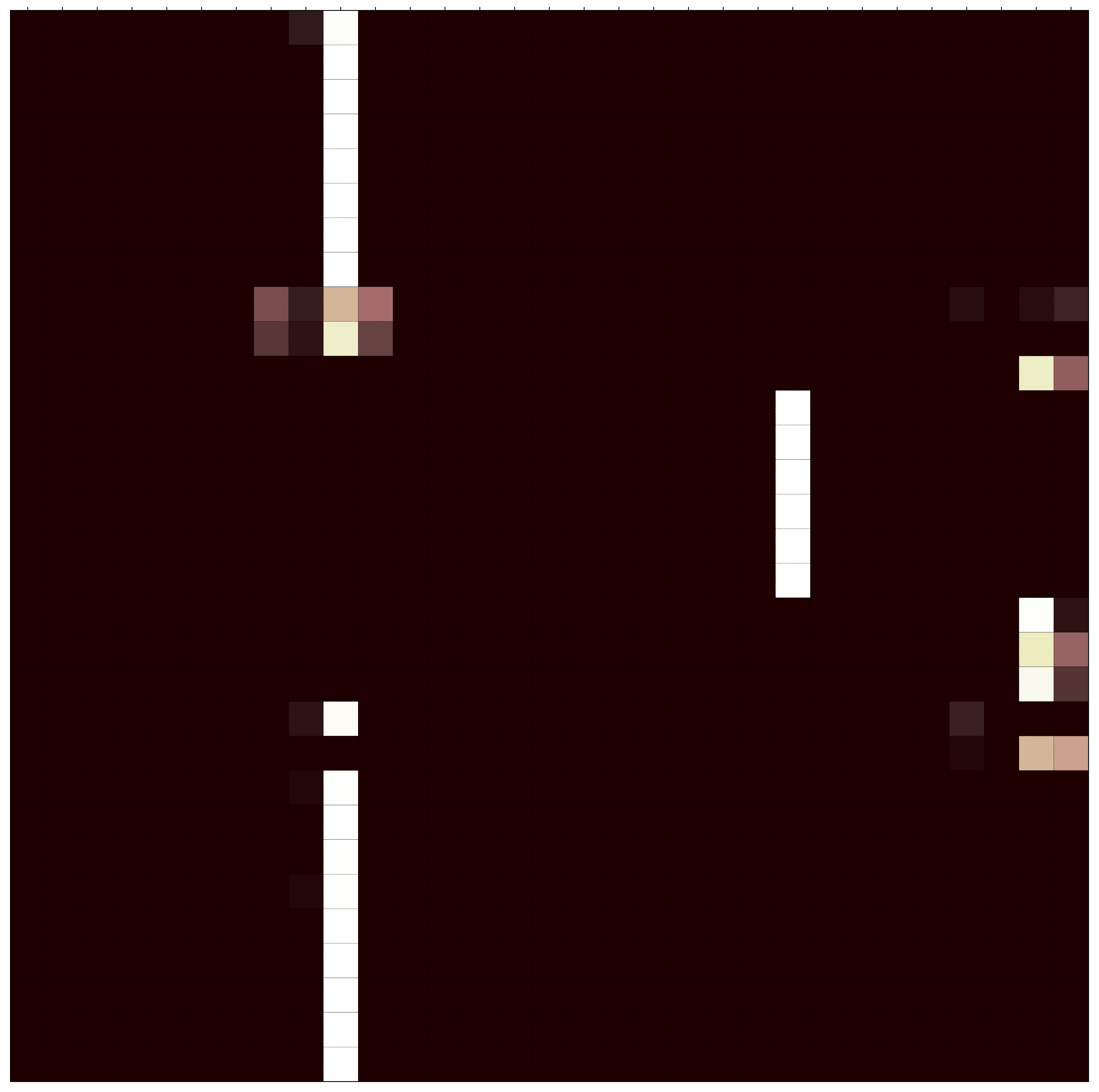}
\includegraphics[width=0.15\textwidth]{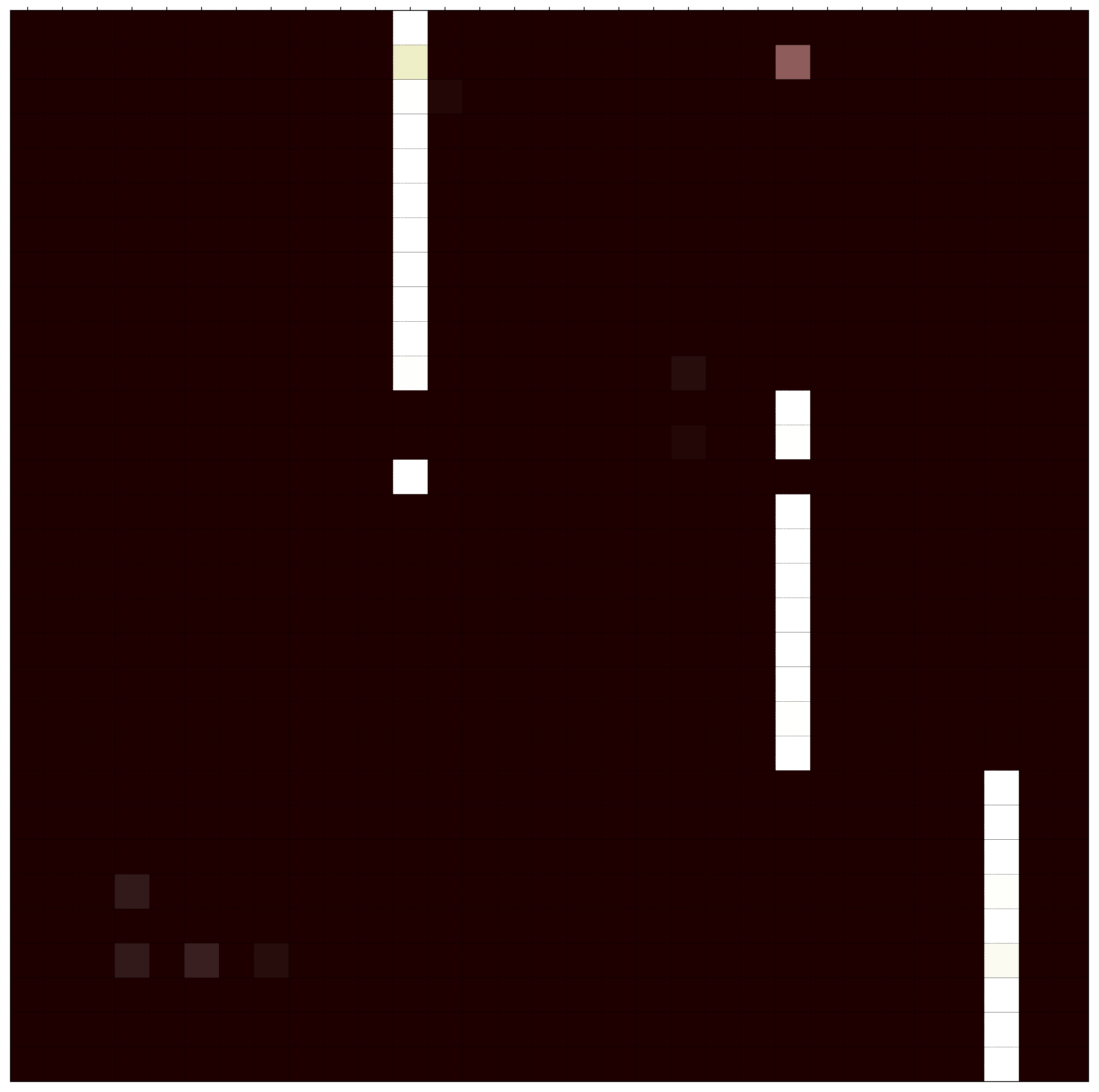}
\includegraphics[width=0.15\textwidth]{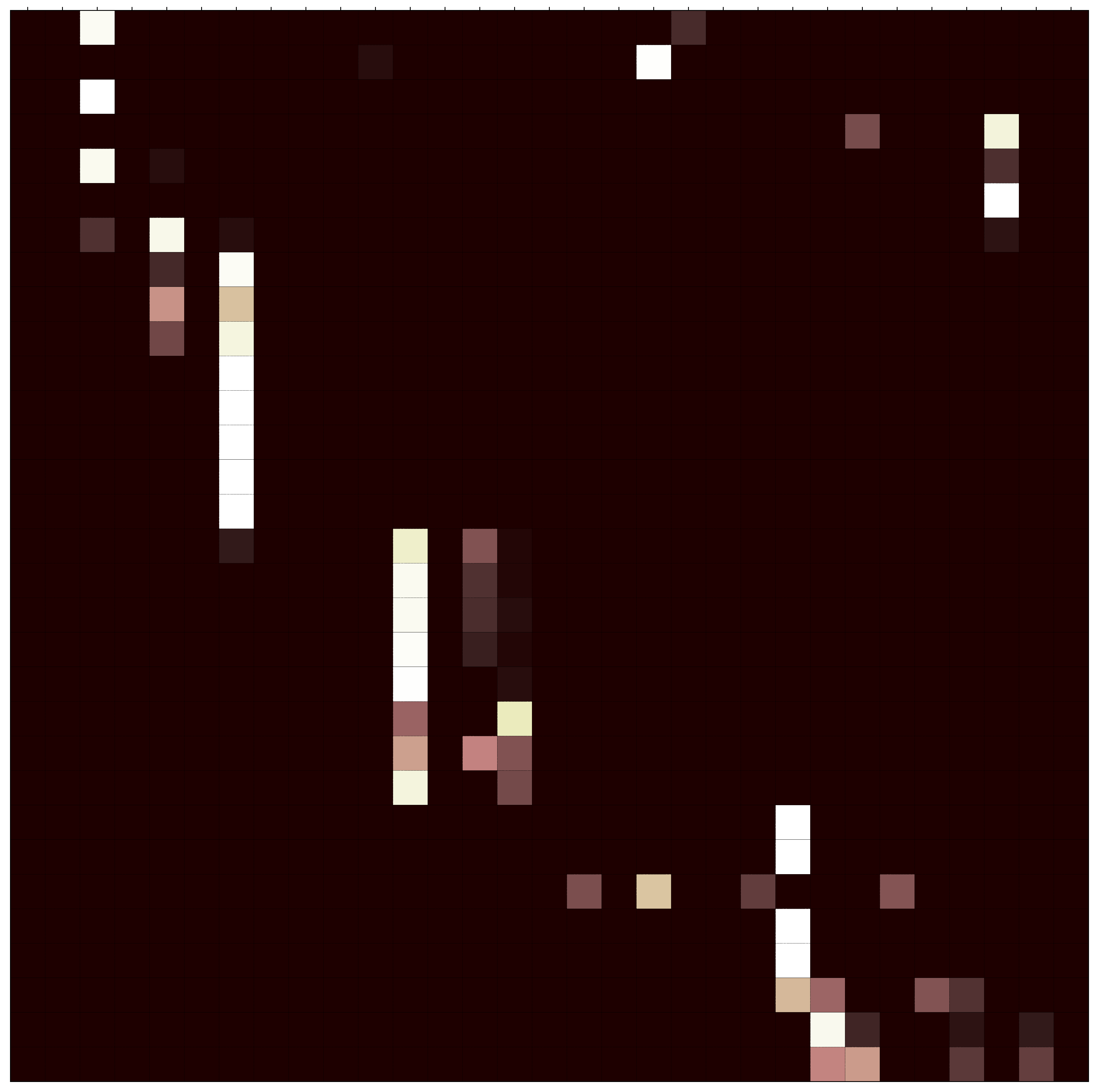}
\includegraphics[width=0.15\textwidth]{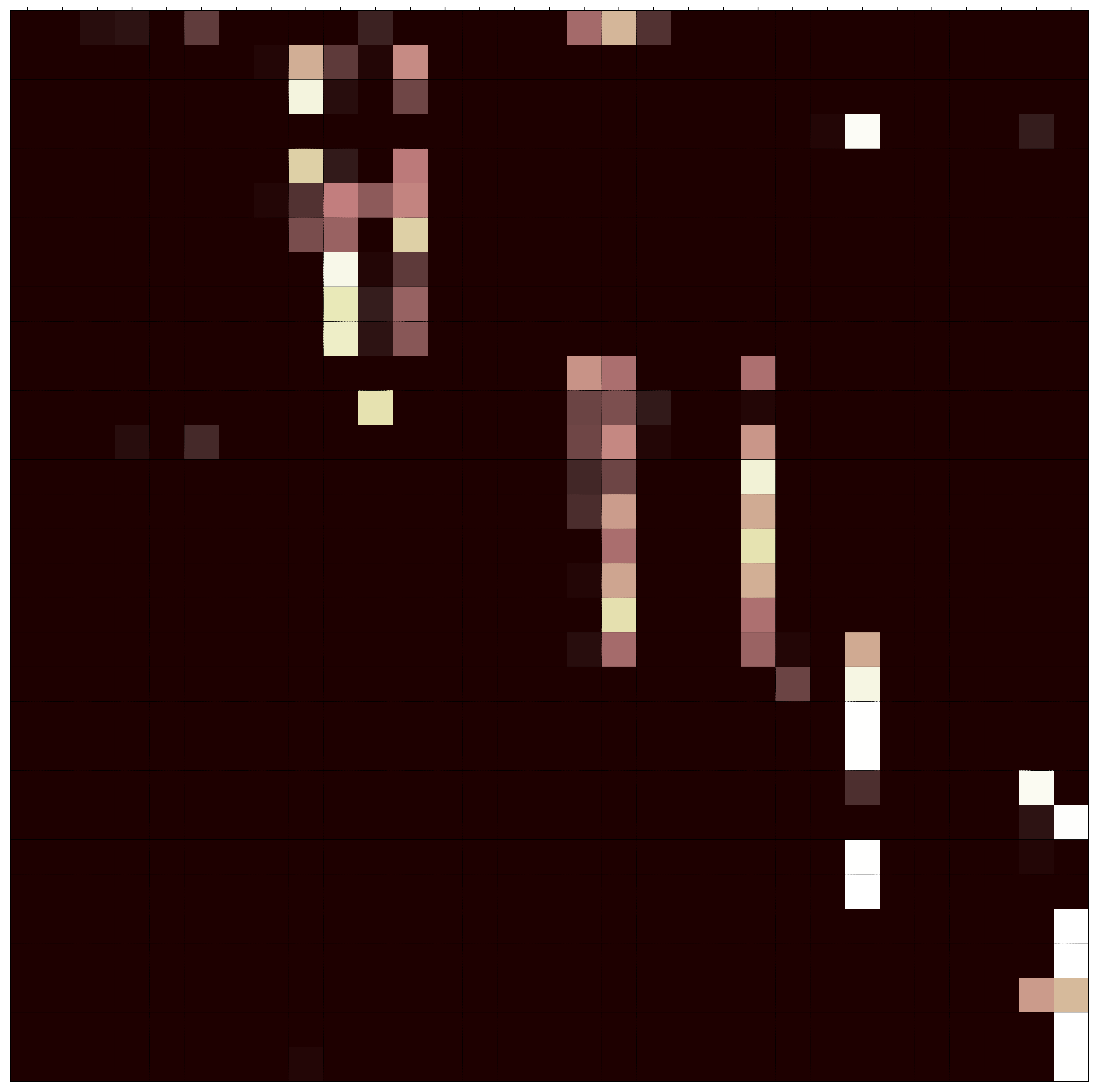}
\includegraphics[width=0.15\textwidth]{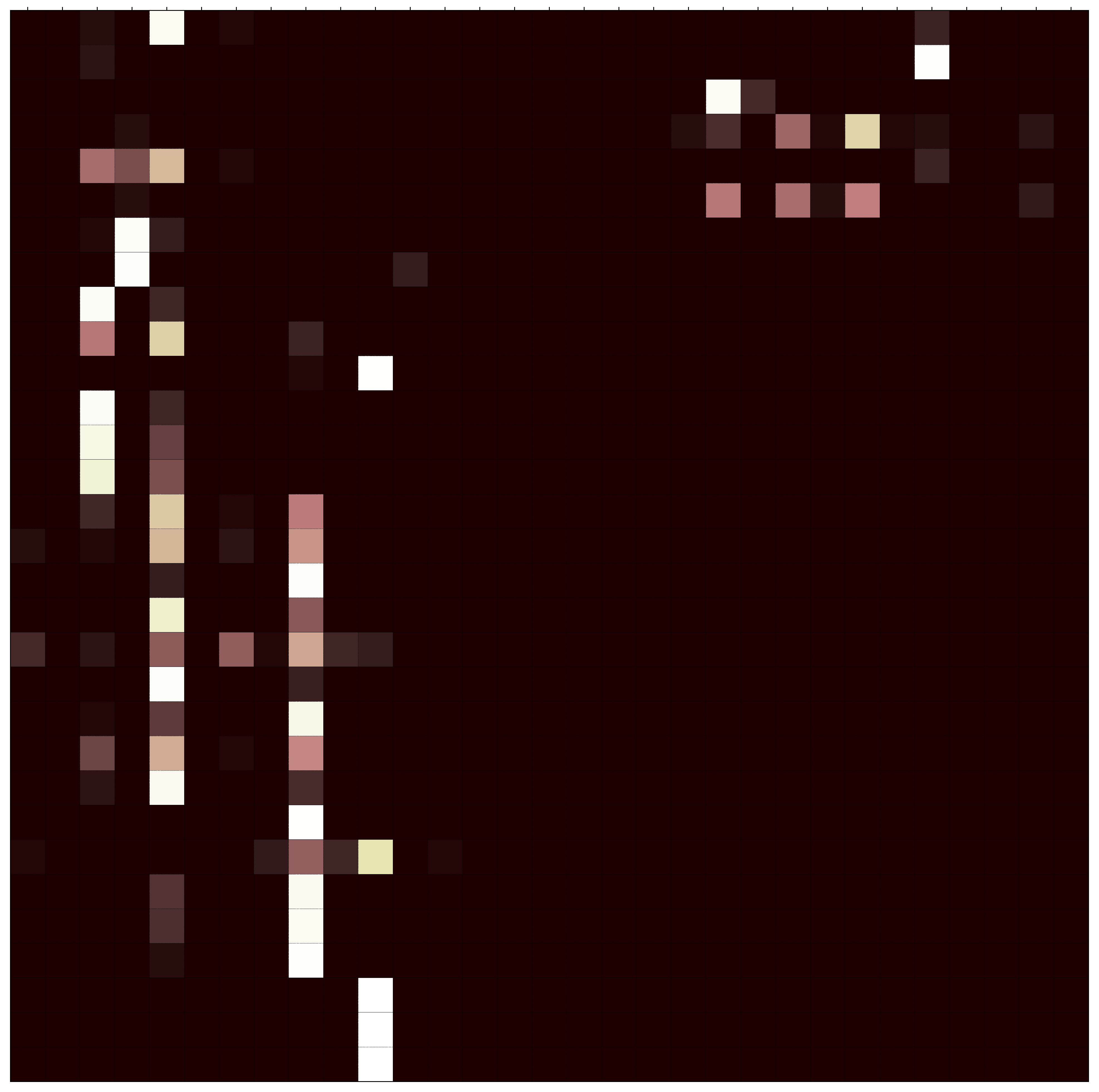}
\includegraphics[width=0.15\textwidth]{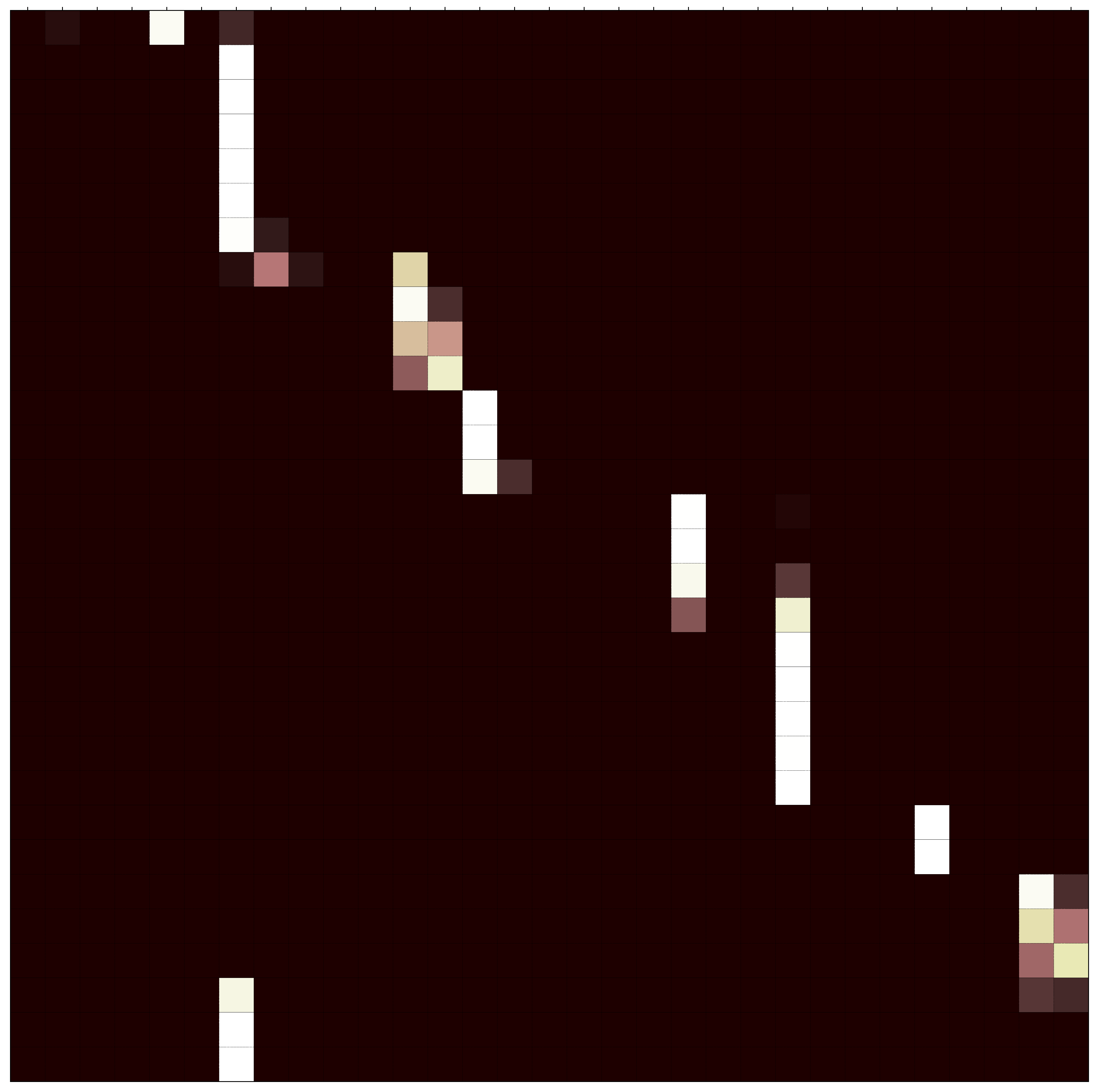}
\includegraphics[width=0.15\textwidth]{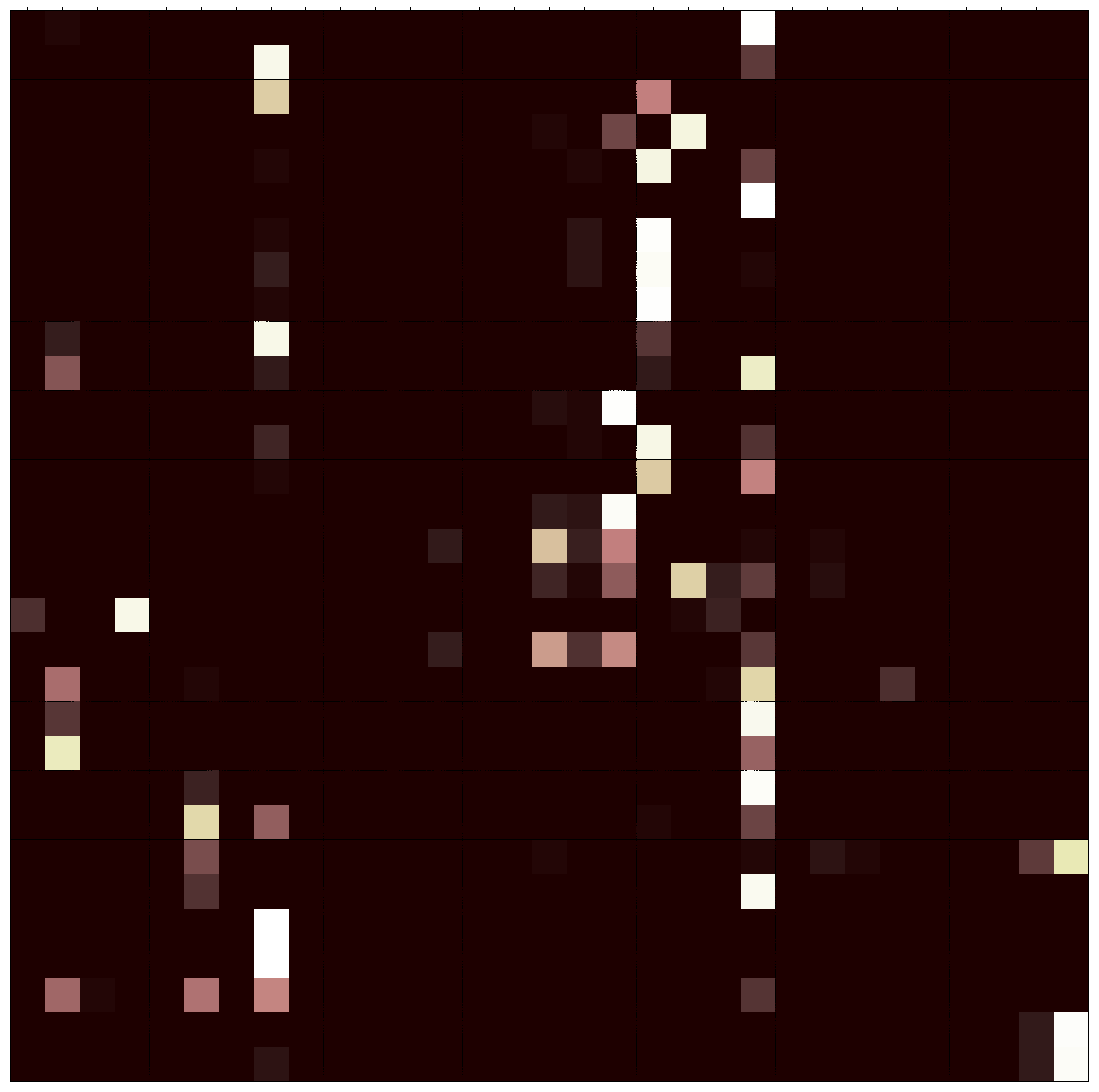}
\includegraphics[width=0.15\textwidth]{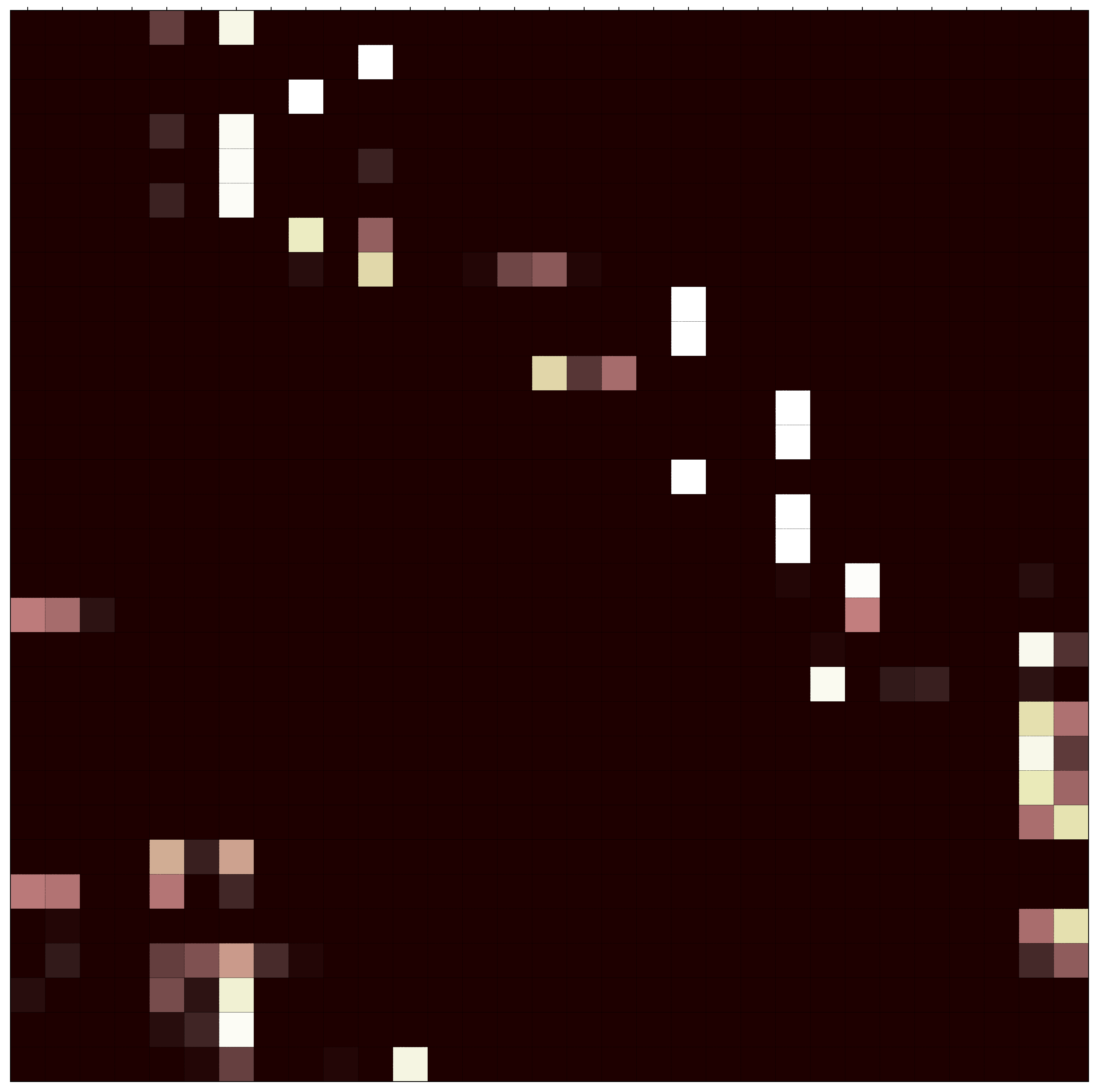}
\includegraphics[width=0.15\textwidth]{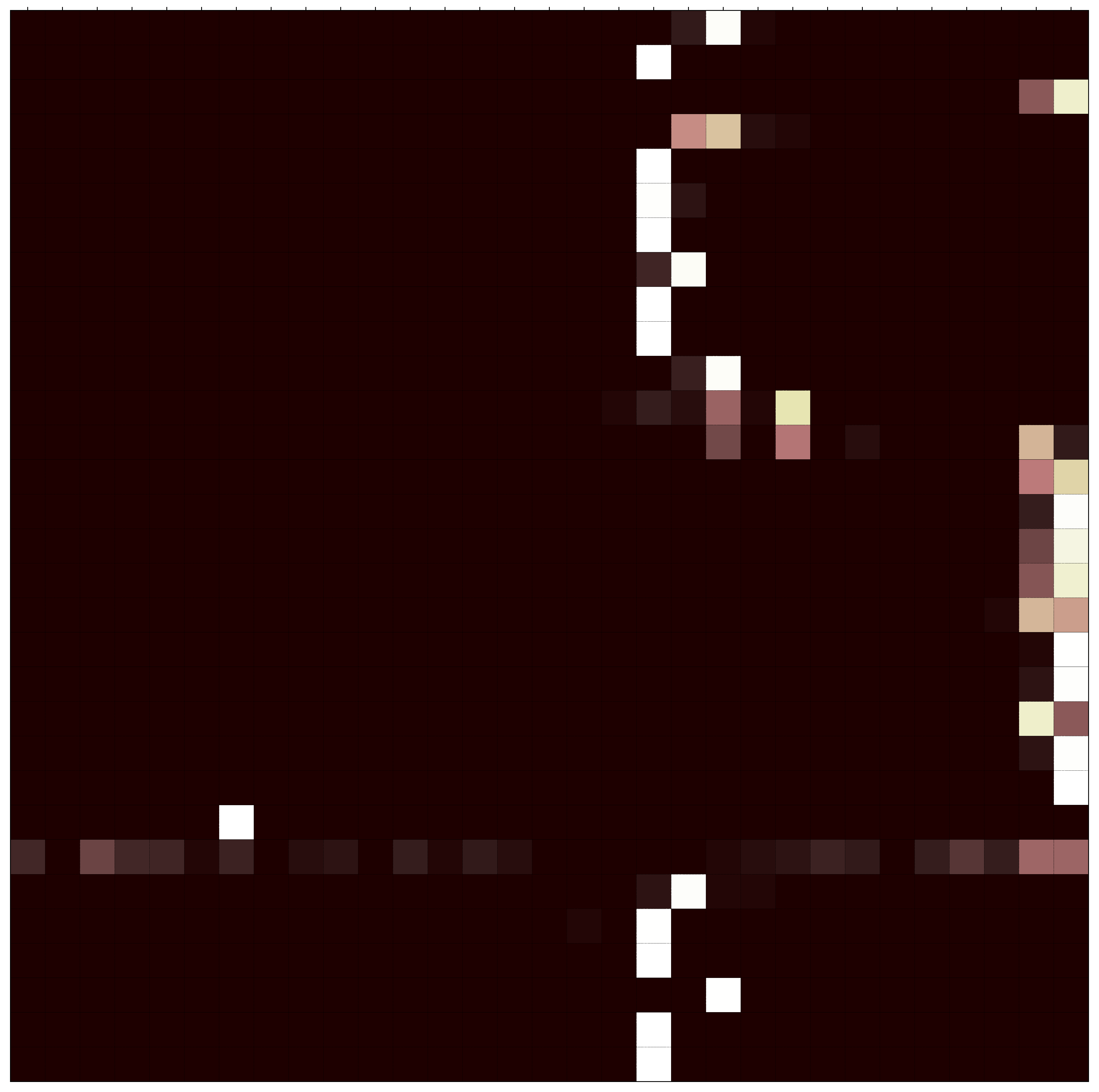}
\includegraphics[width=0.15\textwidth]{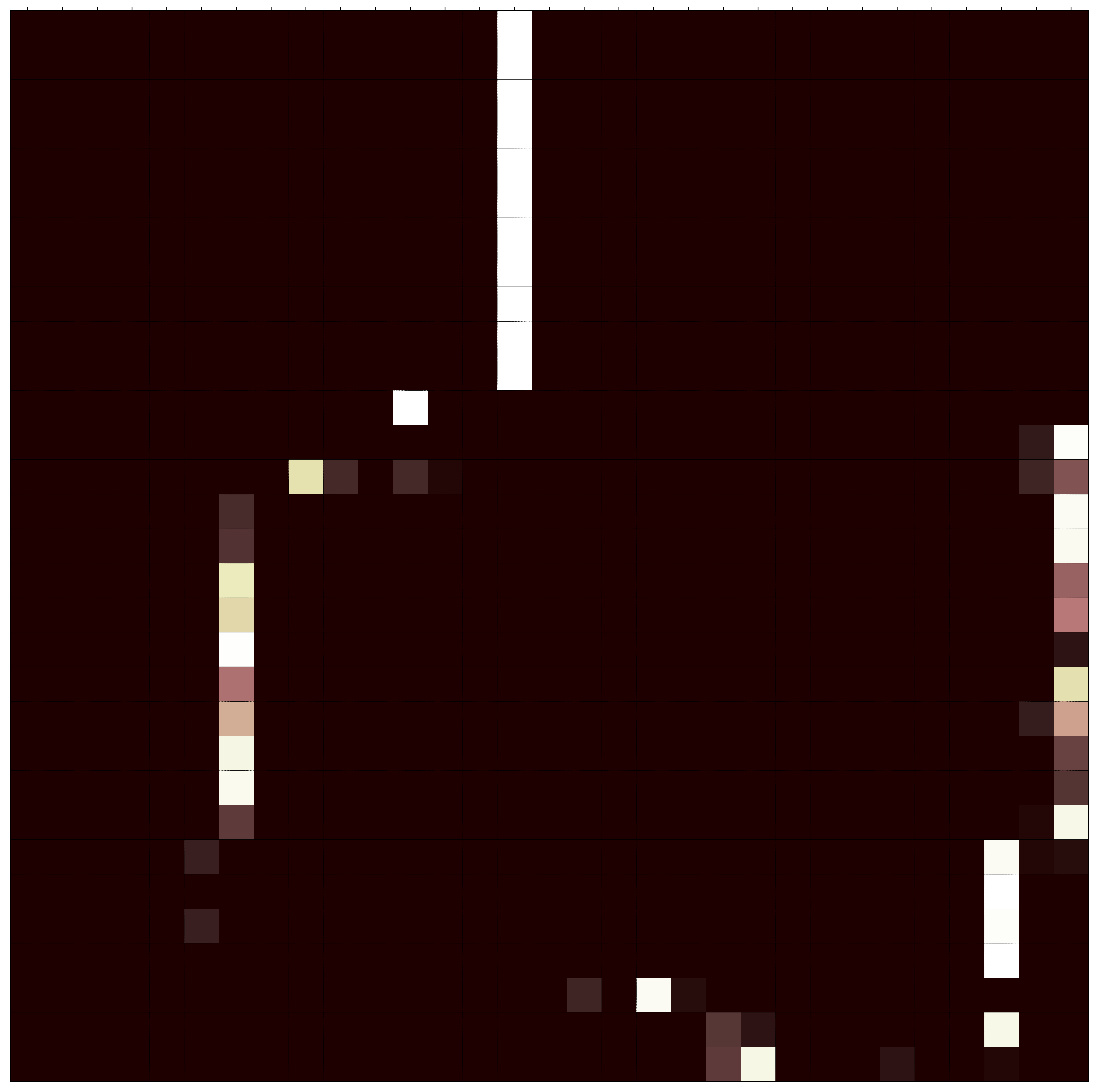}
\includegraphics[width=0.15\textwidth]{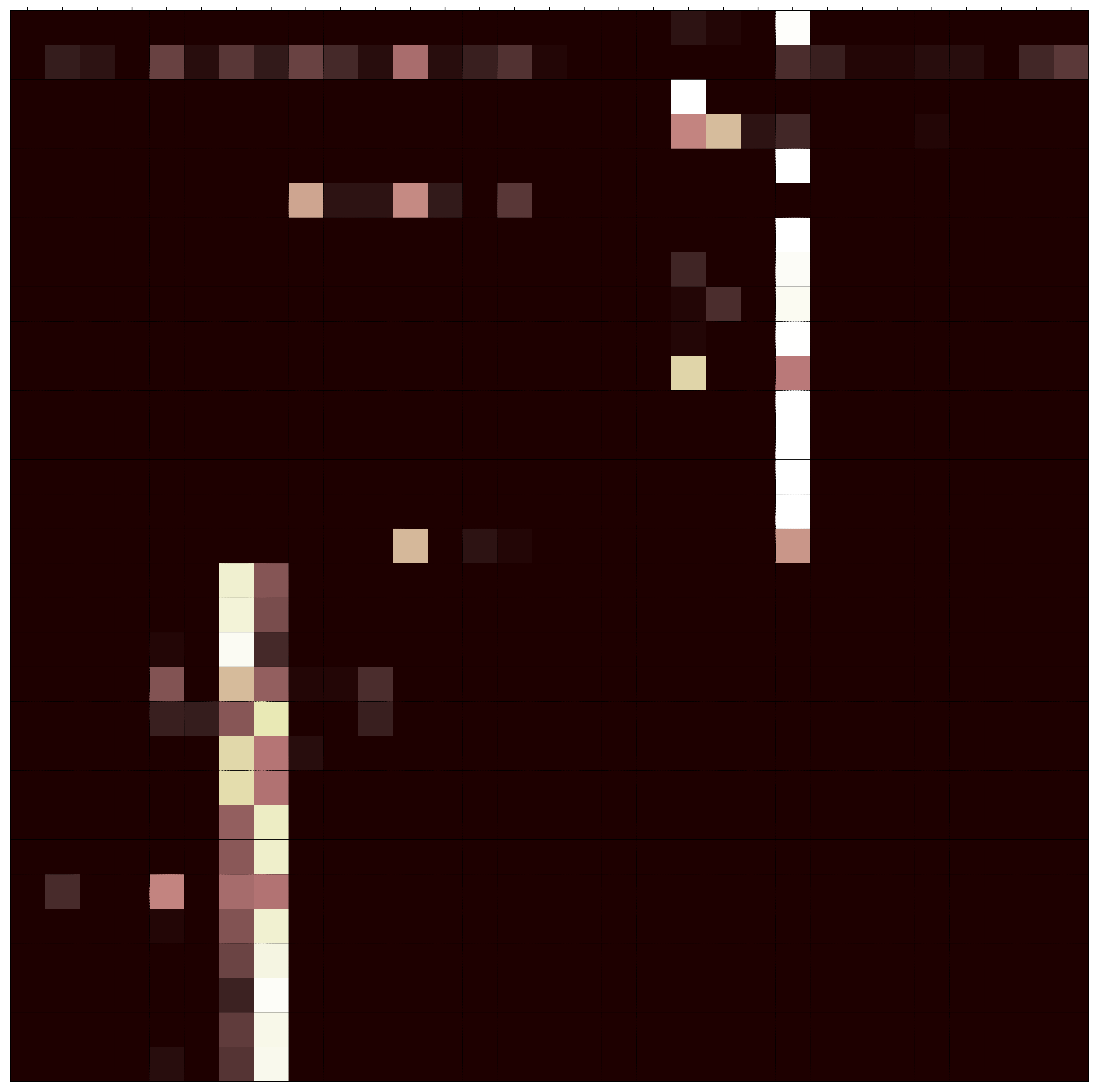}
\includegraphics[width=0.15\textwidth]{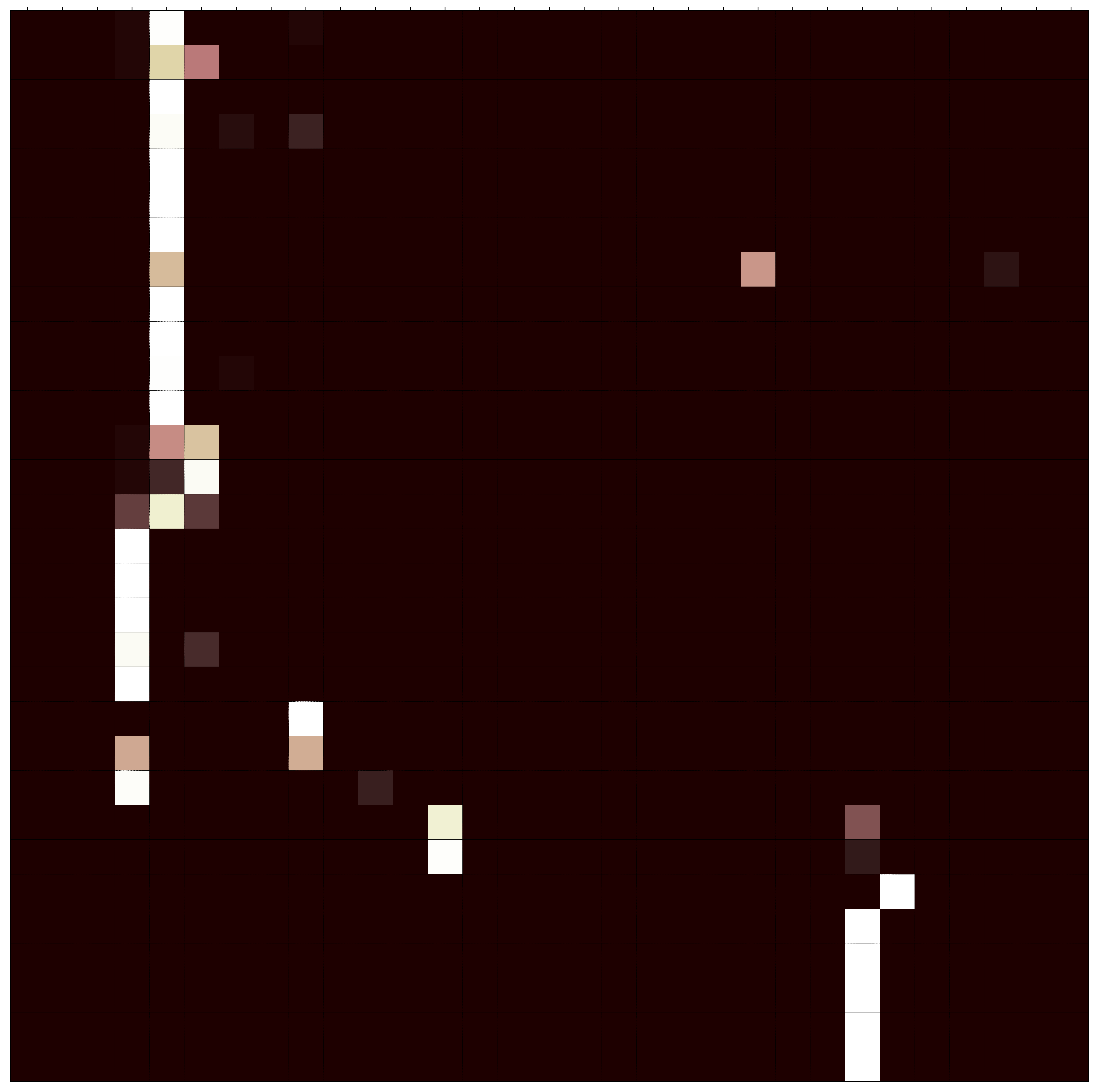}
\includegraphics[width=0.15\textwidth]{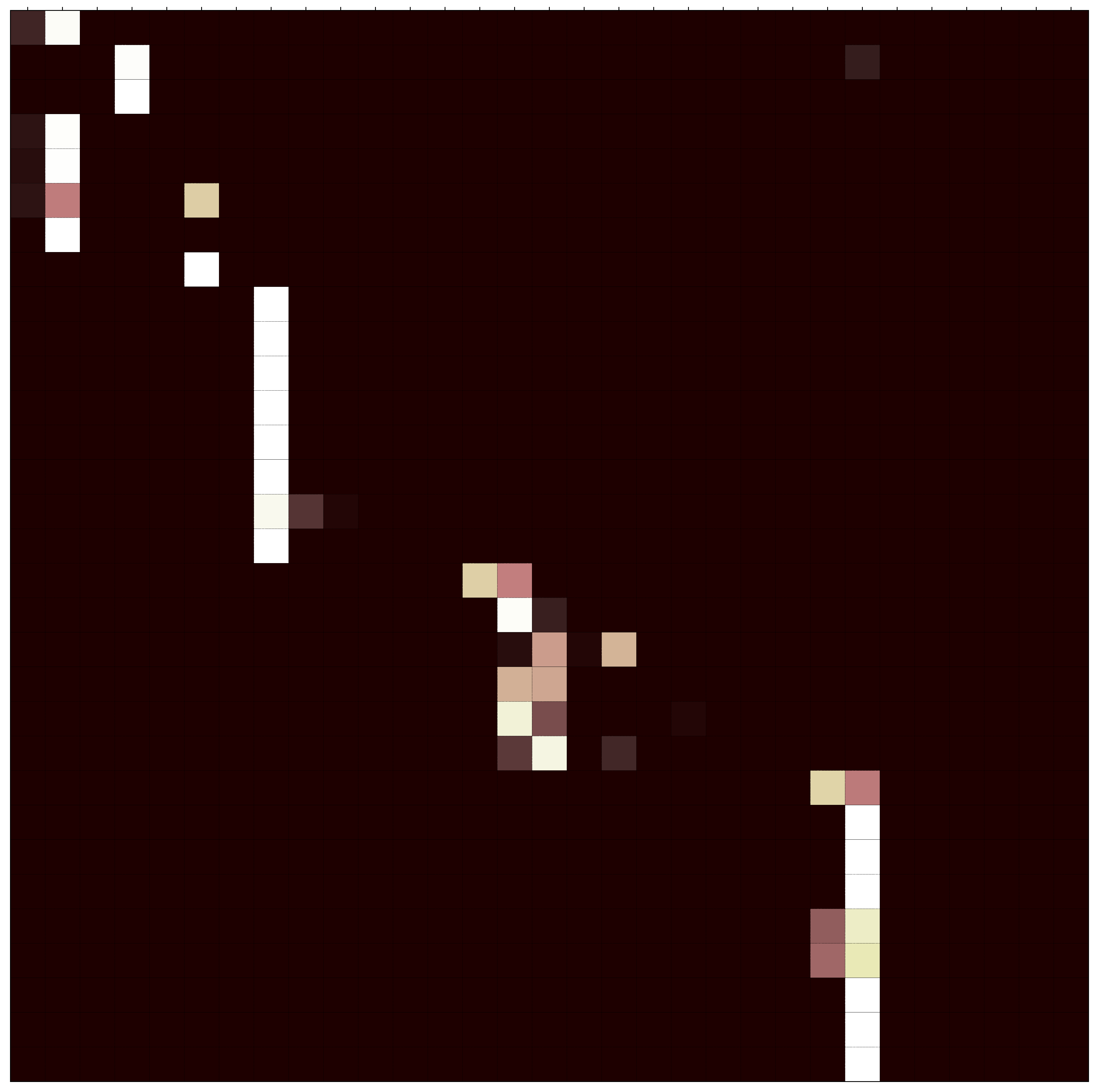}
\end{center}
\caption{Layer 3}
\end{figure}

\begin{figure}
\begin{center}
\includegraphics[width=0.15\textwidth]{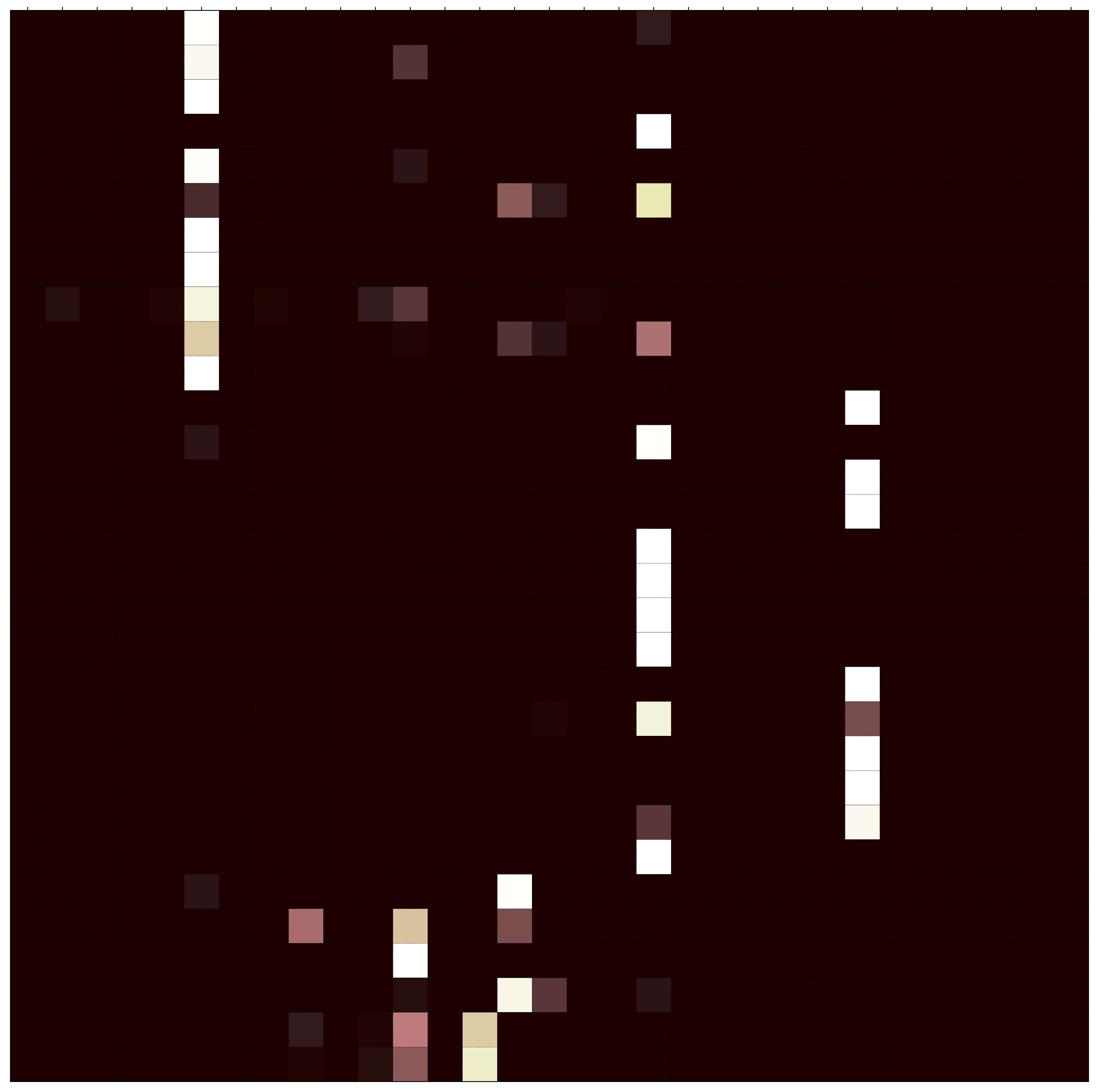}
\includegraphics[width=0.15\textwidth]{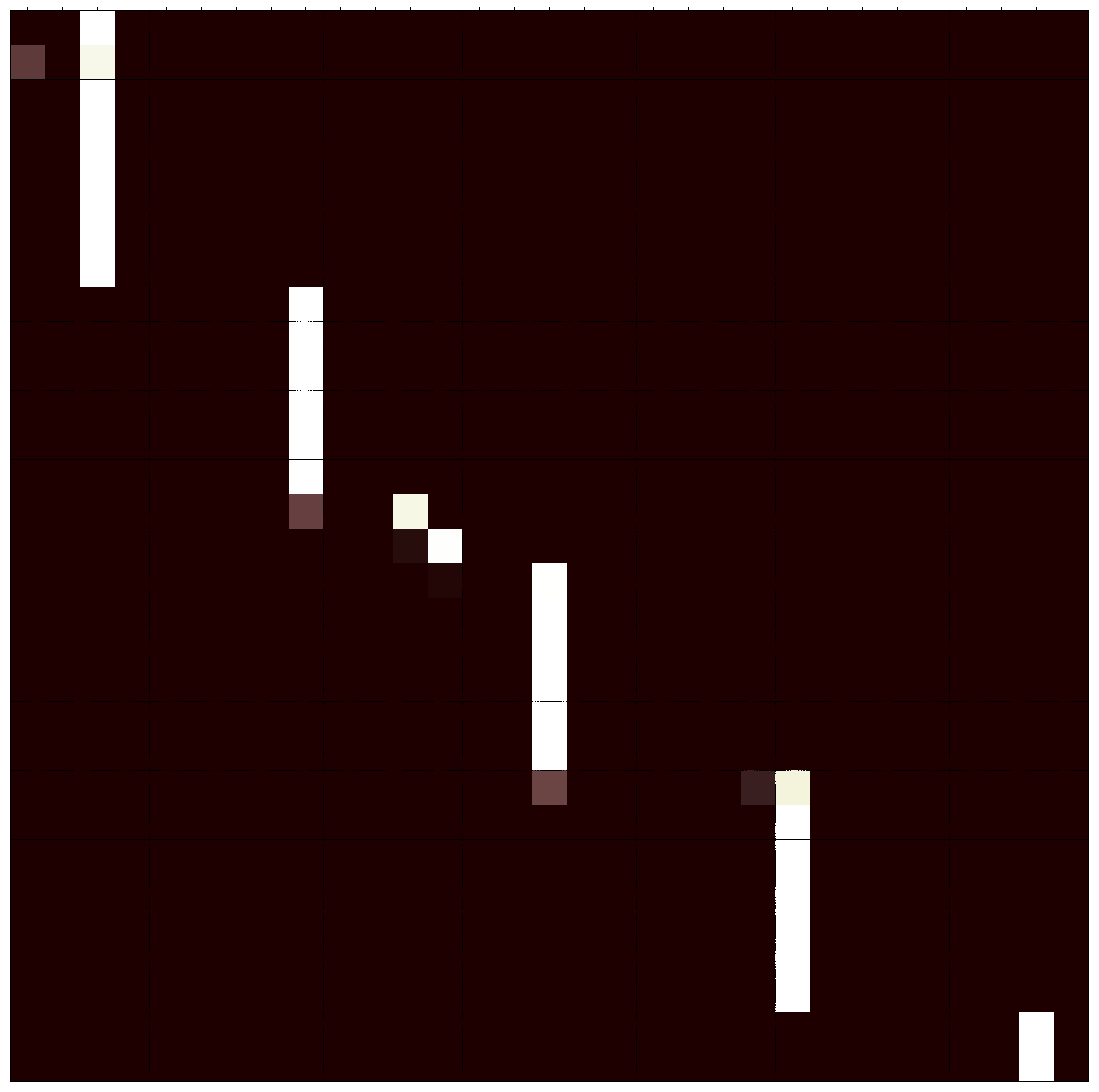}
\includegraphics[width=0.15\textwidth]{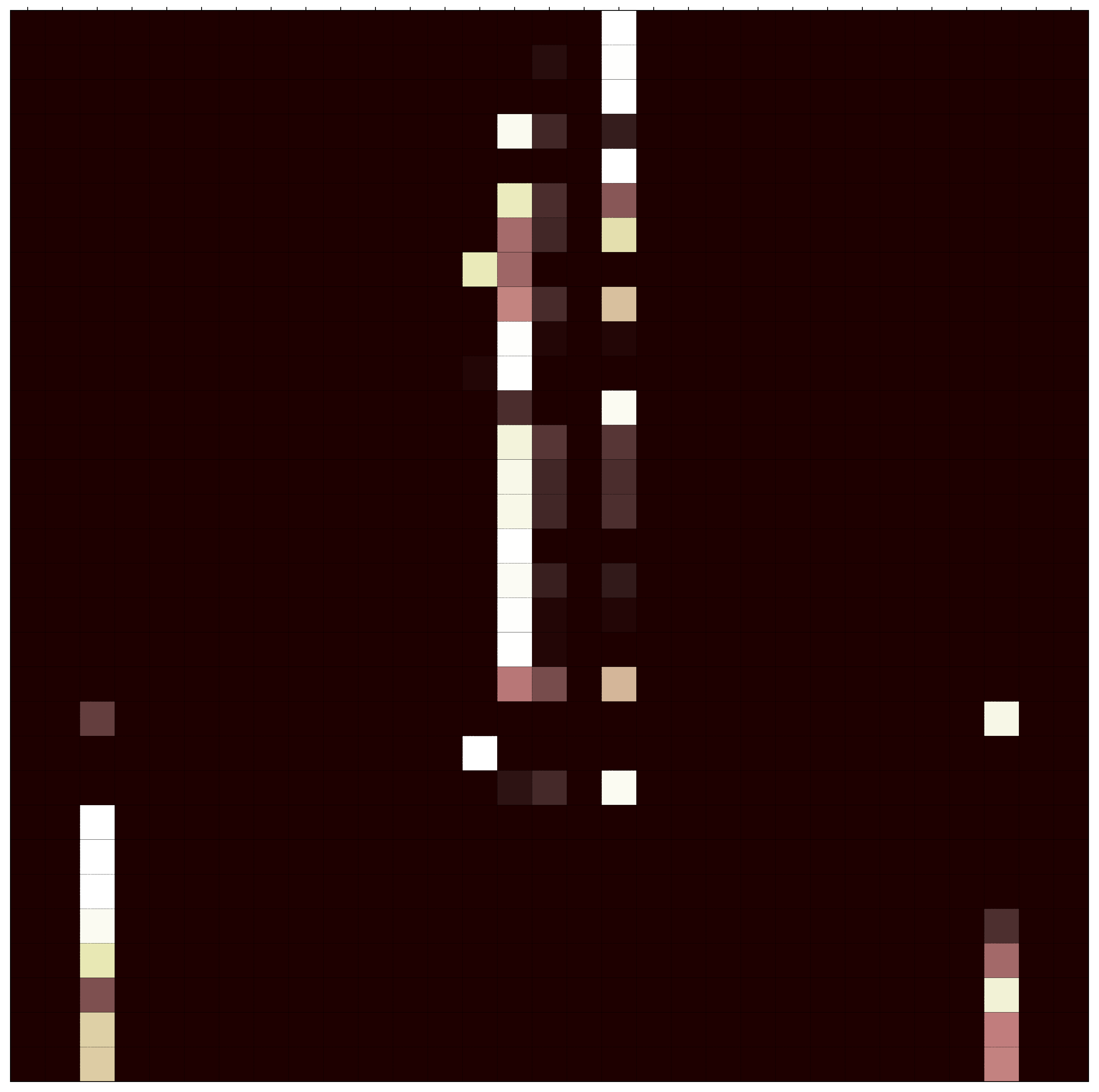}
\includegraphics[width=0.15\textwidth]{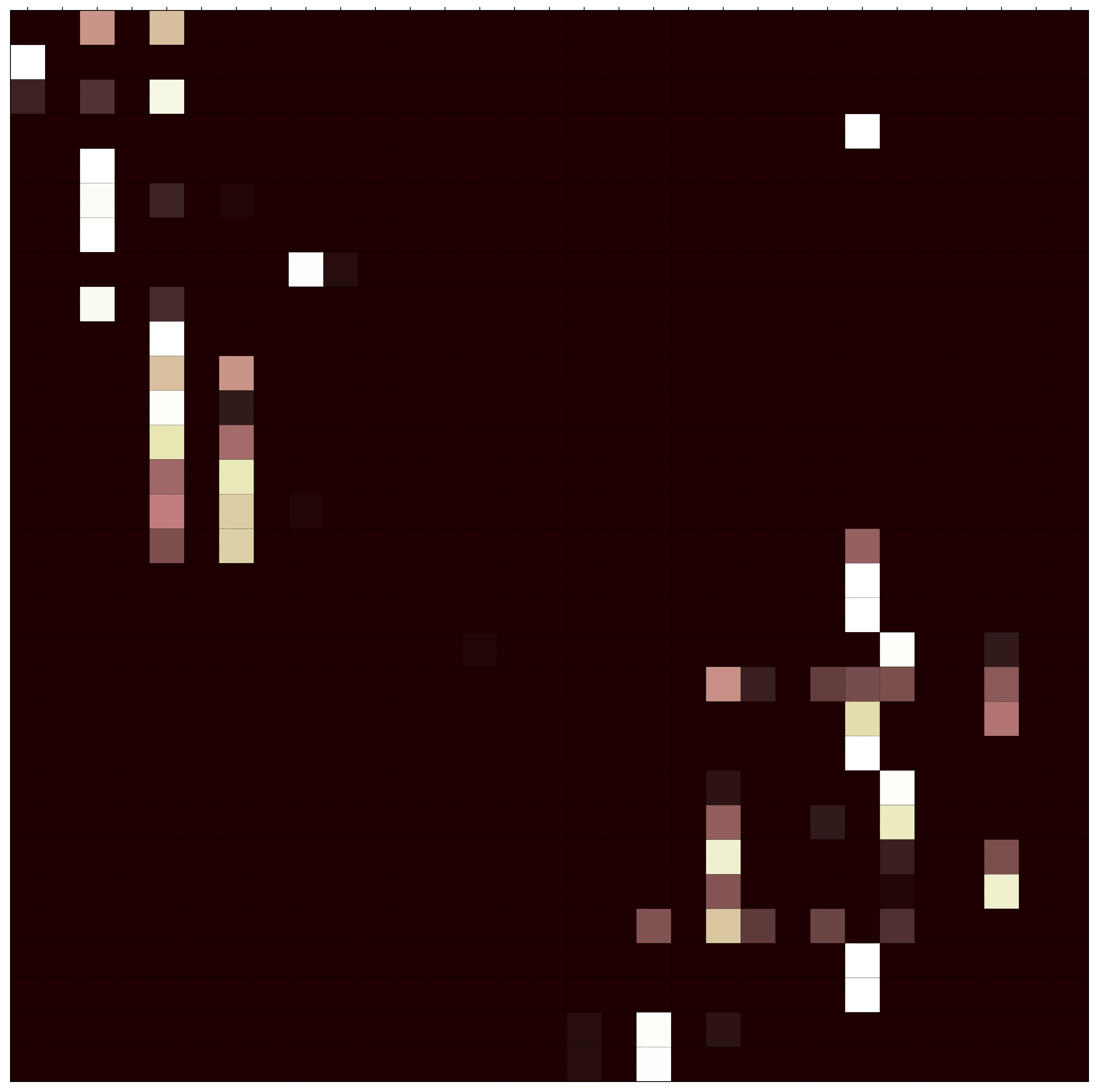}
\includegraphics[width=0.15\textwidth]{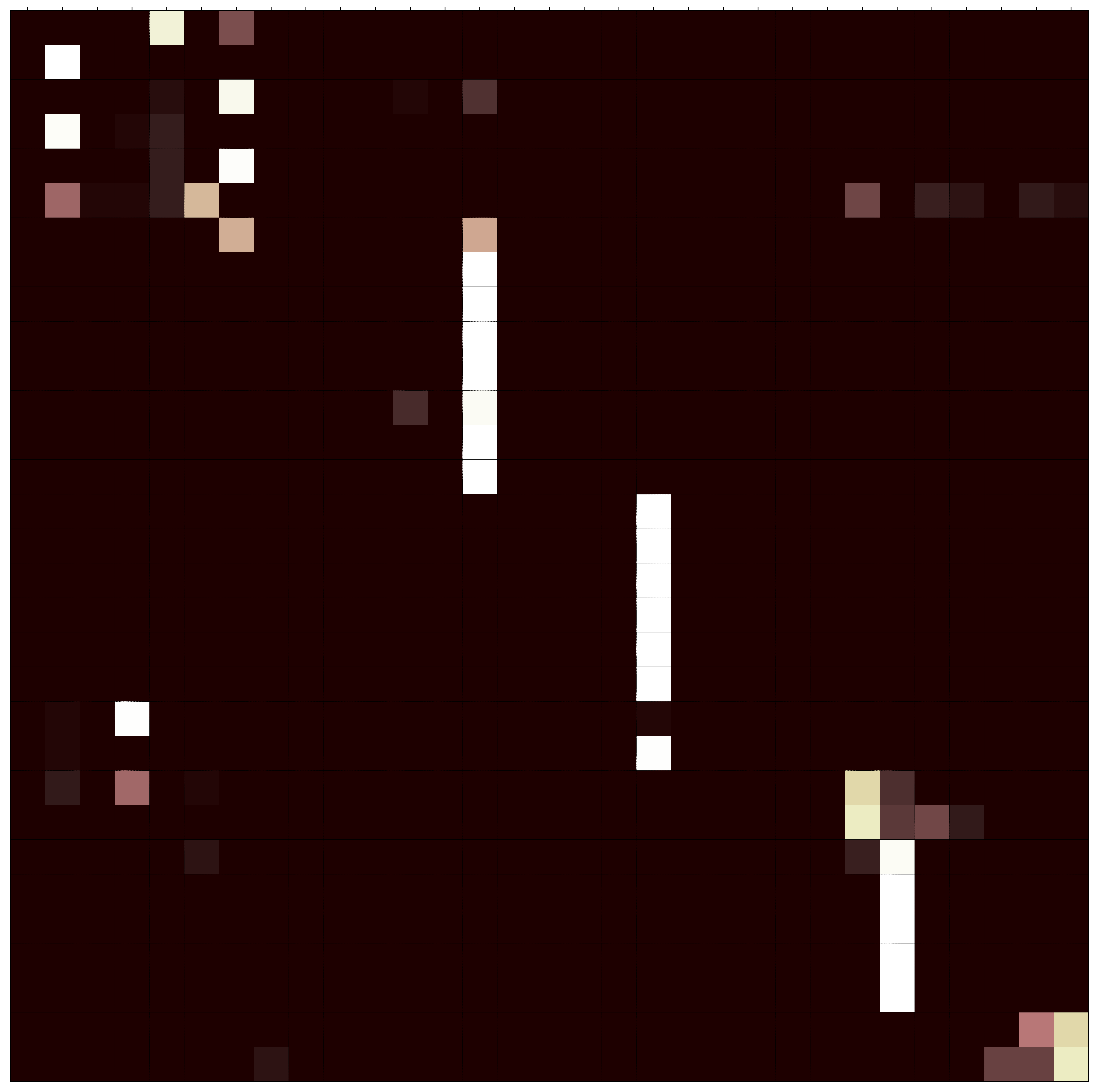}
\includegraphics[width=0.15\textwidth]{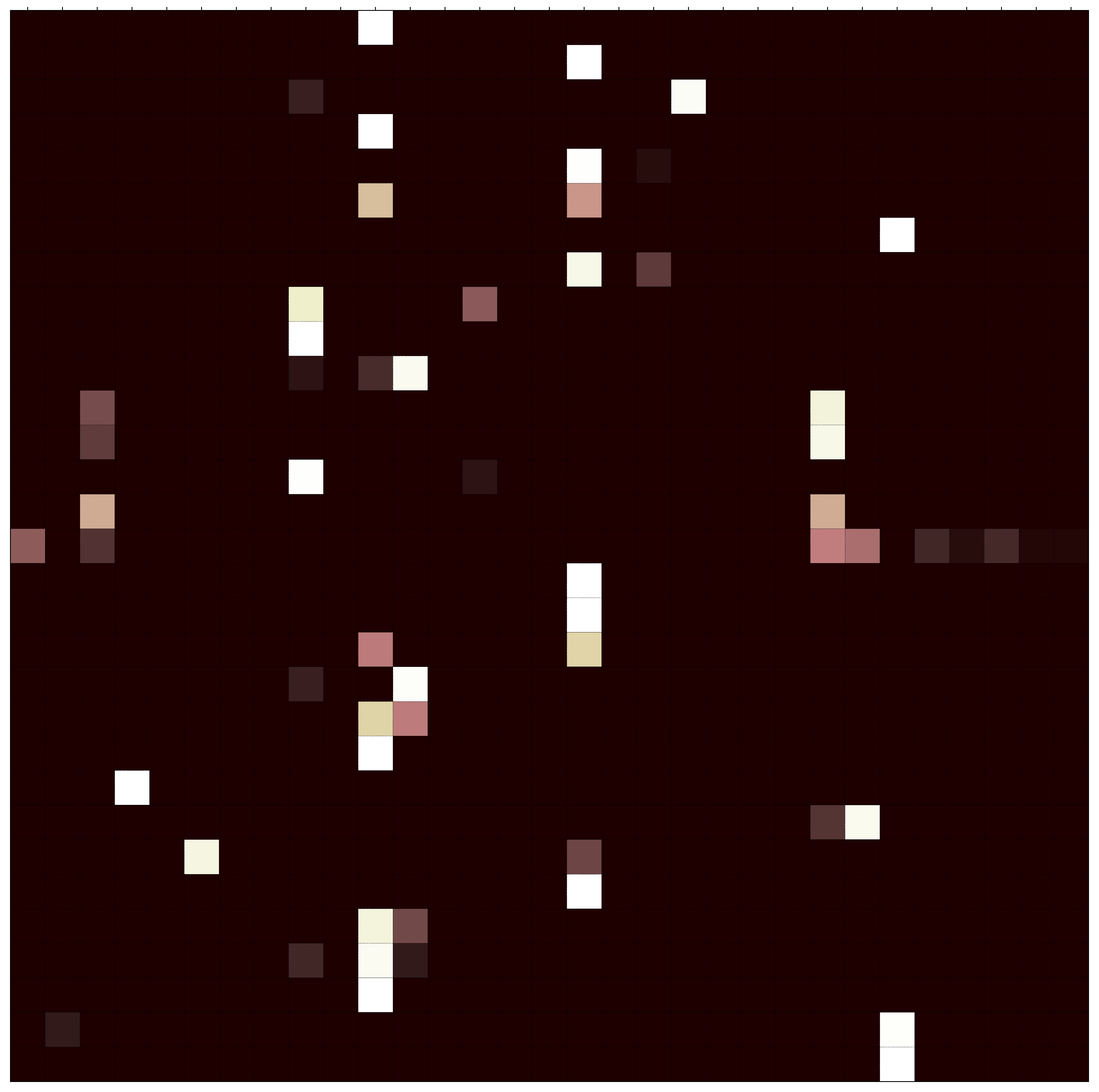}
\includegraphics[width=0.15\textwidth]{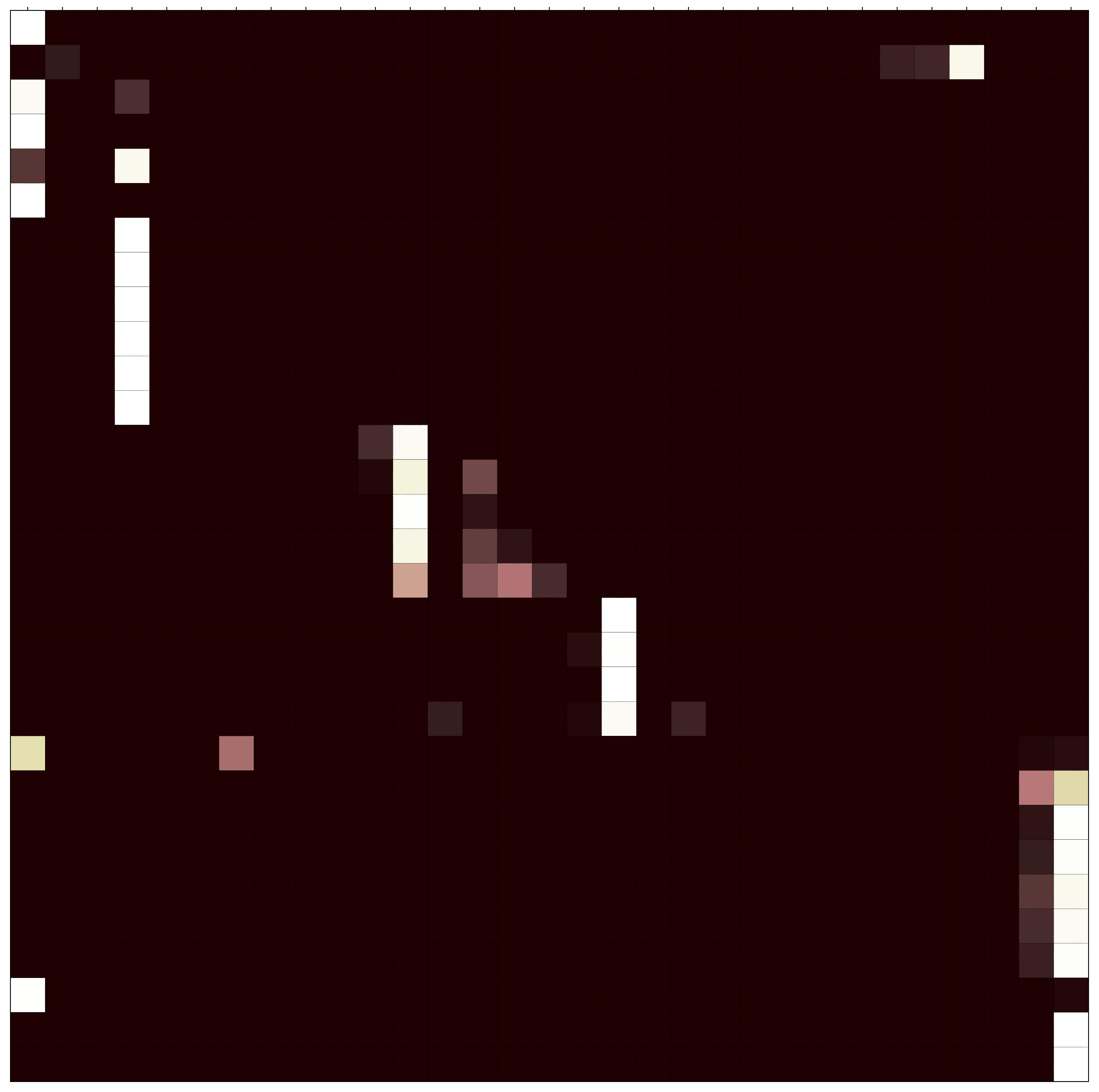}
\includegraphics[width=0.15\textwidth]{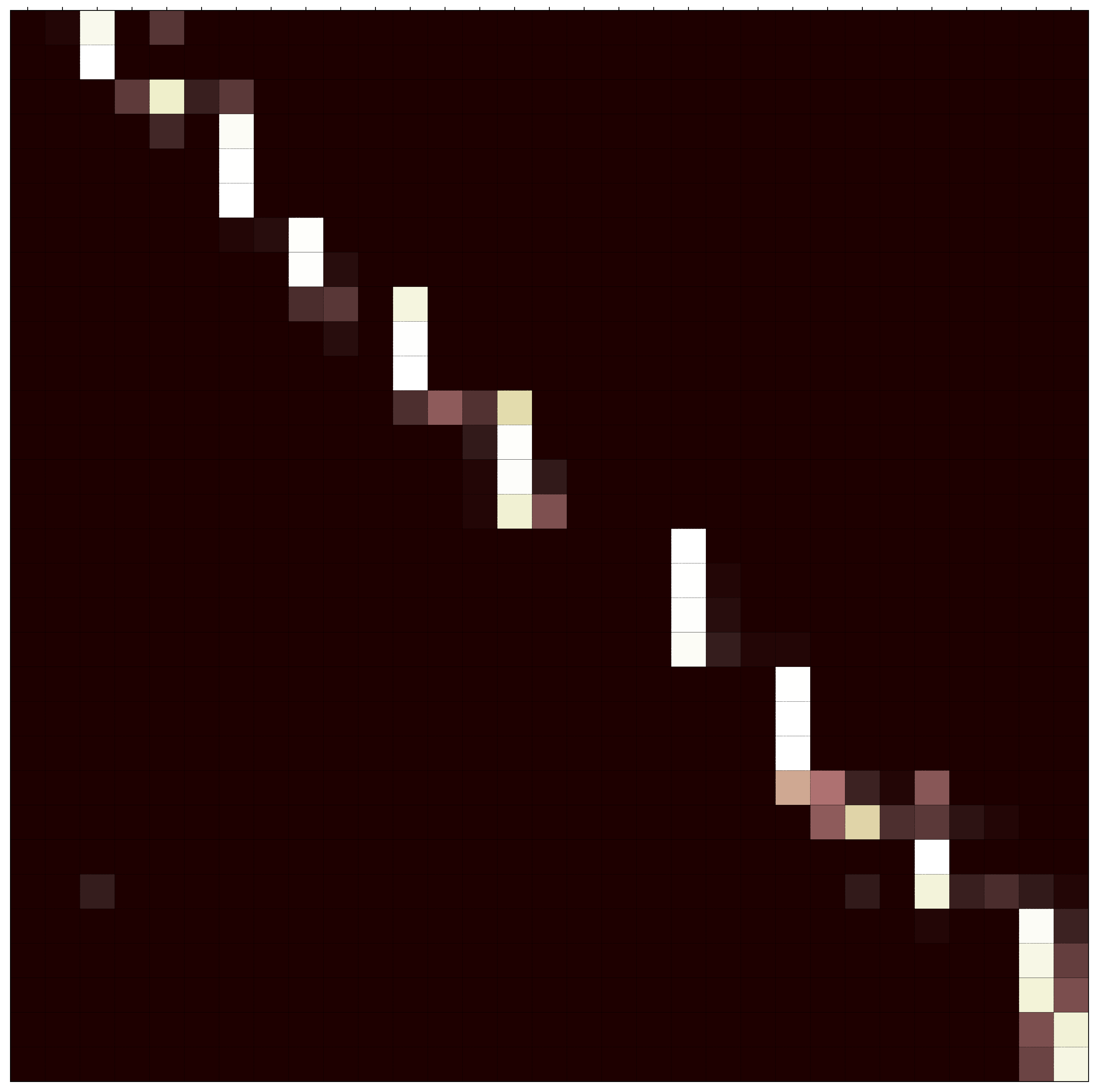}
\includegraphics[width=0.15\textwidth]{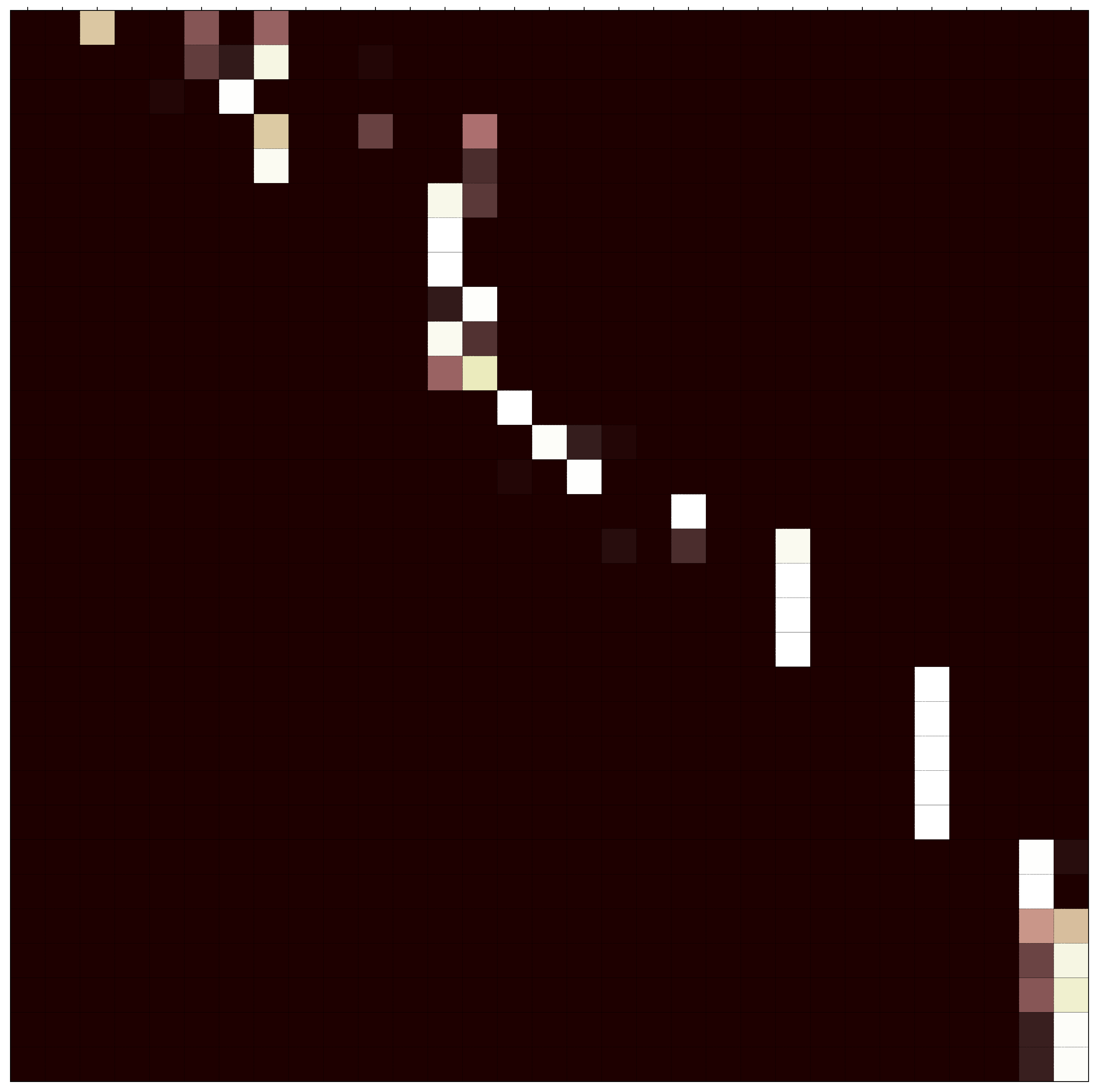}
\includegraphics[width=0.15\textwidth]{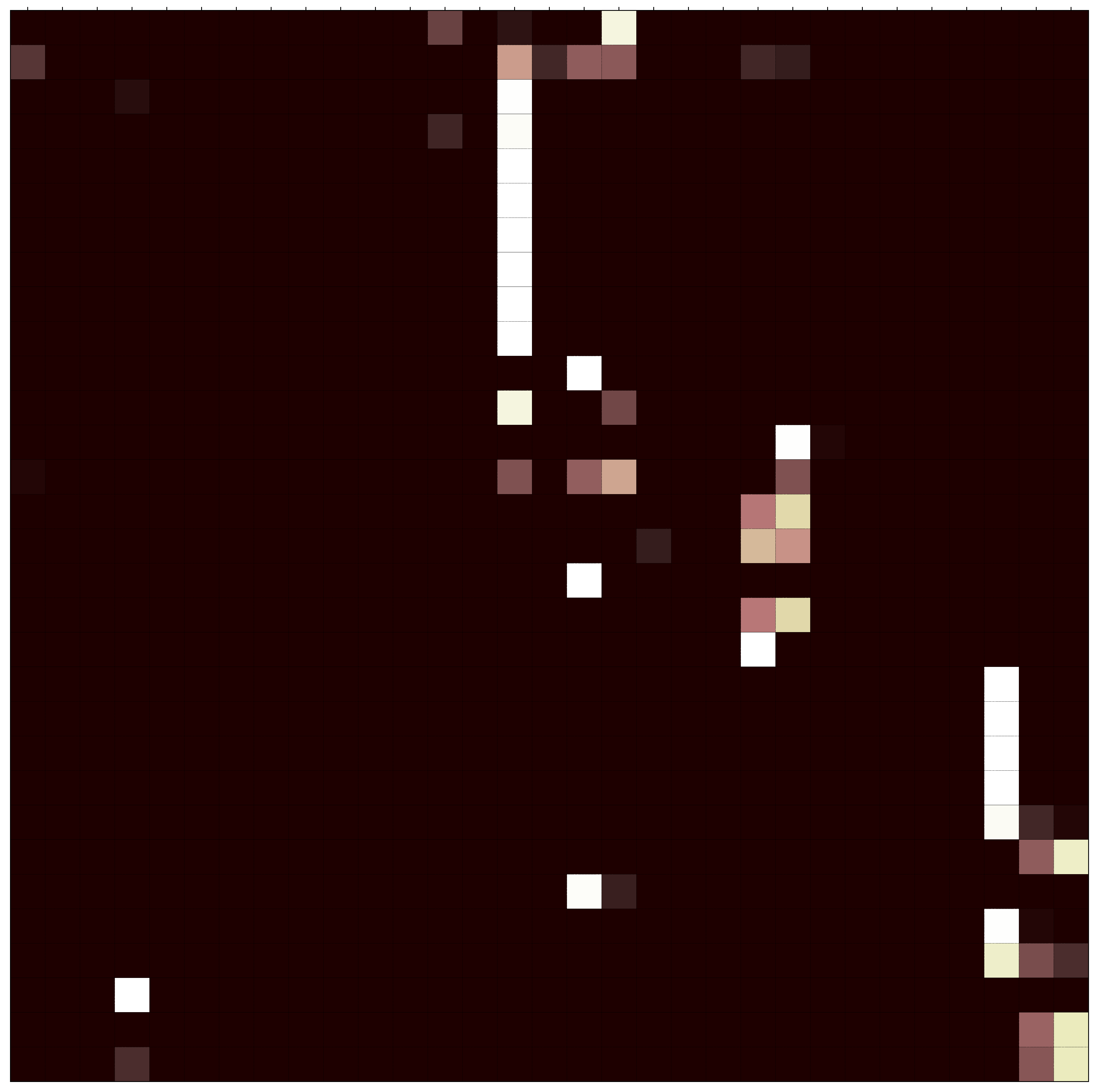}
\includegraphics[width=0.15\textwidth]{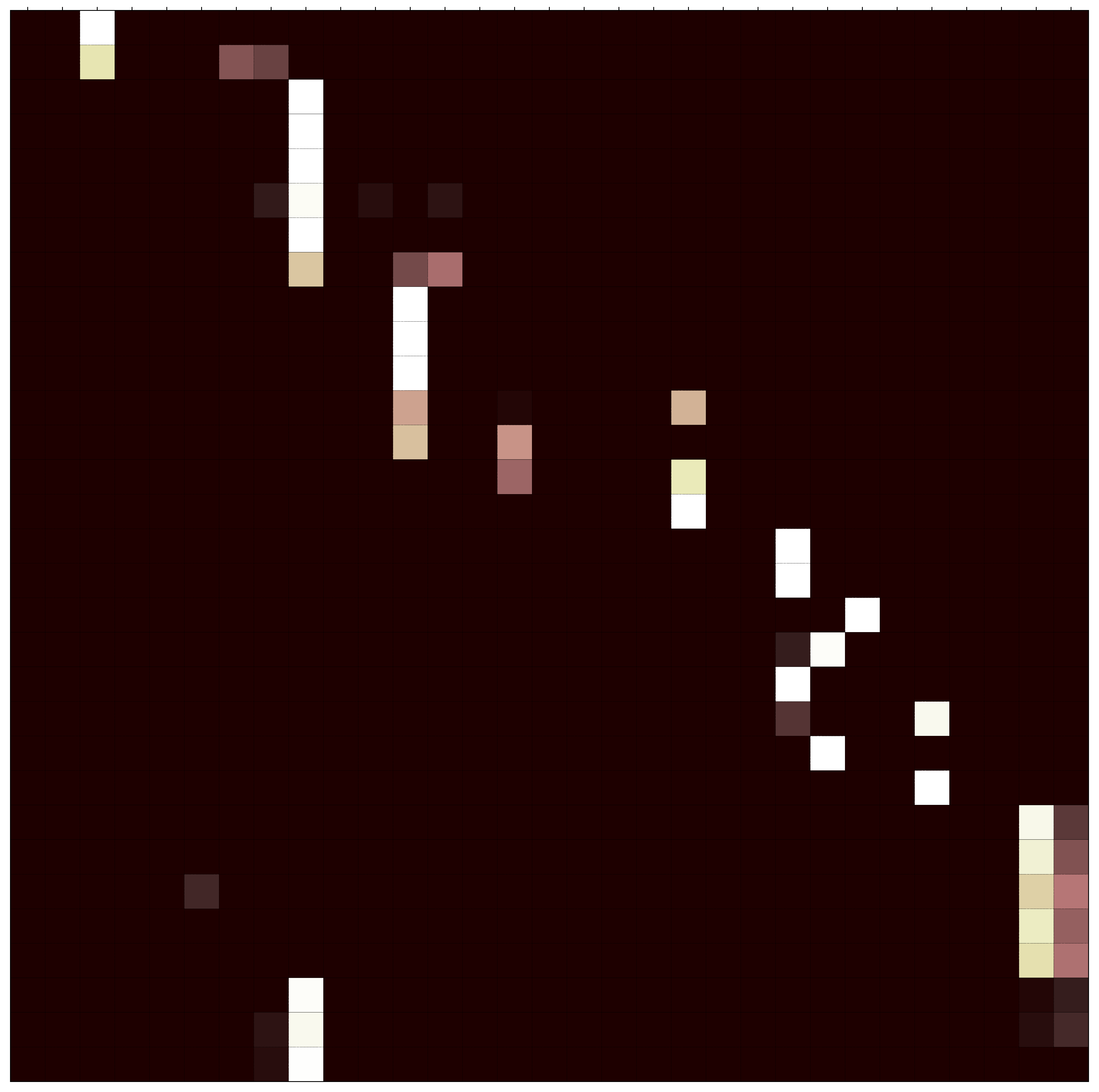}
\includegraphics[width=0.15\textwidth]{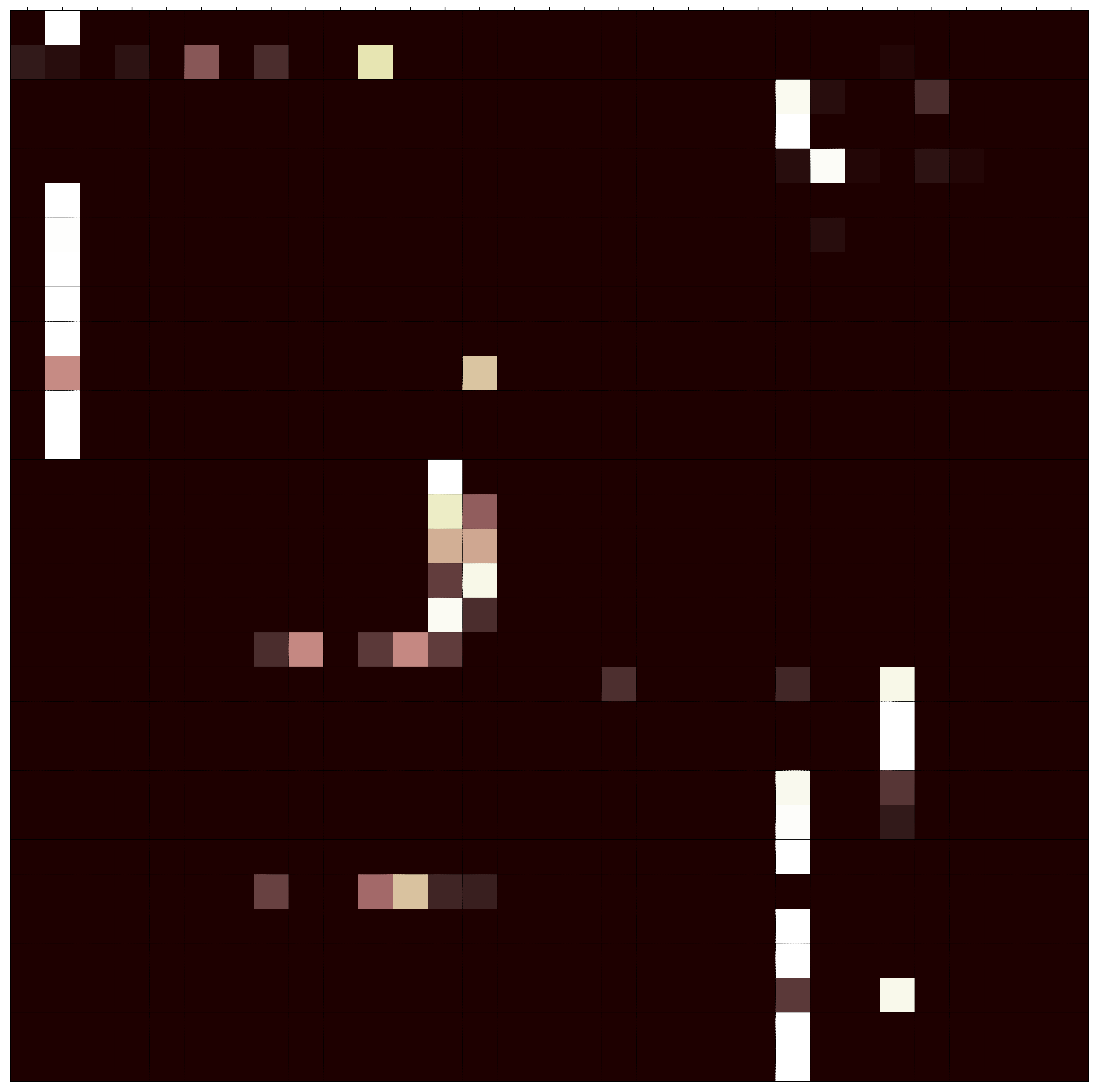}
\includegraphics[width=0.15\textwidth]{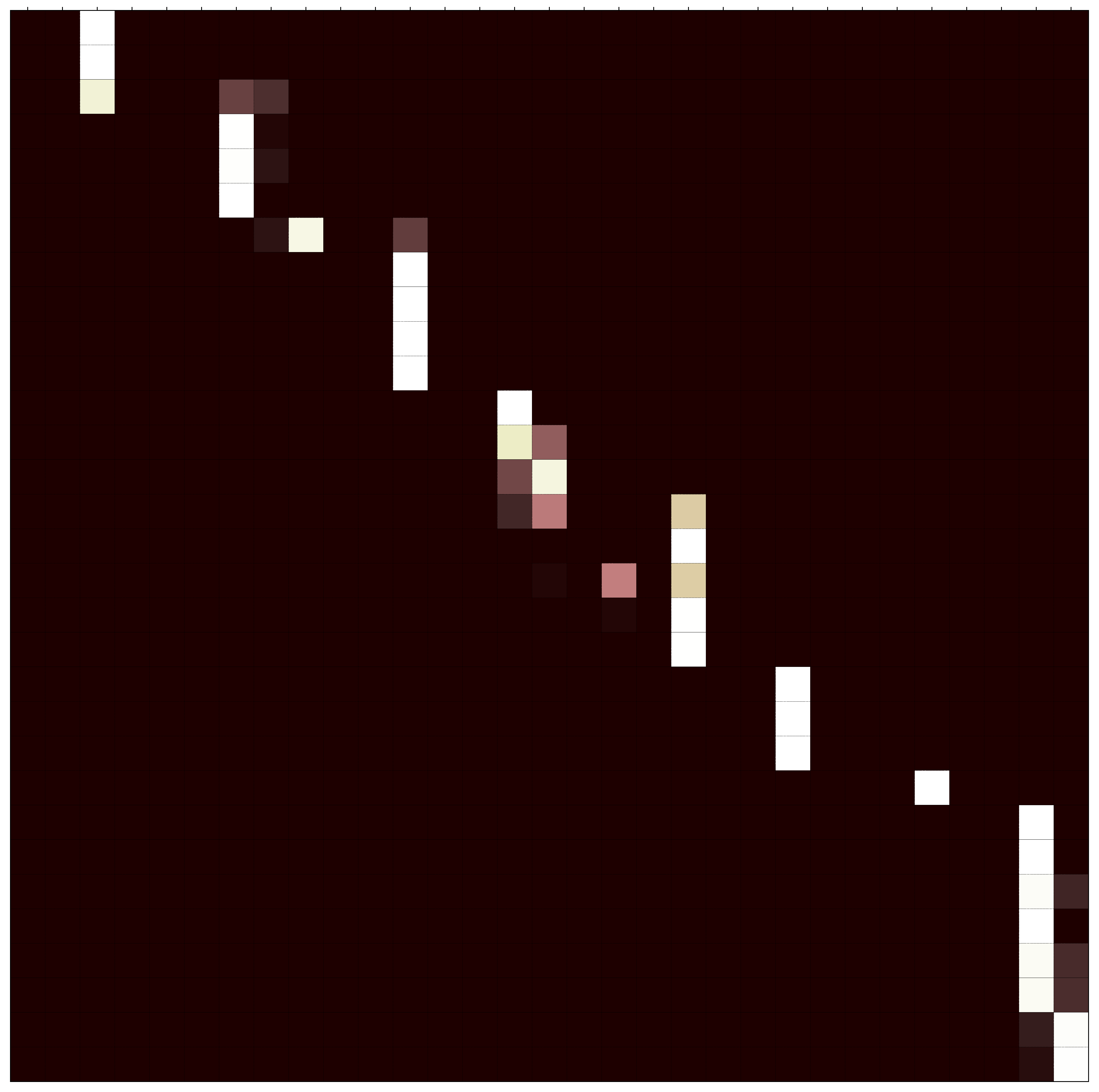}
\includegraphics[width=0.15\textwidth]{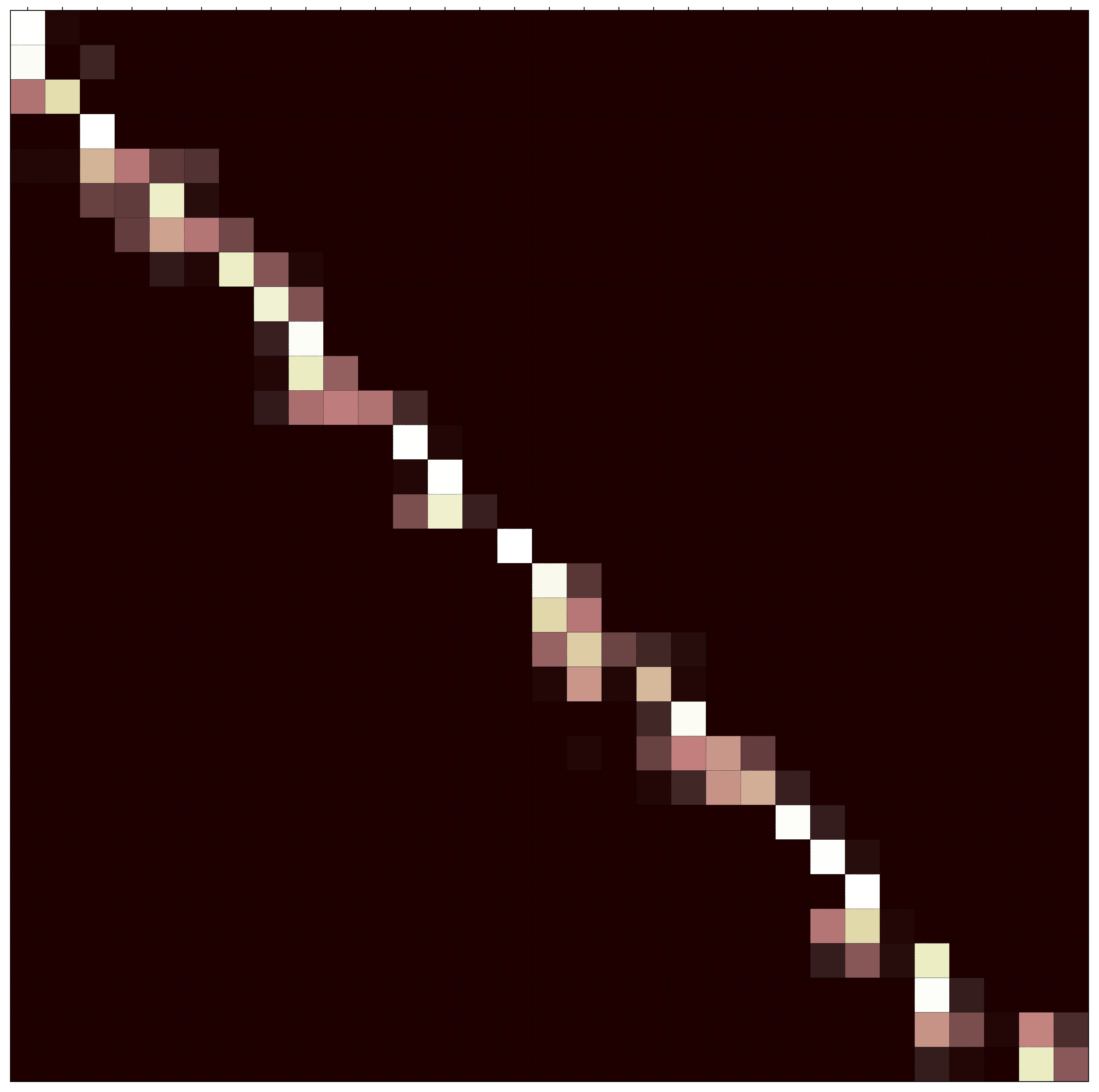}
\includegraphics[width=0.15\textwidth]{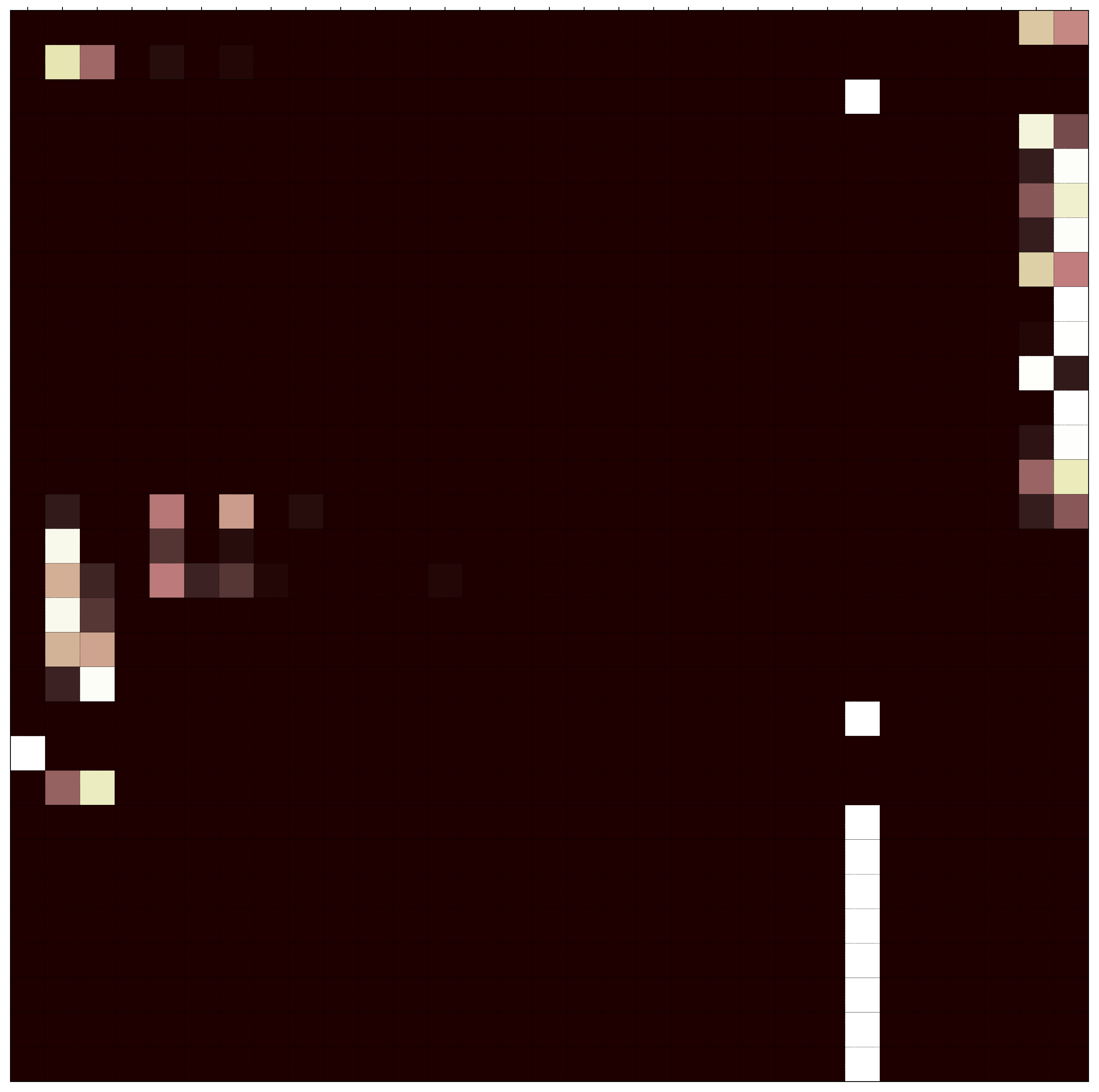}
\includegraphics[width=0.15\textwidth]{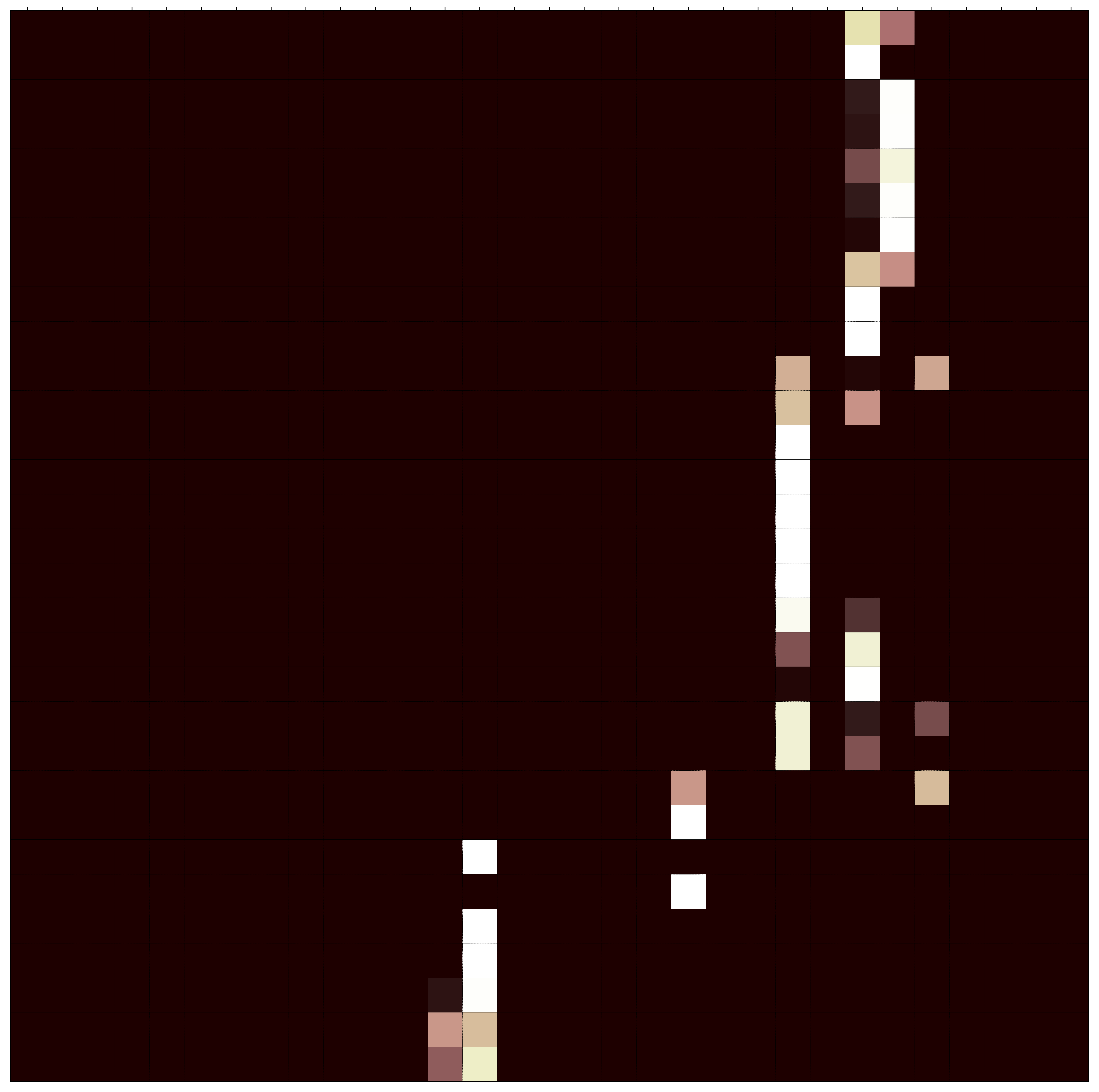}
\end{center}
\caption{Layer 4}
\end{figure}

\begin{figure}
\begin{center}
\includegraphics[width=0.15\textwidth]{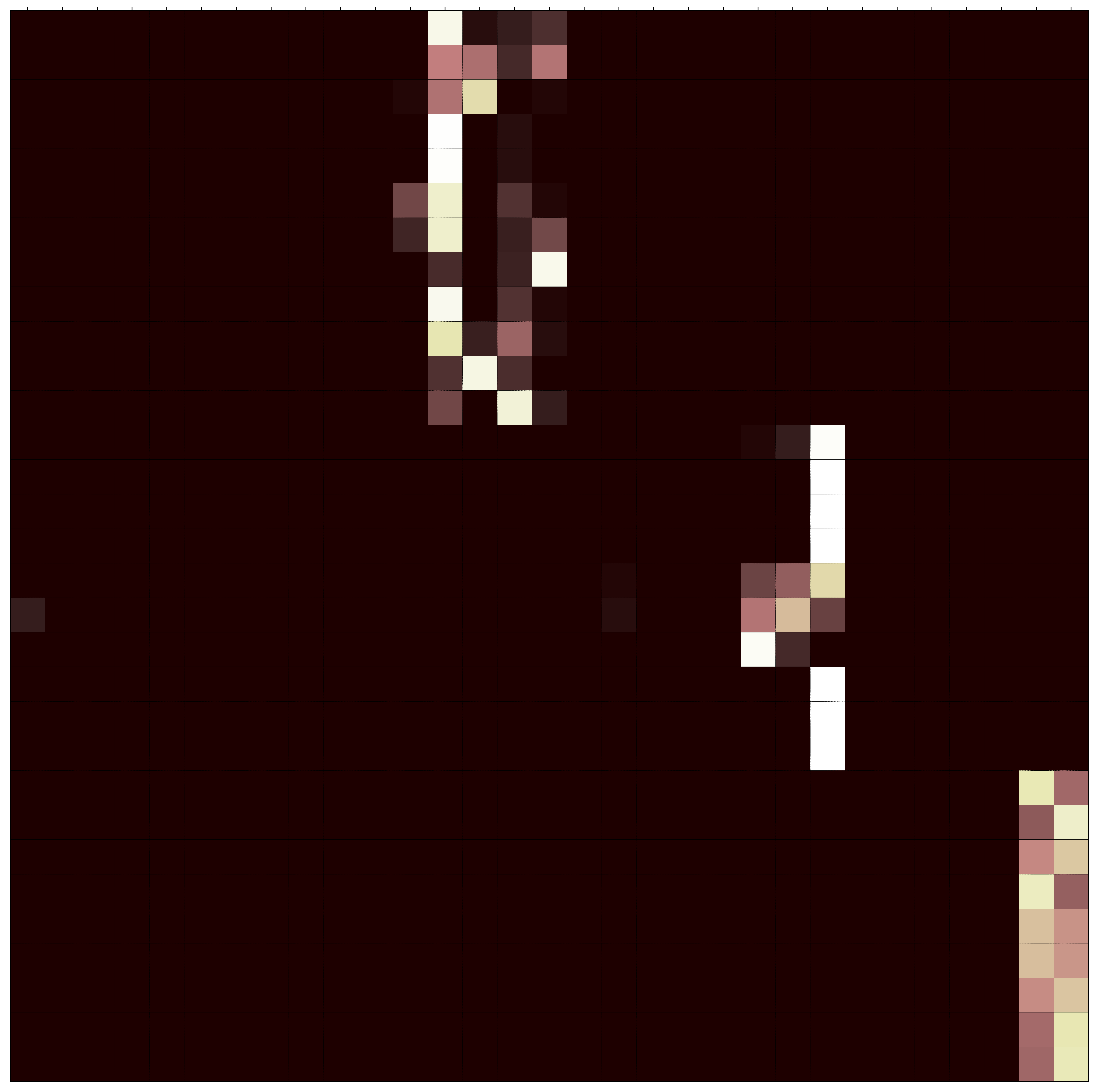}
\includegraphics[width=0.15\textwidth]{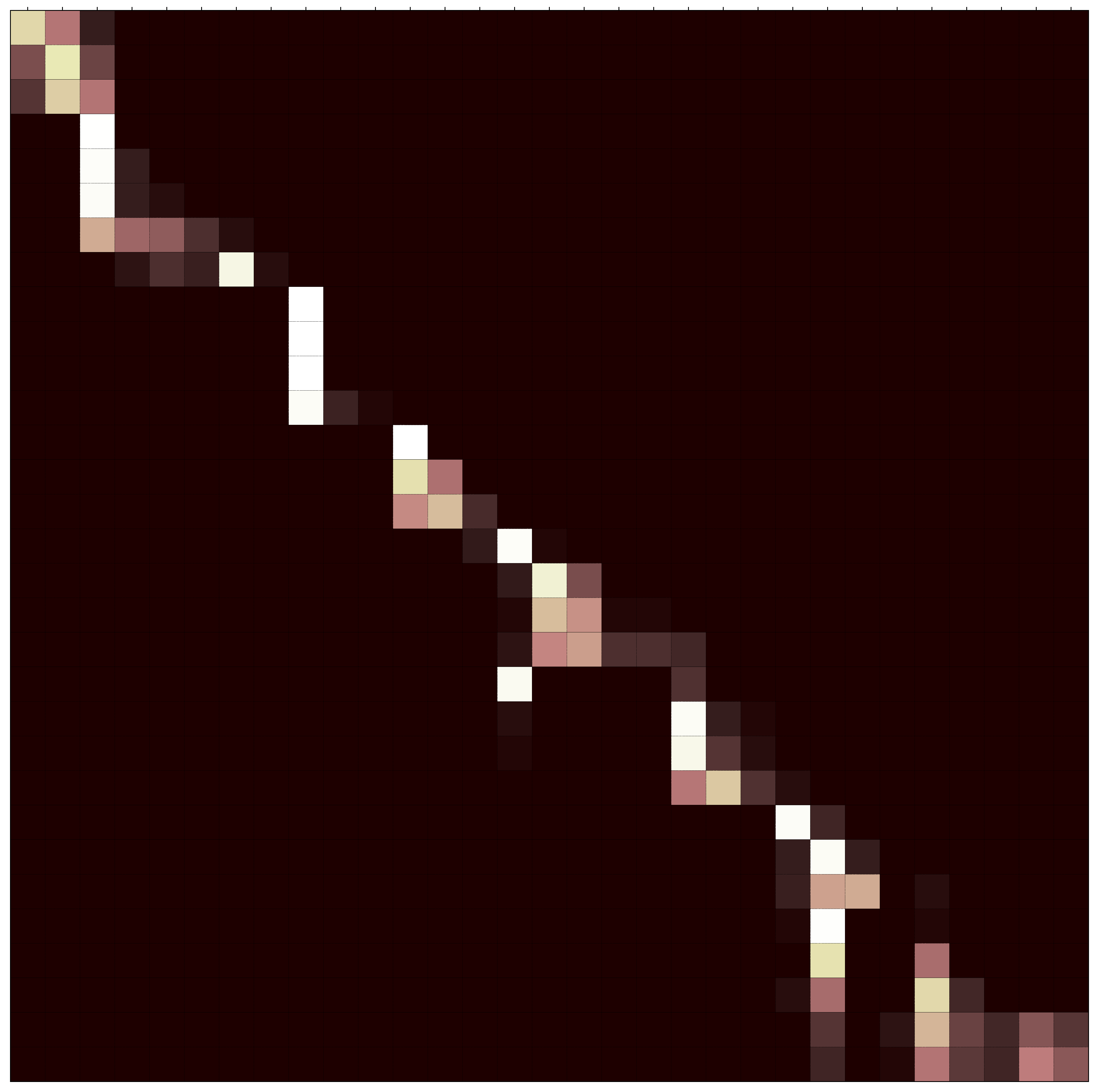}
\includegraphics[width=0.15\textwidth]{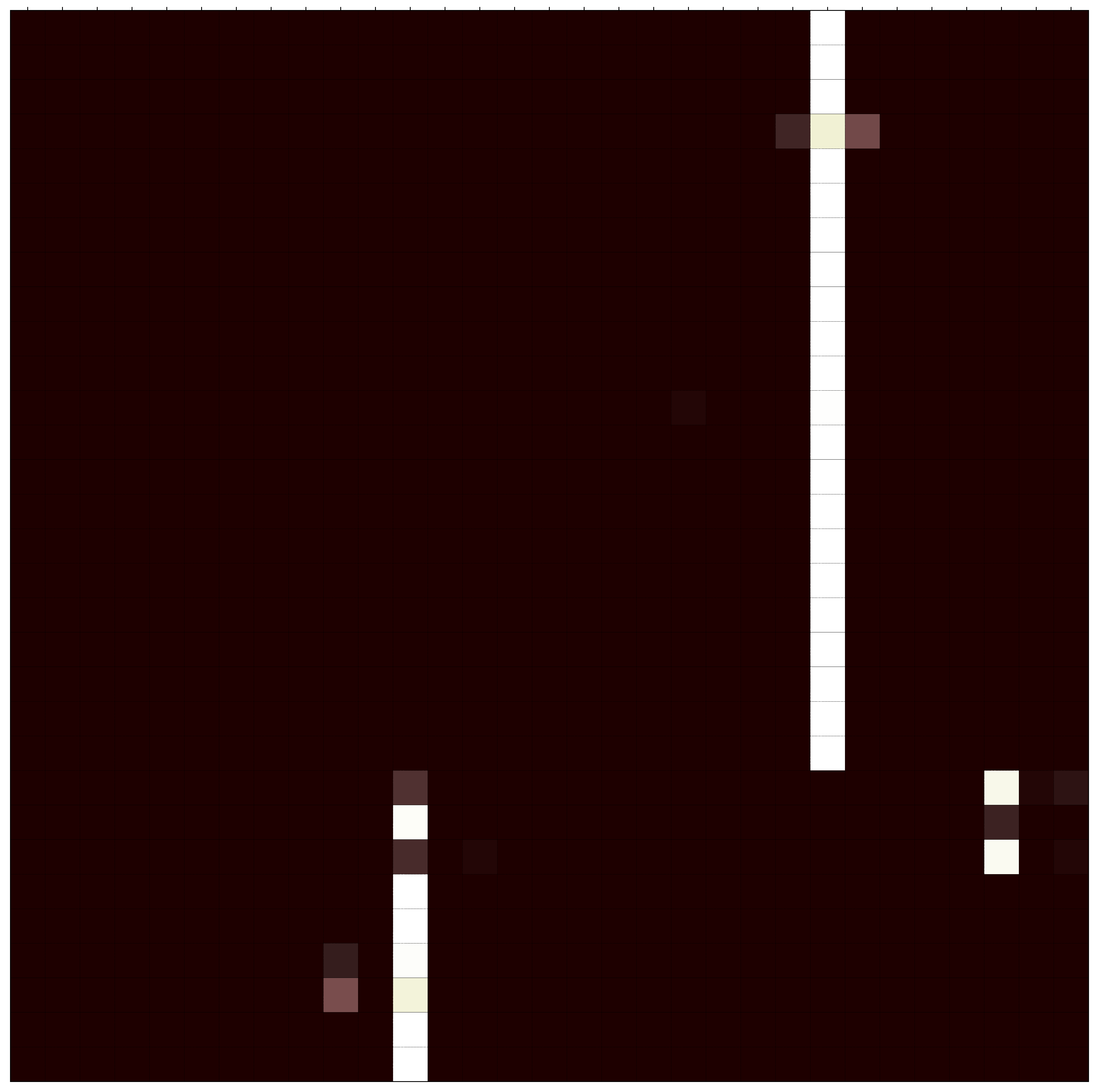}
\includegraphics[width=0.15\textwidth]{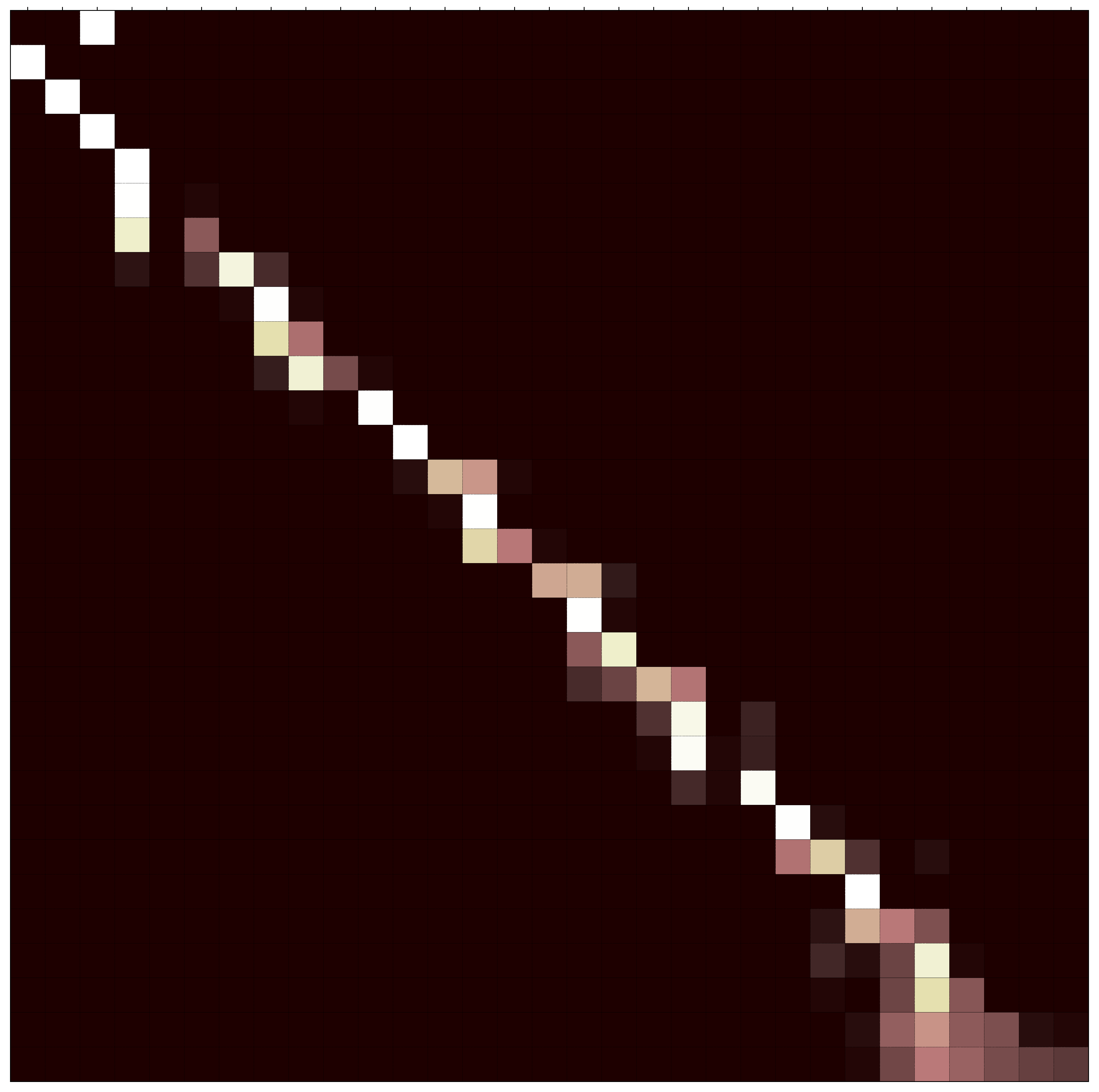}
\includegraphics[width=0.15\textwidth]{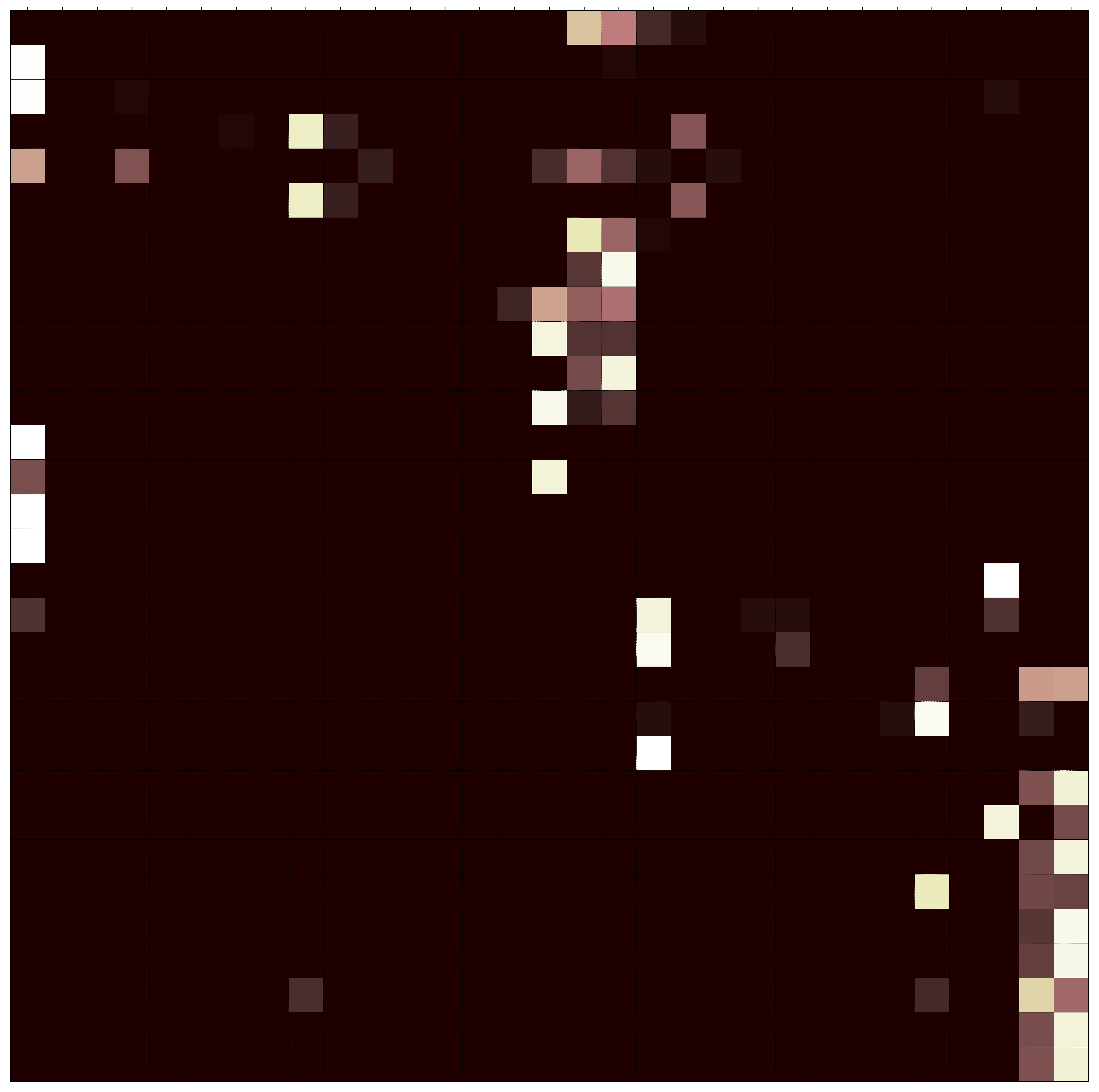}
\includegraphics[width=0.15\textwidth]{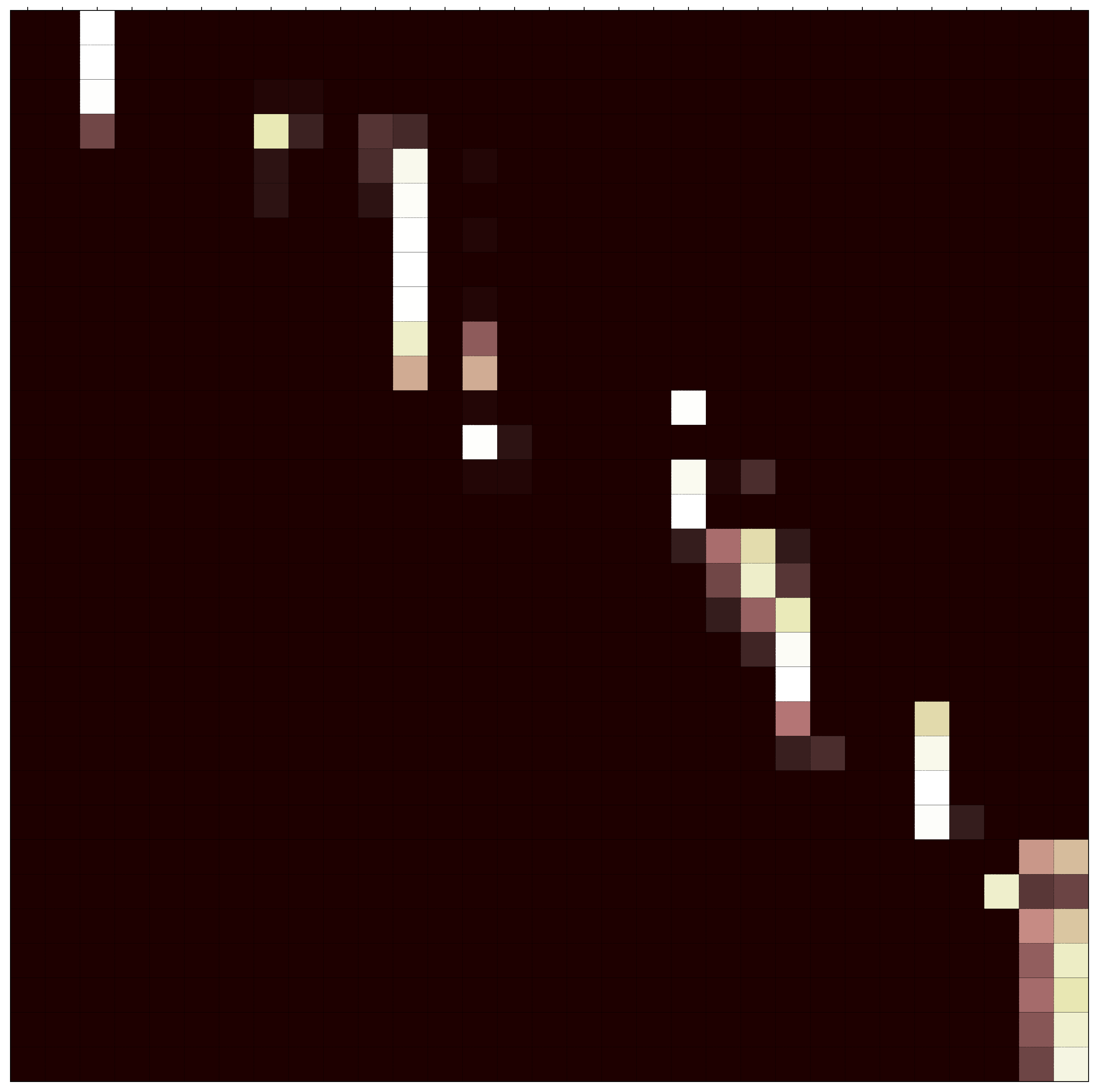}
\includegraphics[width=0.15\textwidth]{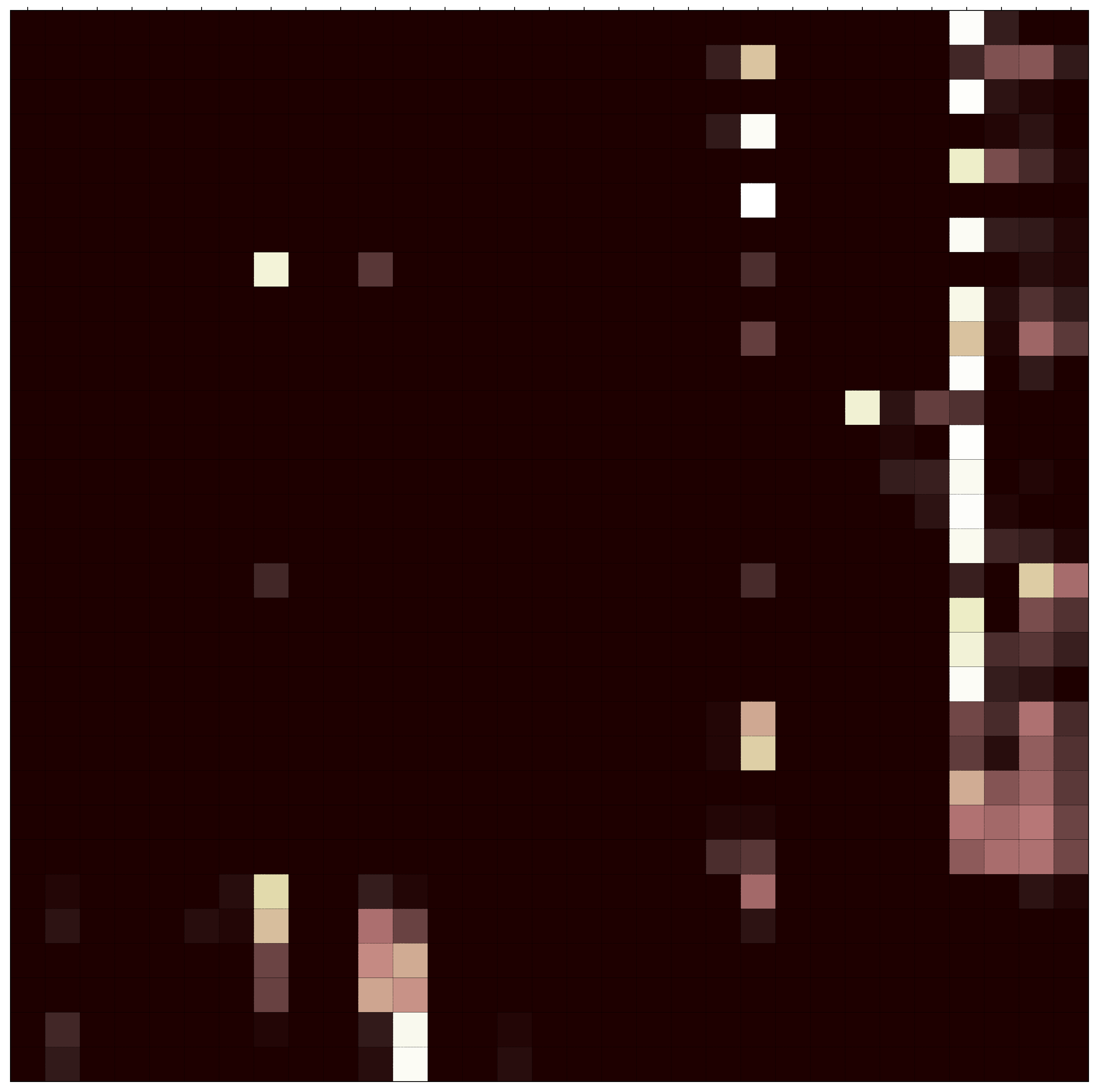}
\includegraphics[width=0.15\textwidth]{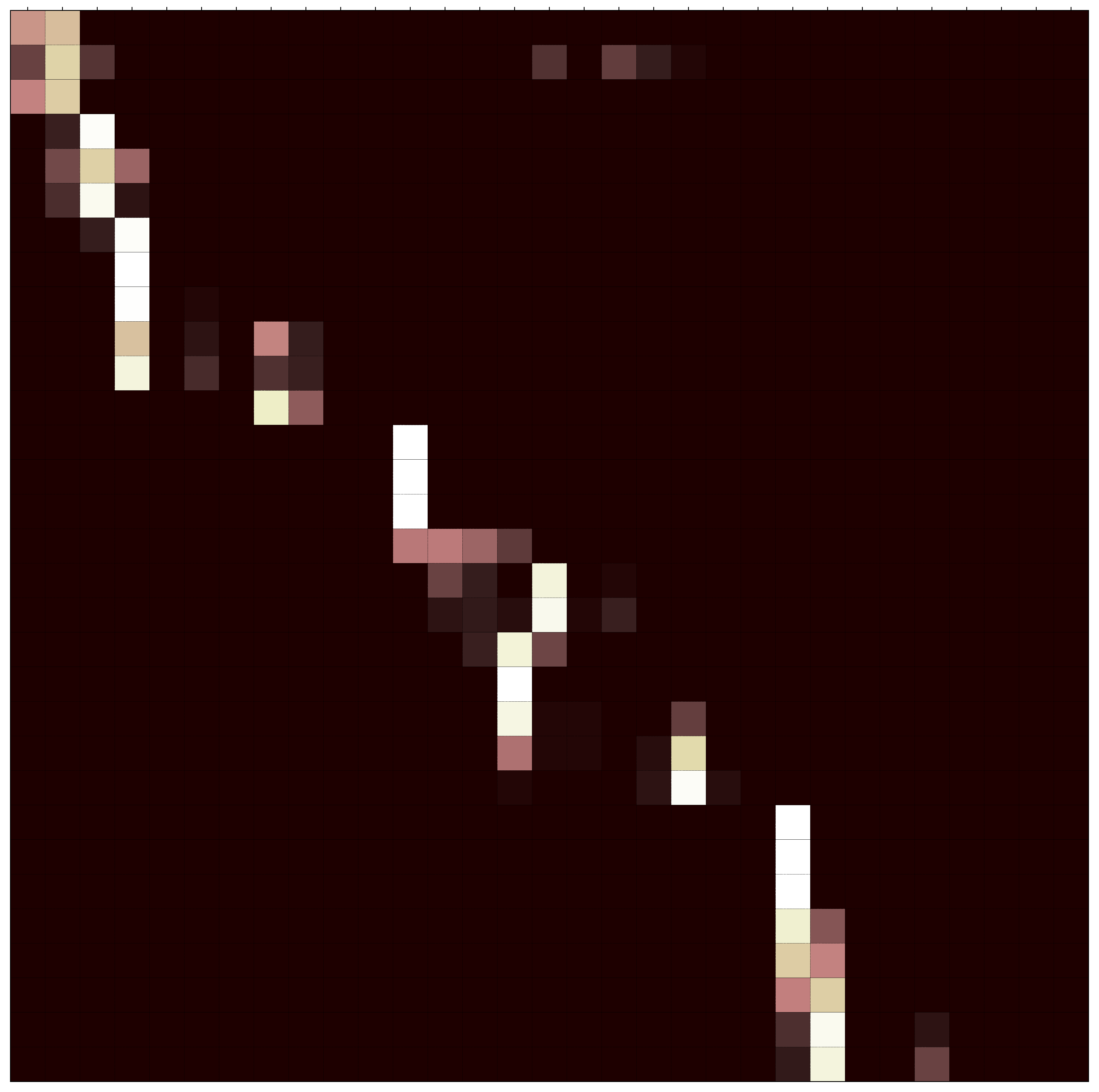}
\includegraphics[width=0.15\textwidth]{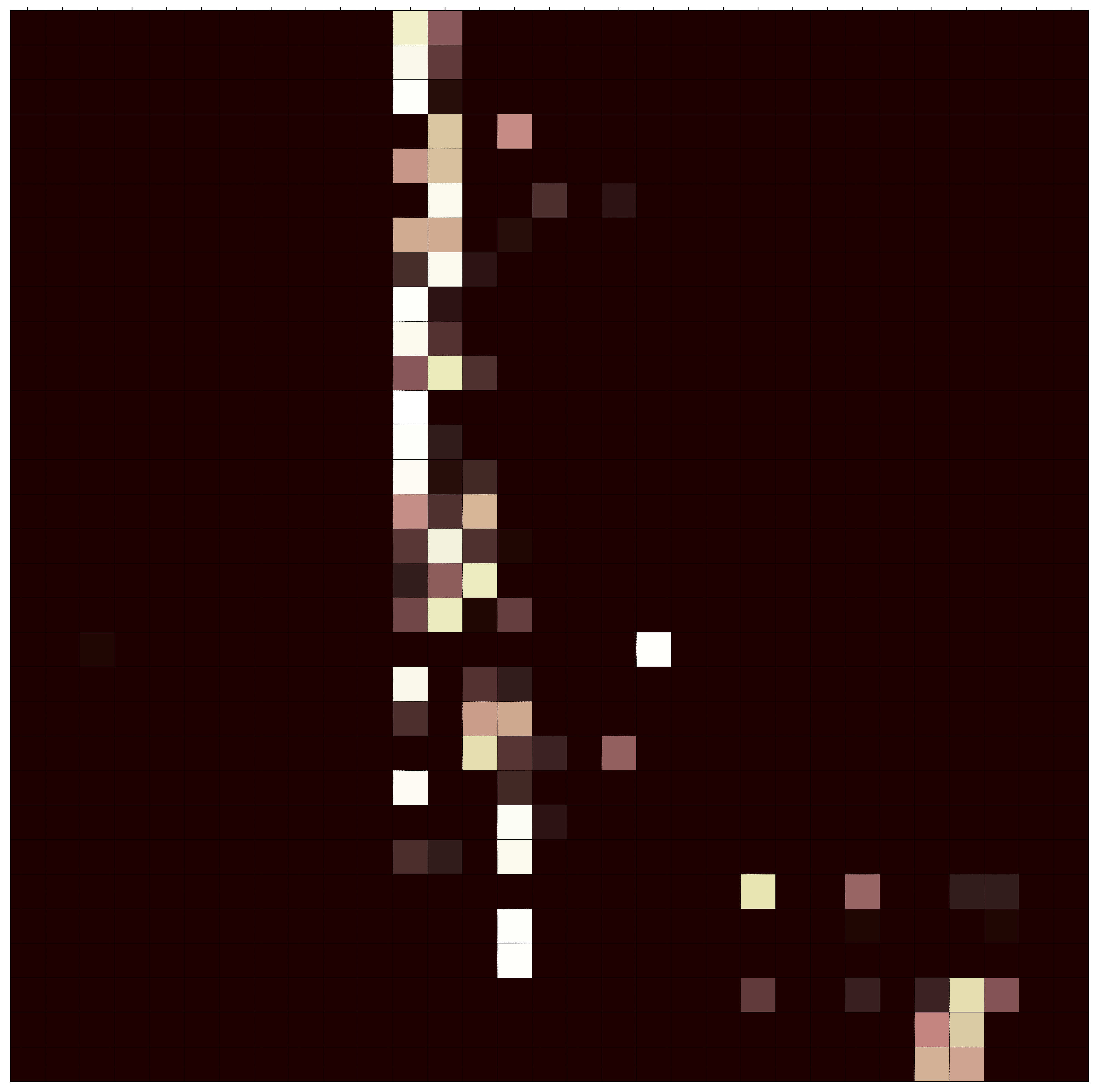}
\includegraphics[width=0.15\textwidth]{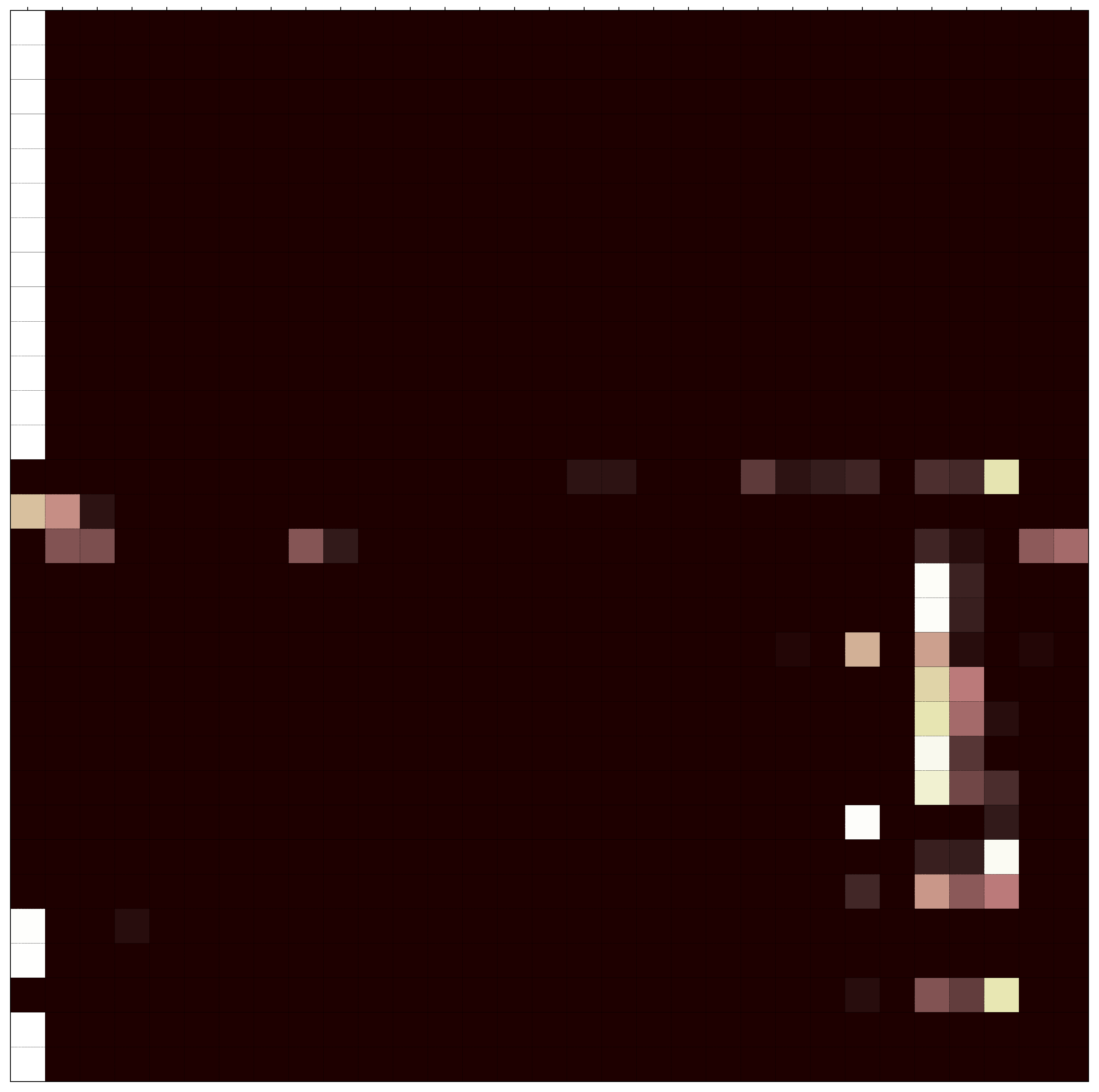}
\includegraphics[width=0.15\textwidth]{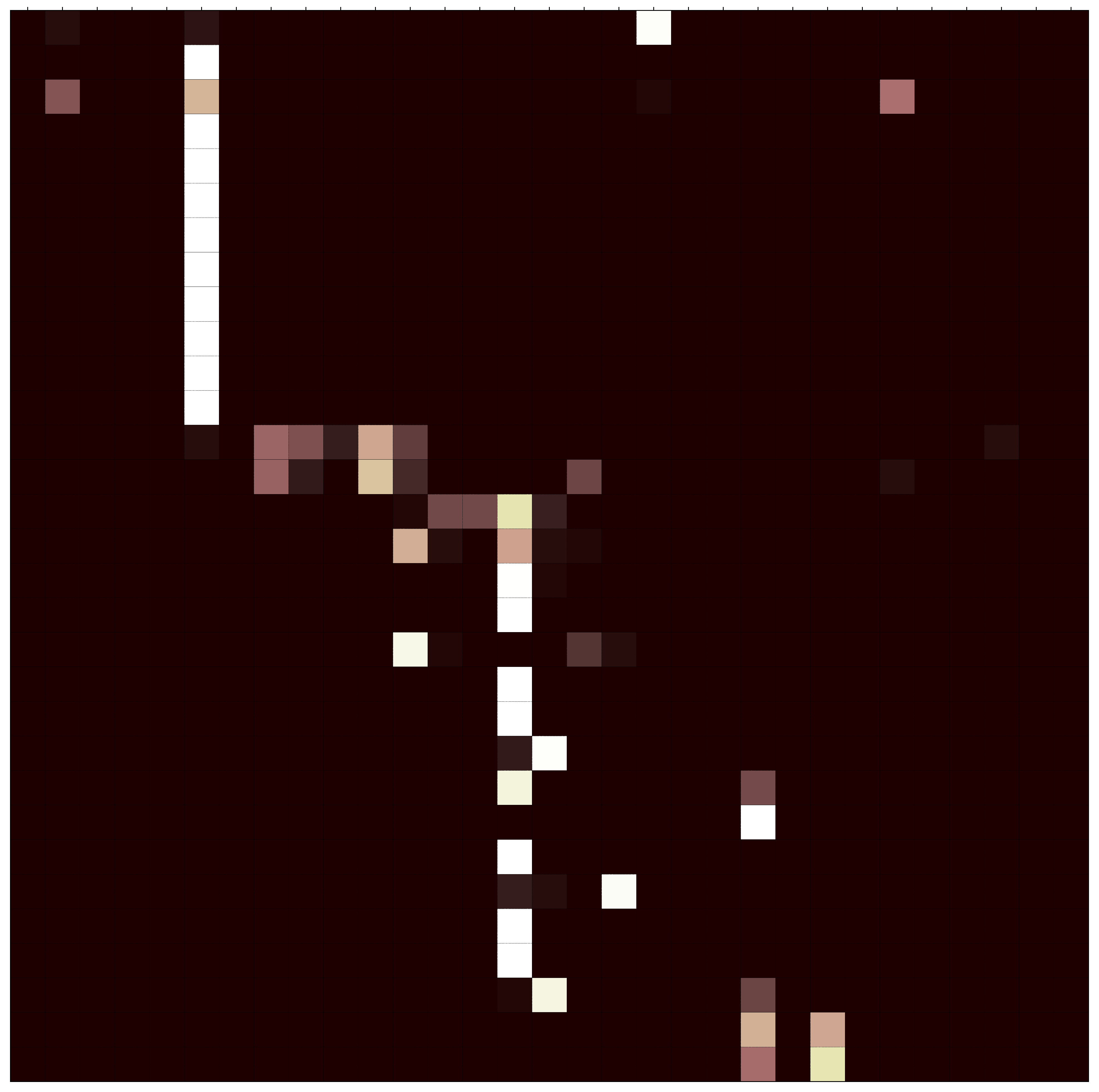}
\includegraphics[width=0.15\textwidth]{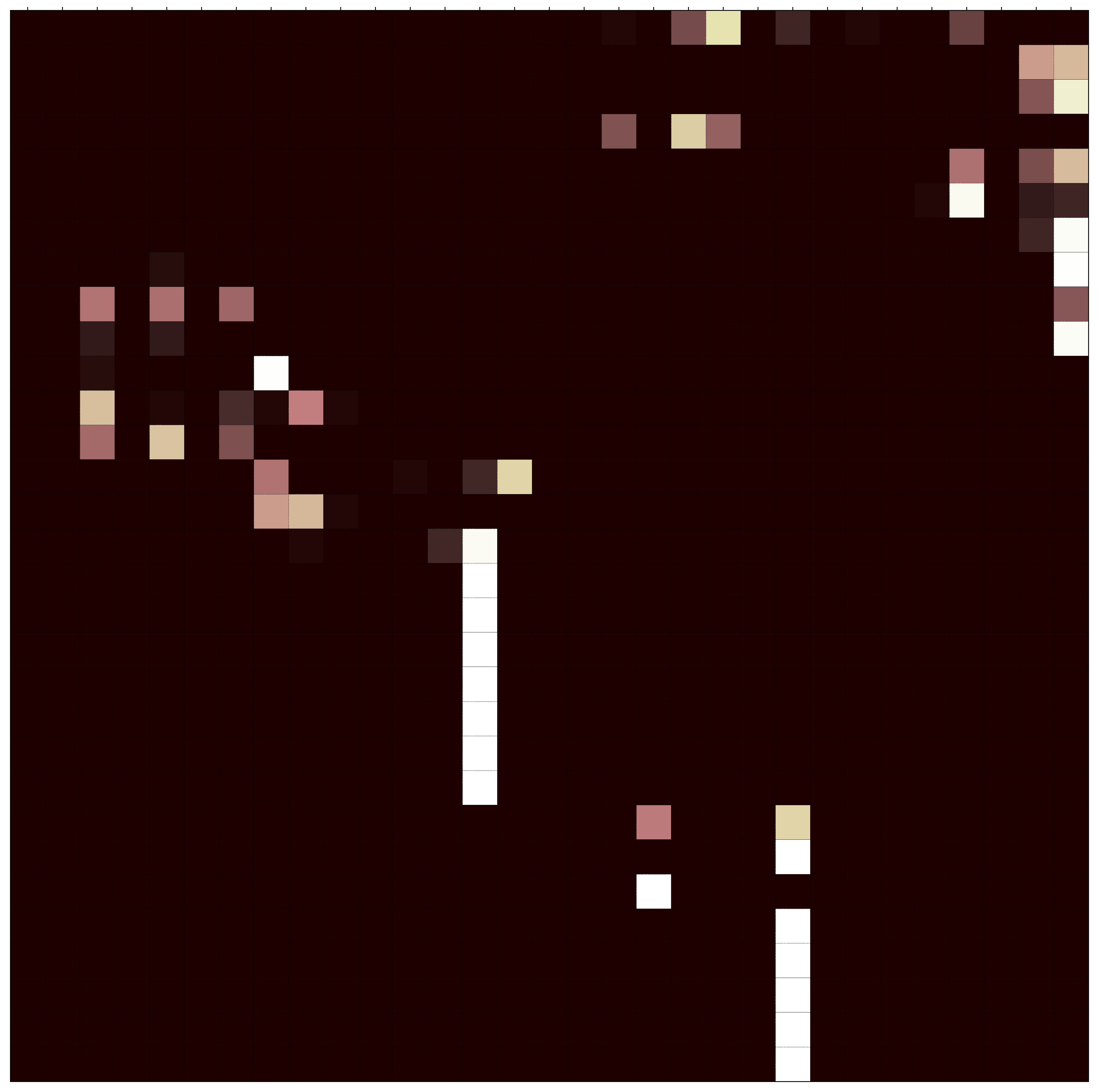}
\includegraphics[width=0.15\textwidth]{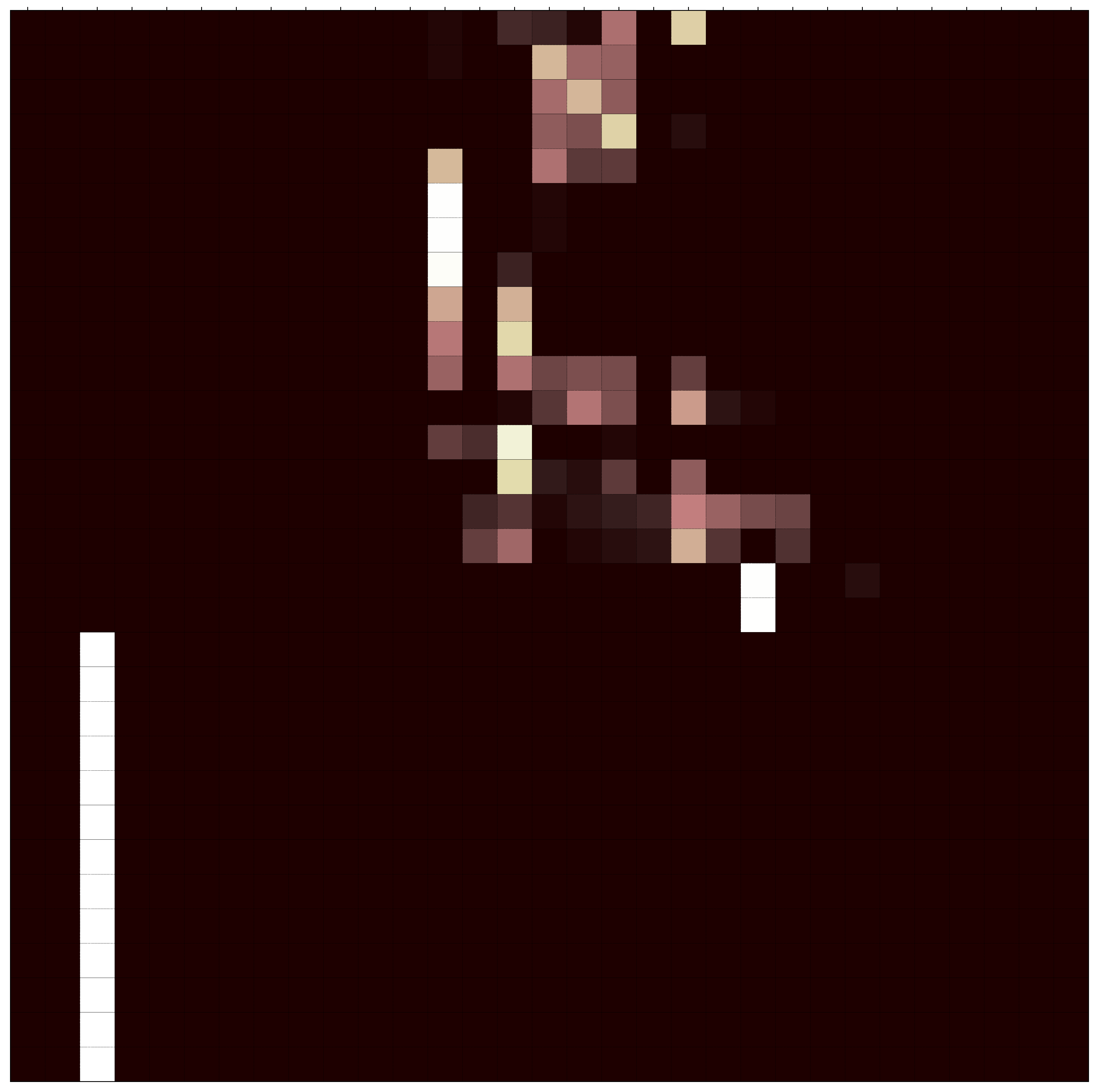}
\includegraphics[width=0.15\textwidth]{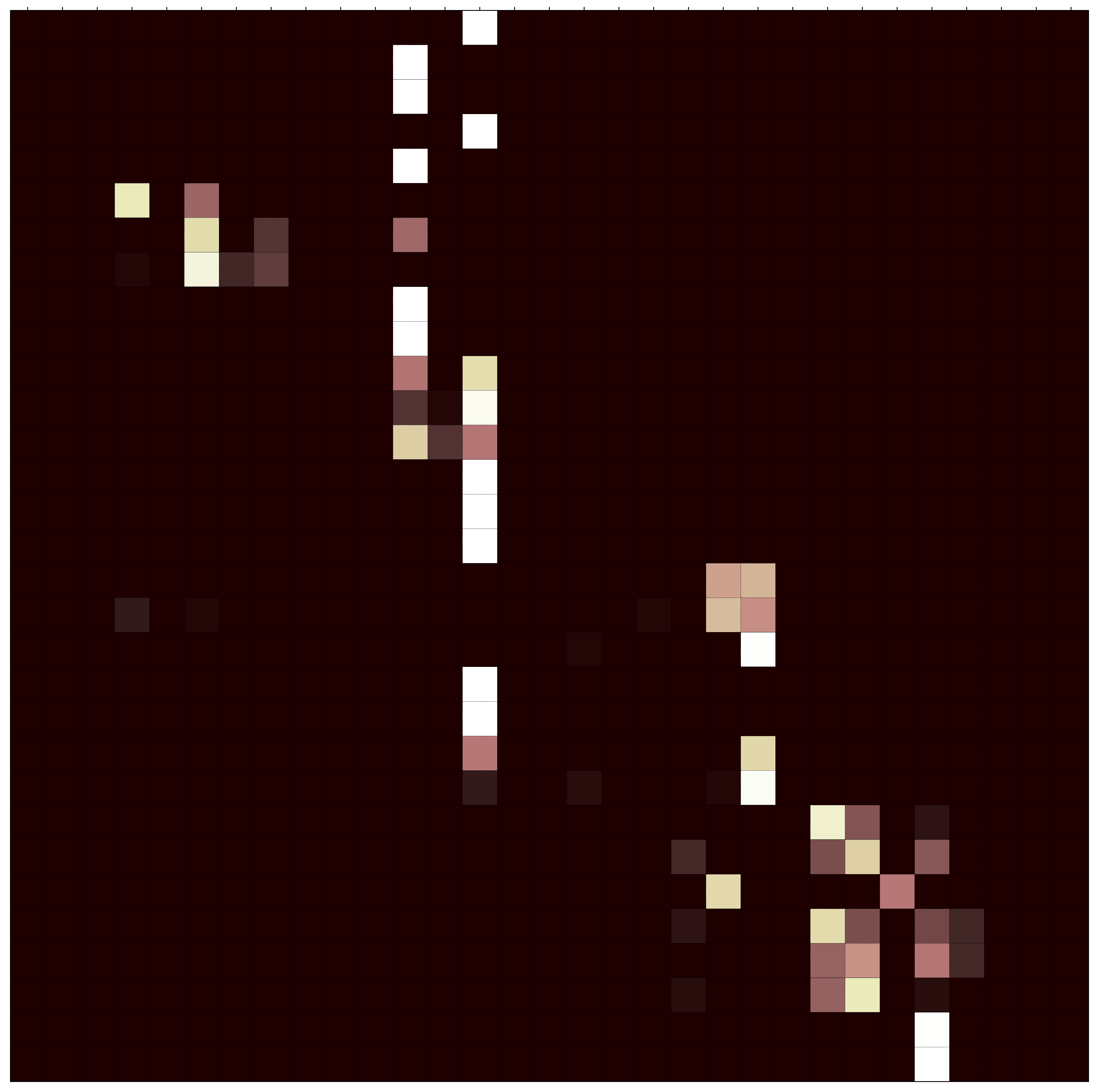}
\includegraphics[width=0.15\textwidth]{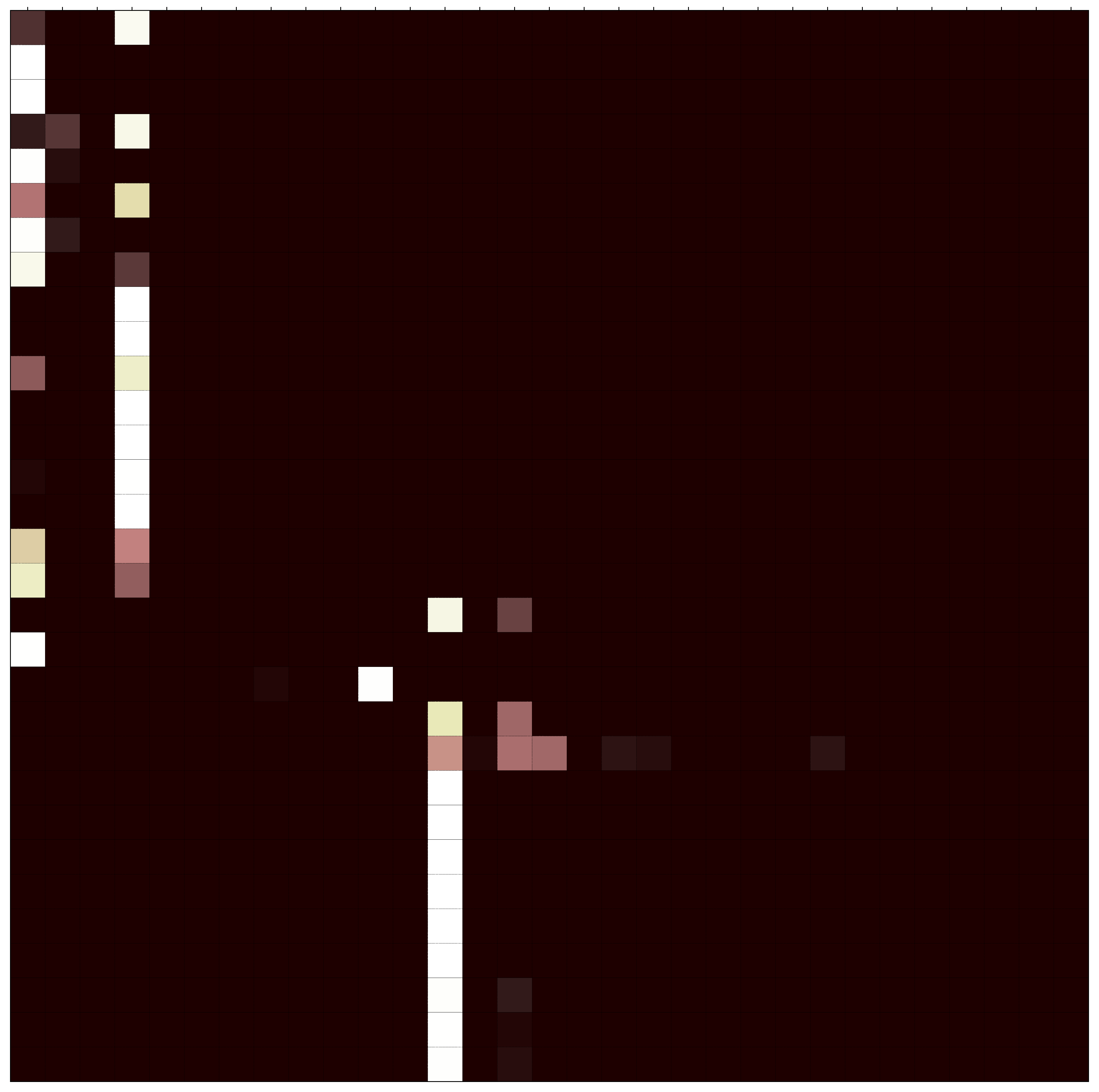}
\includegraphics[width=0.15\textwidth]{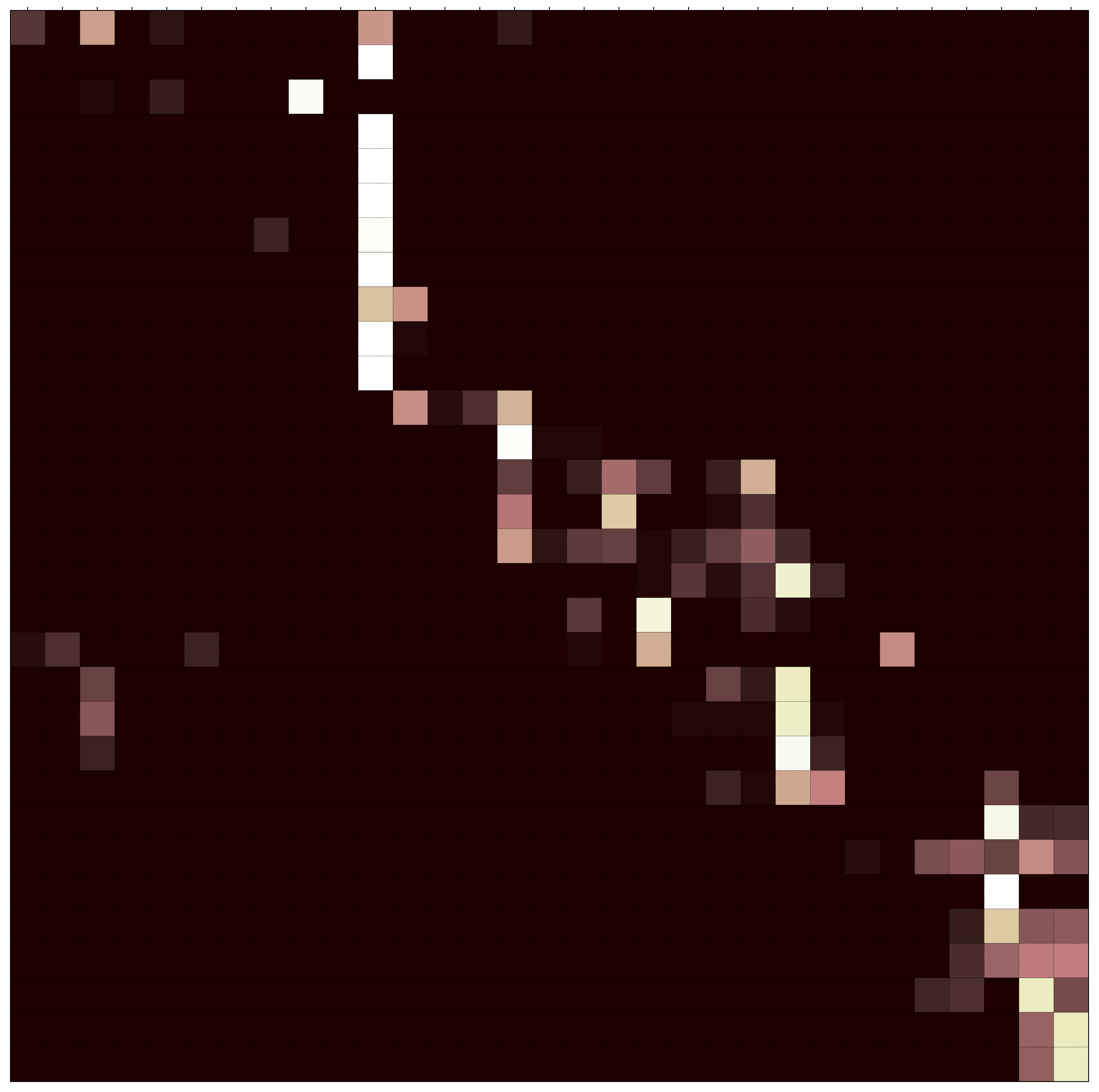}
\end{center}
\caption{Layer 5}
\end{figure}

\begin{figure}
\begin{center}
\includegraphics[width=0.15\textwidth]{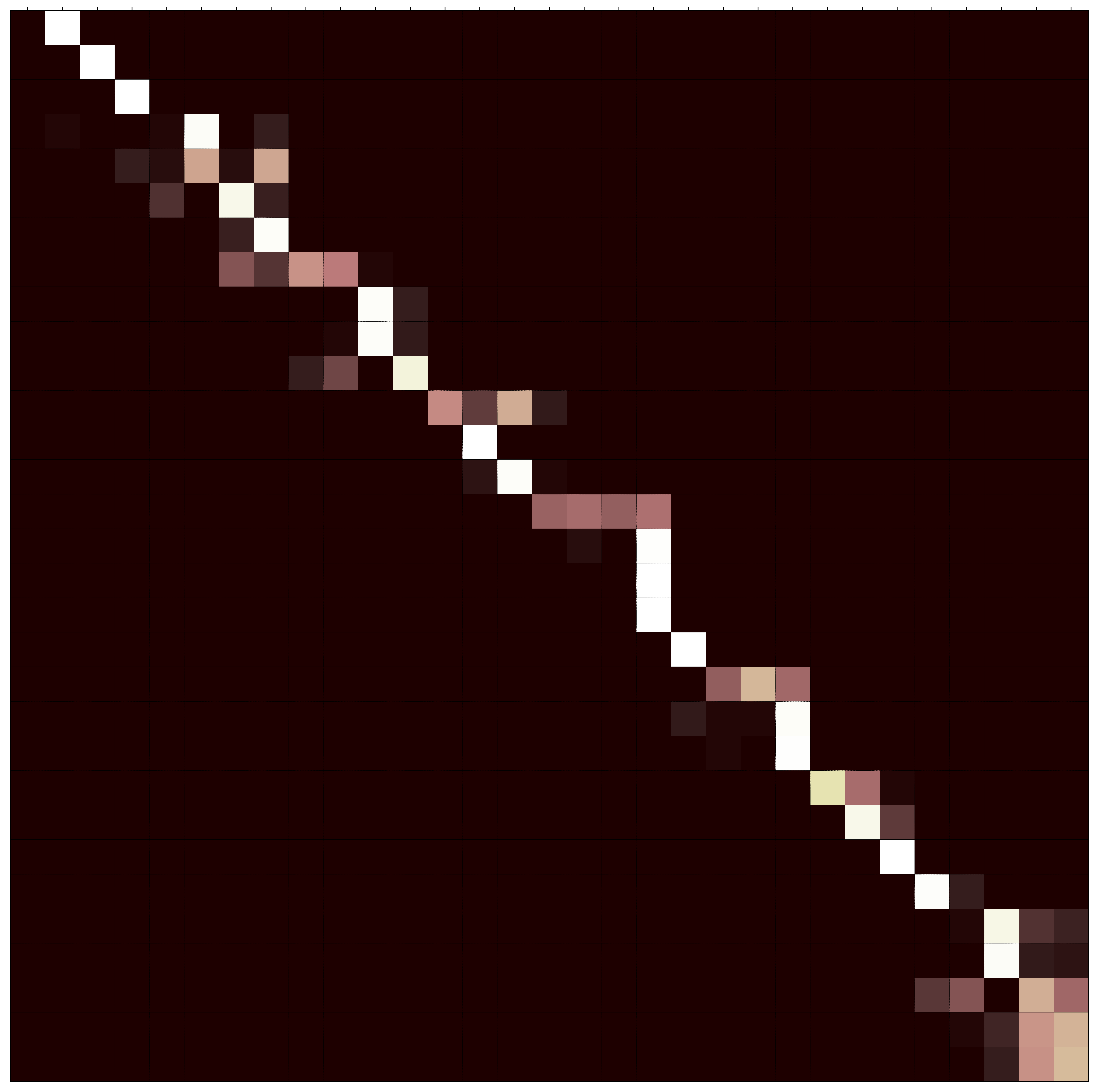}
\includegraphics[width=0.15\textwidth]{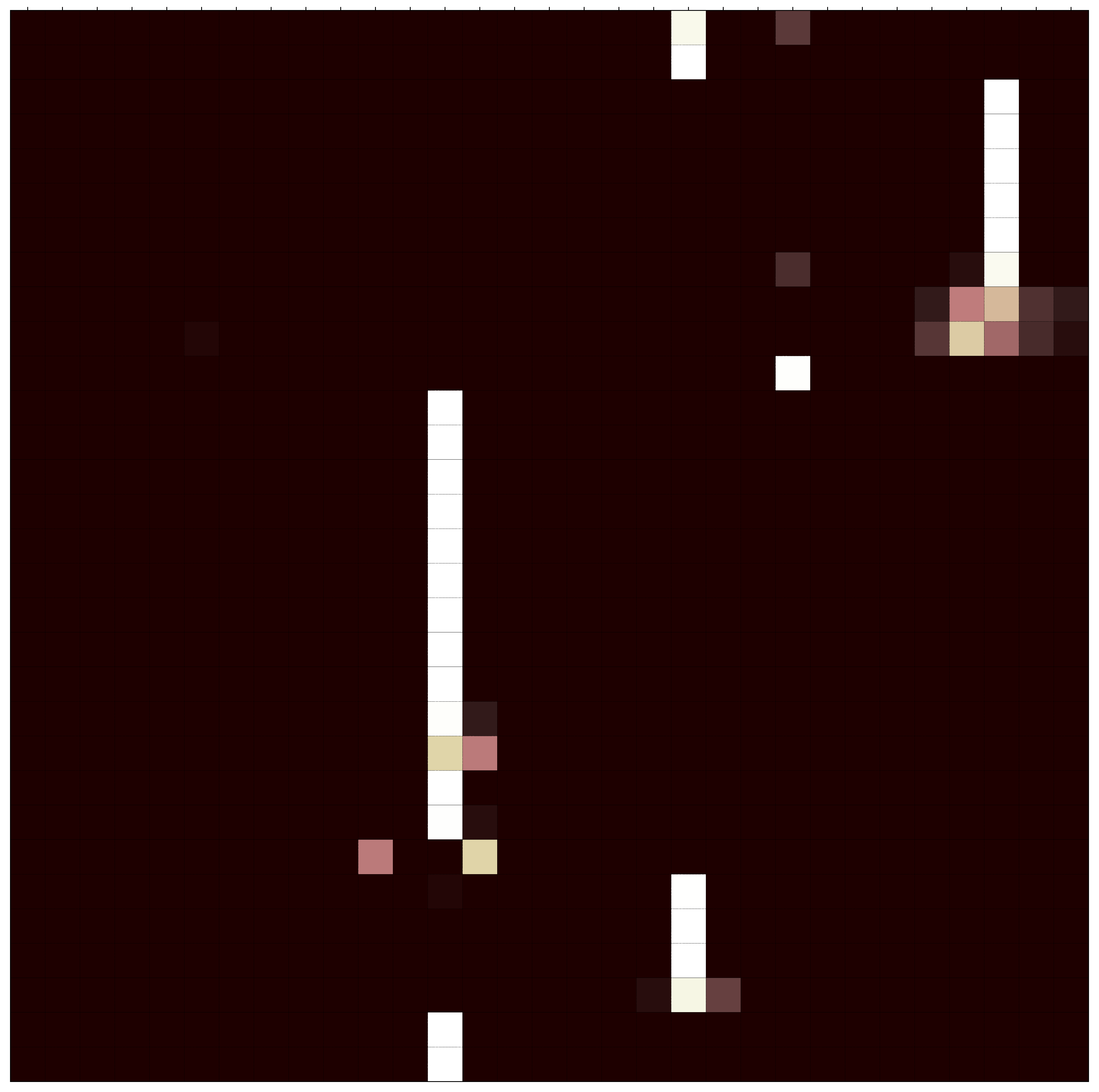}
\includegraphics[width=0.15\textwidth]{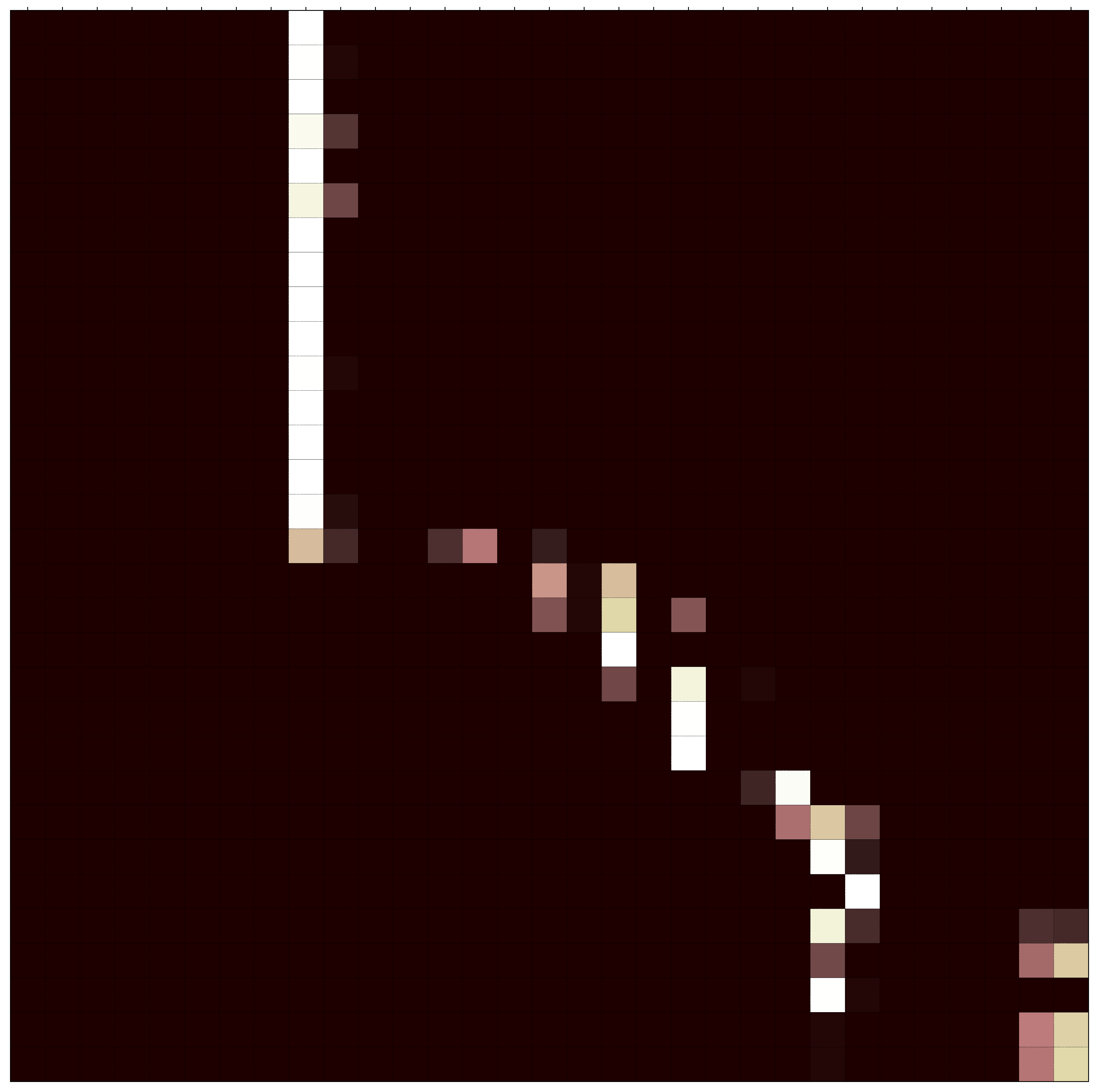}
\includegraphics[width=0.15\textwidth]{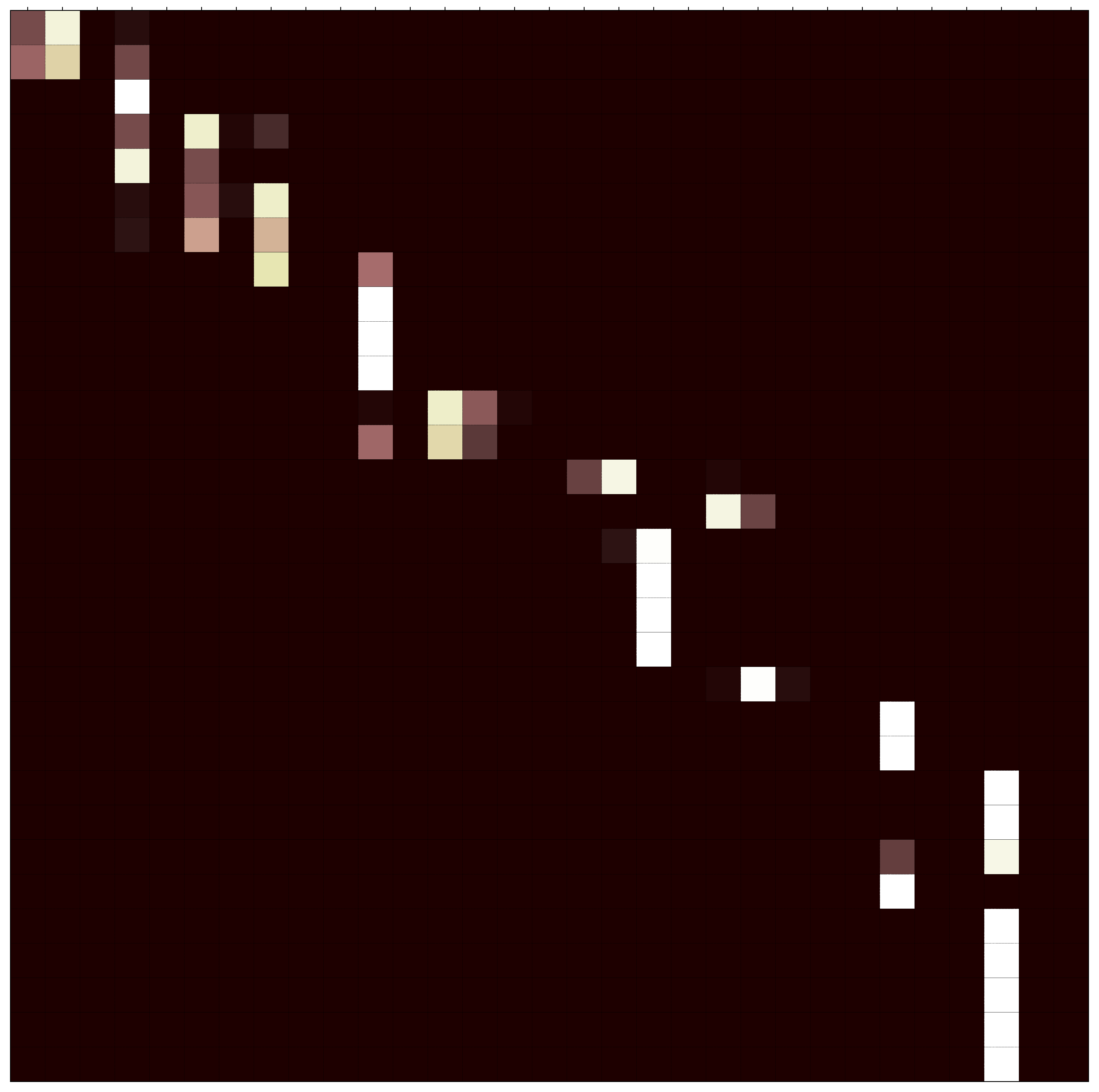}
\includegraphics[width=0.15\textwidth]{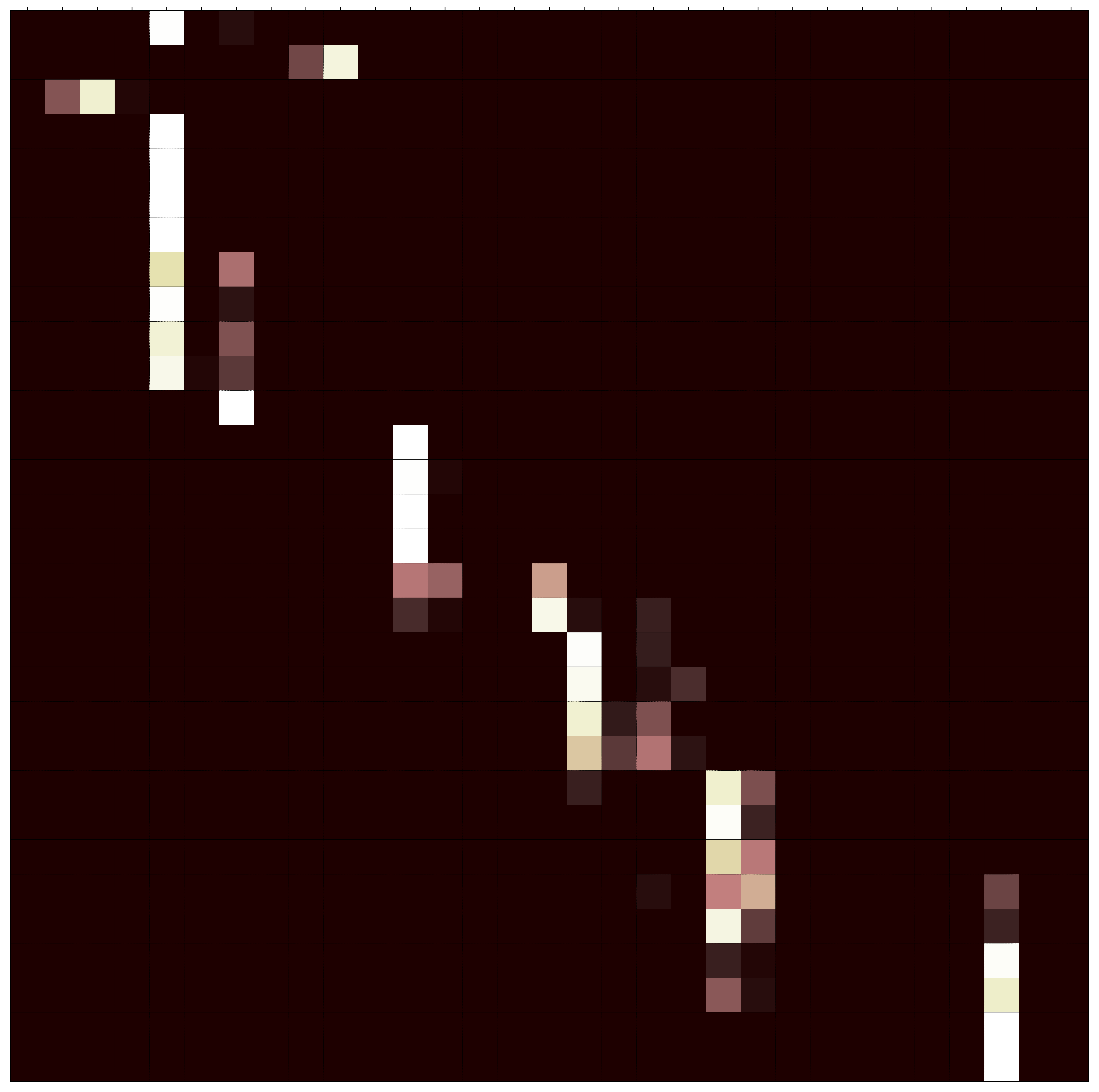}
\includegraphics[width=0.15\textwidth]{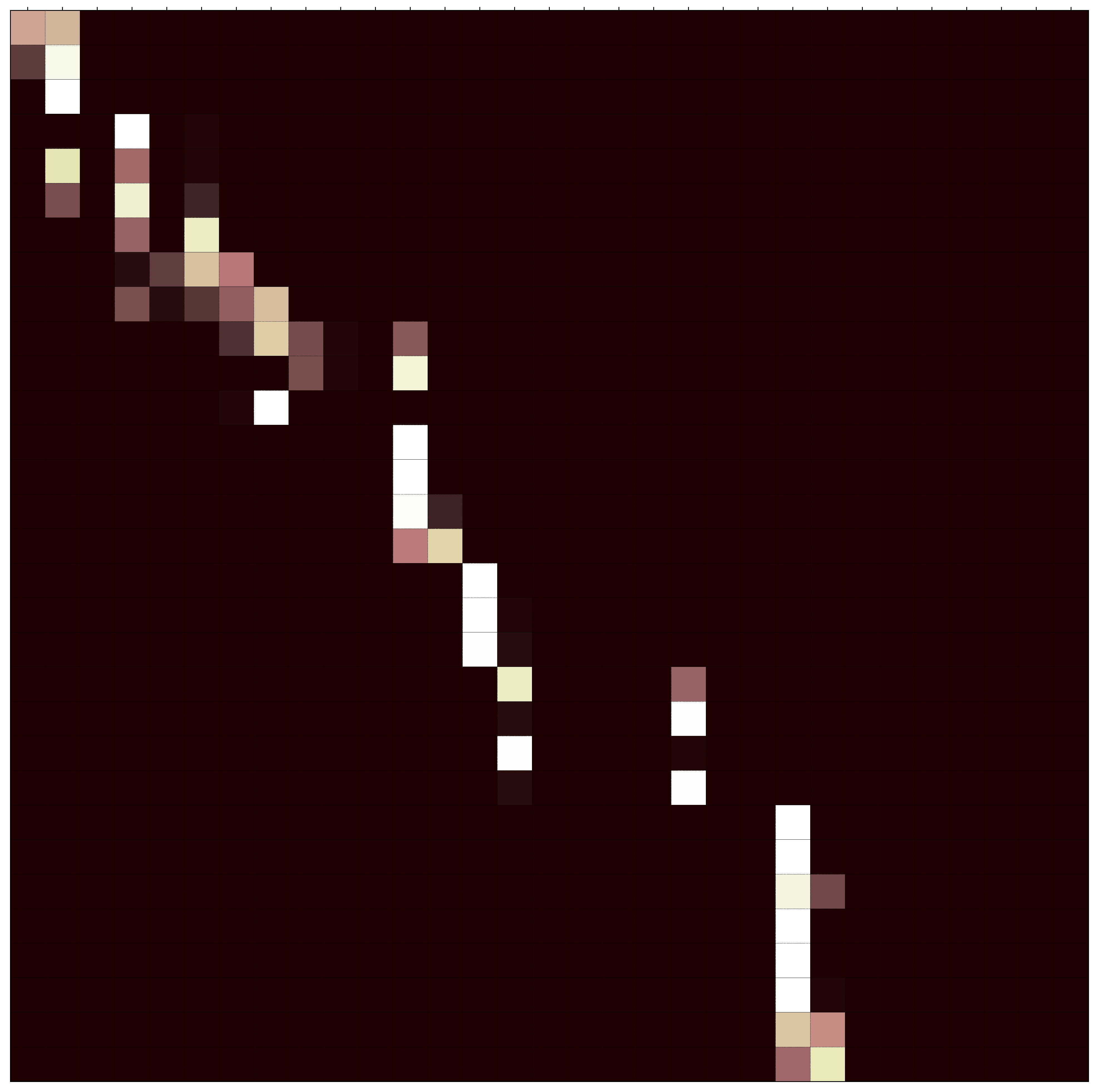}
\includegraphics[width=0.15\textwidth]{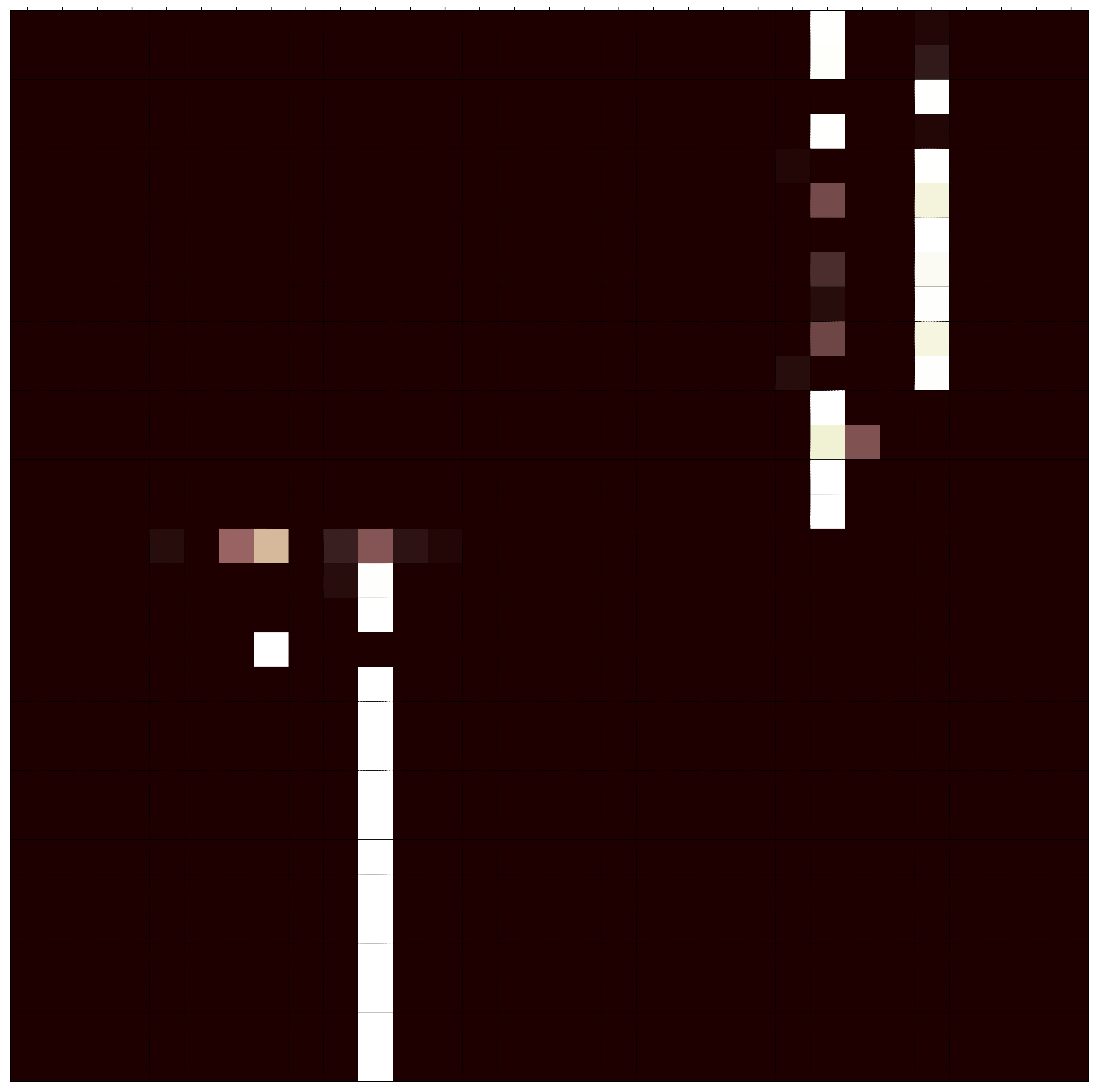}
\includegraphics[width=0.15\textwidth]{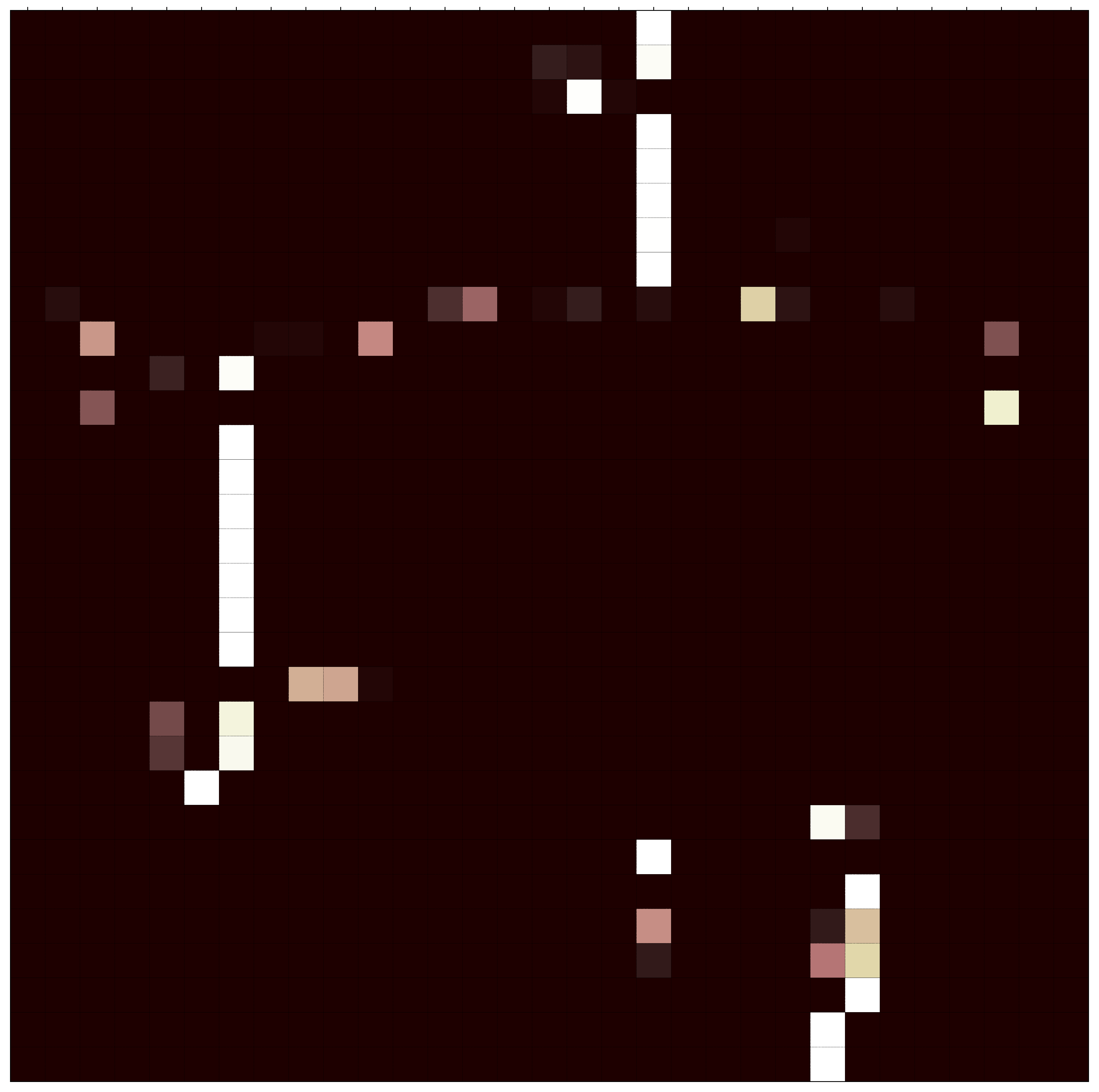}
\includegraphics[width=0.15\textwidth]{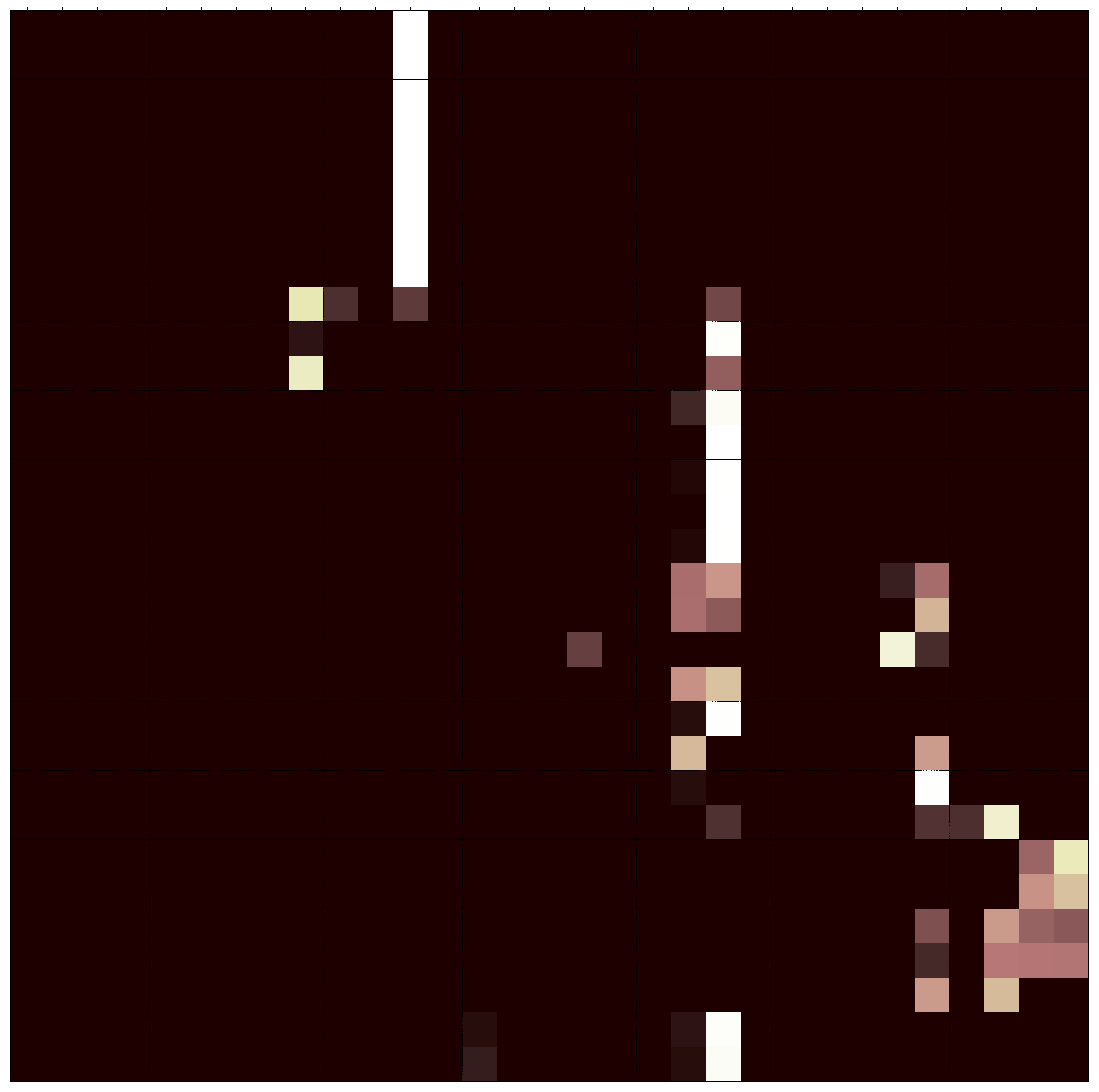}
\includegraphics[width=0.15\textwidth]{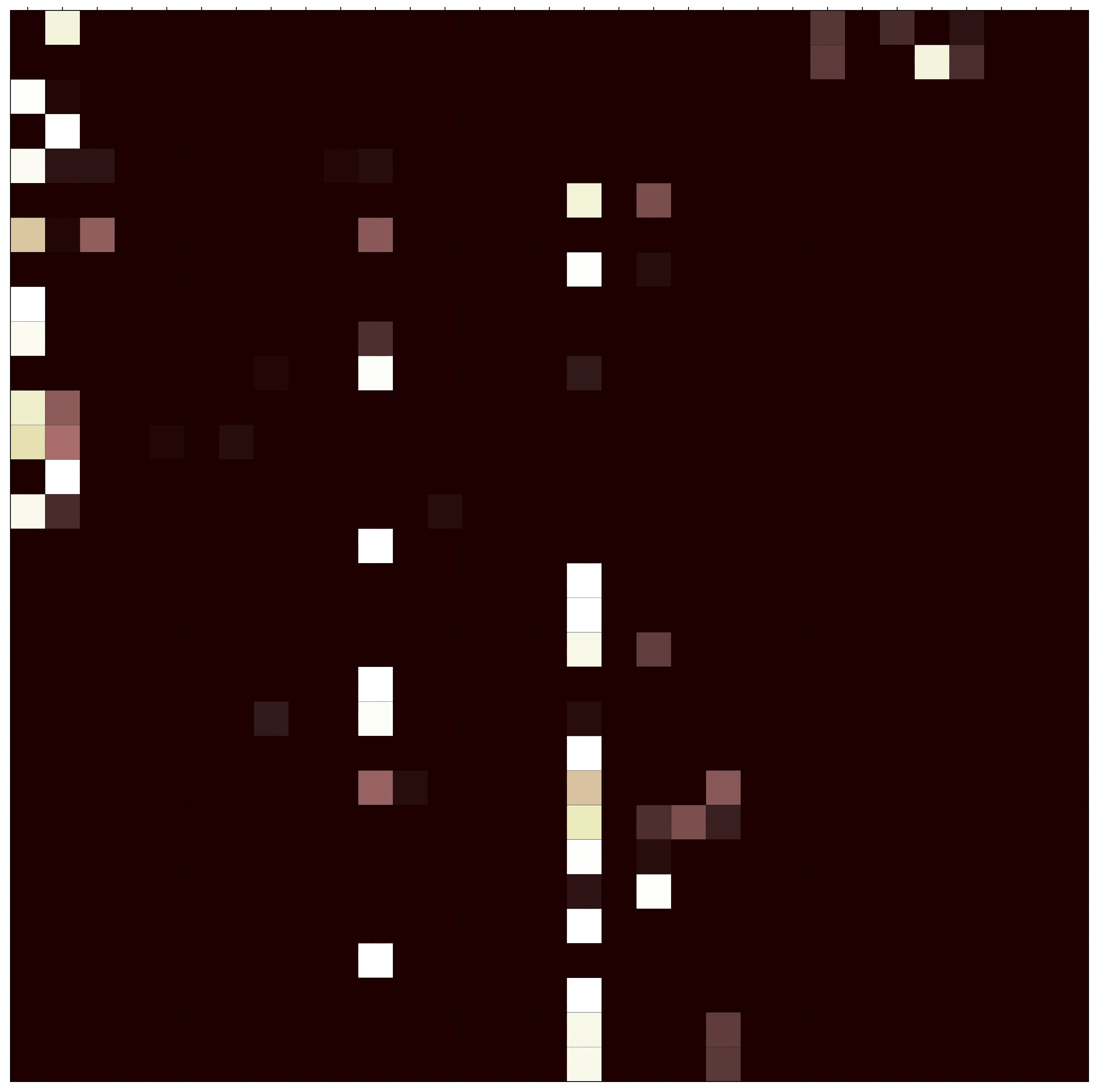}
\includegraphics[width=0.15\textwidth]{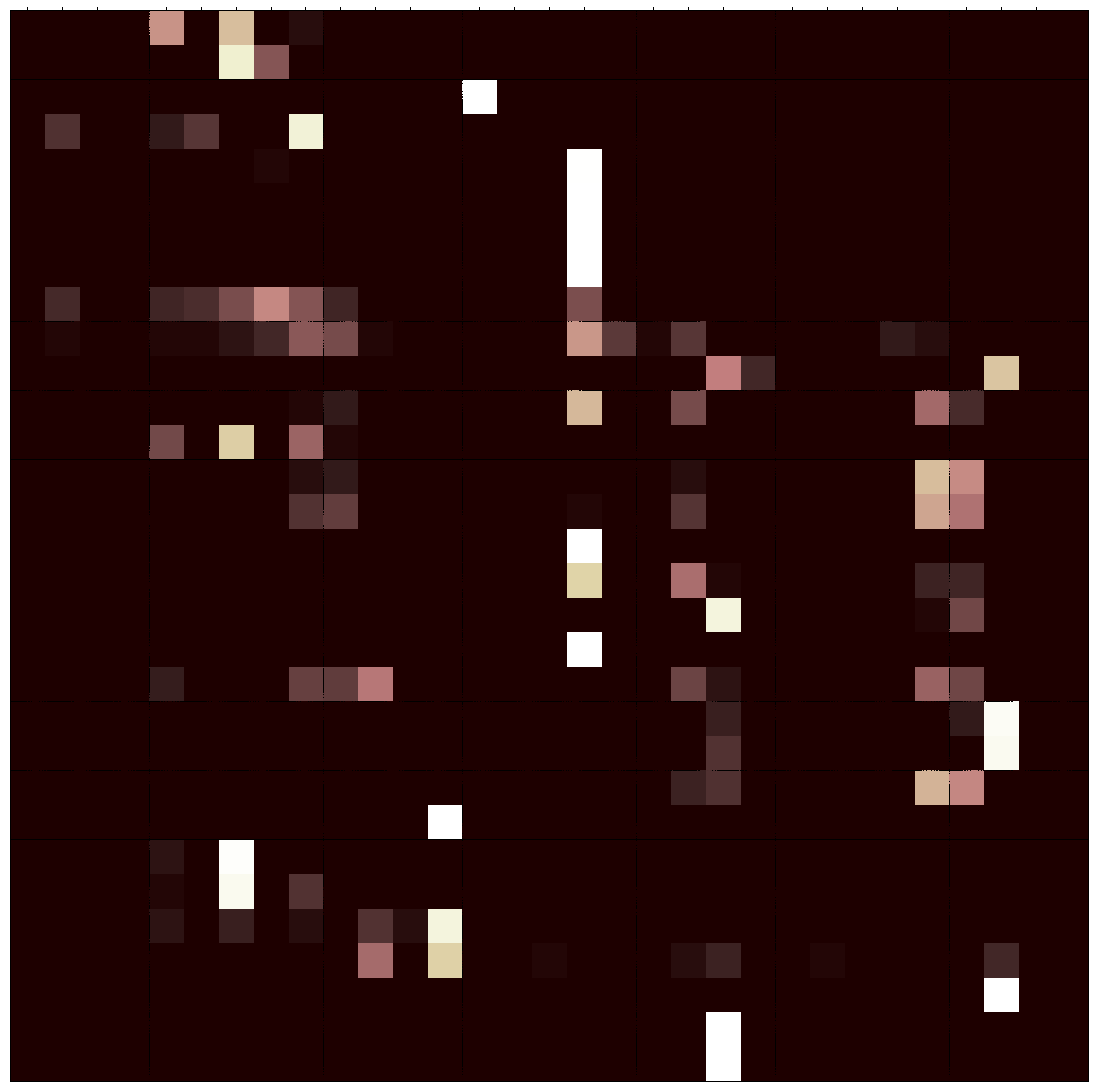}
\includegraphics[width=0.15\textwidth]{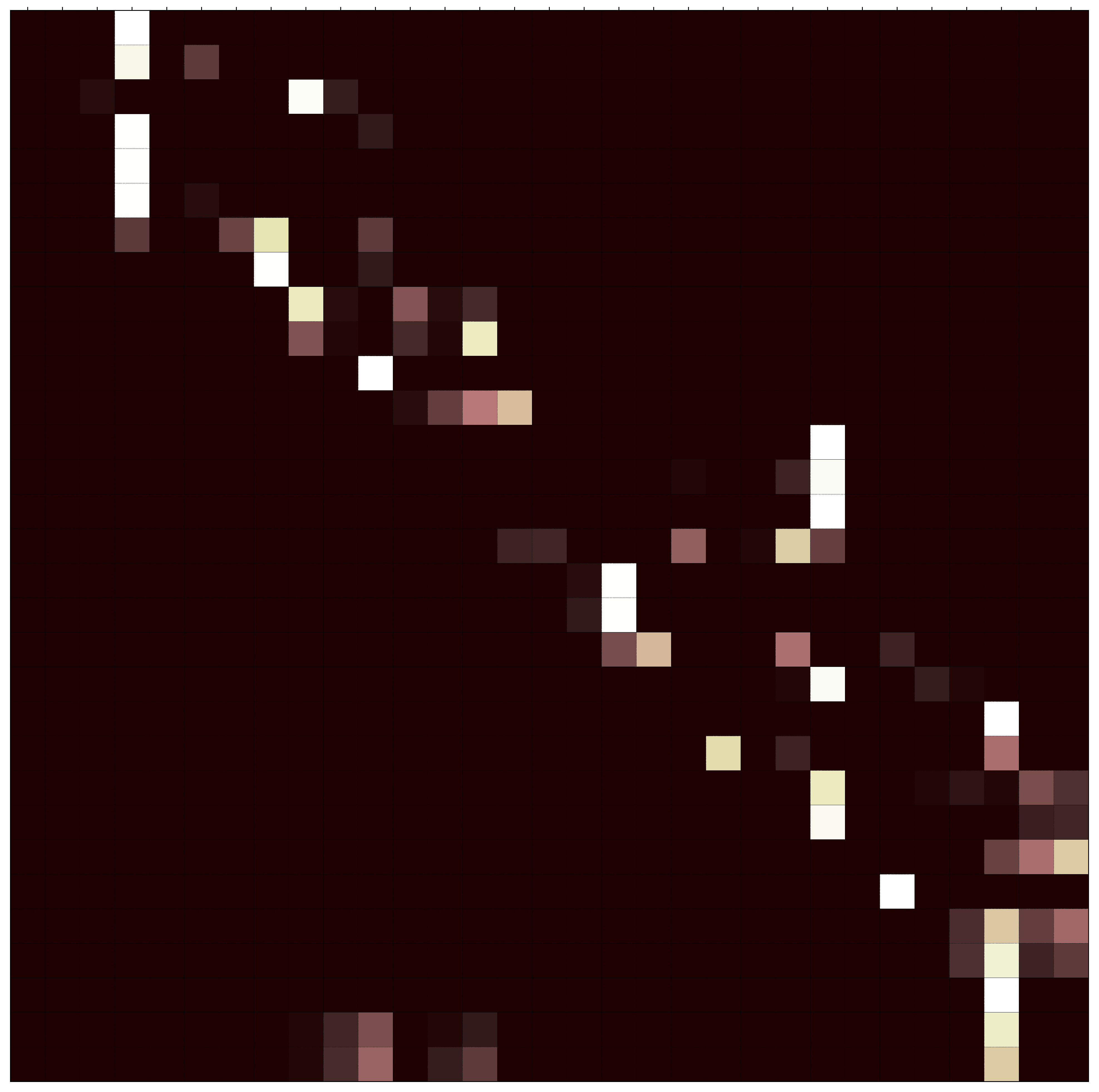}
\includegraphics[width=0.15\textwidth]{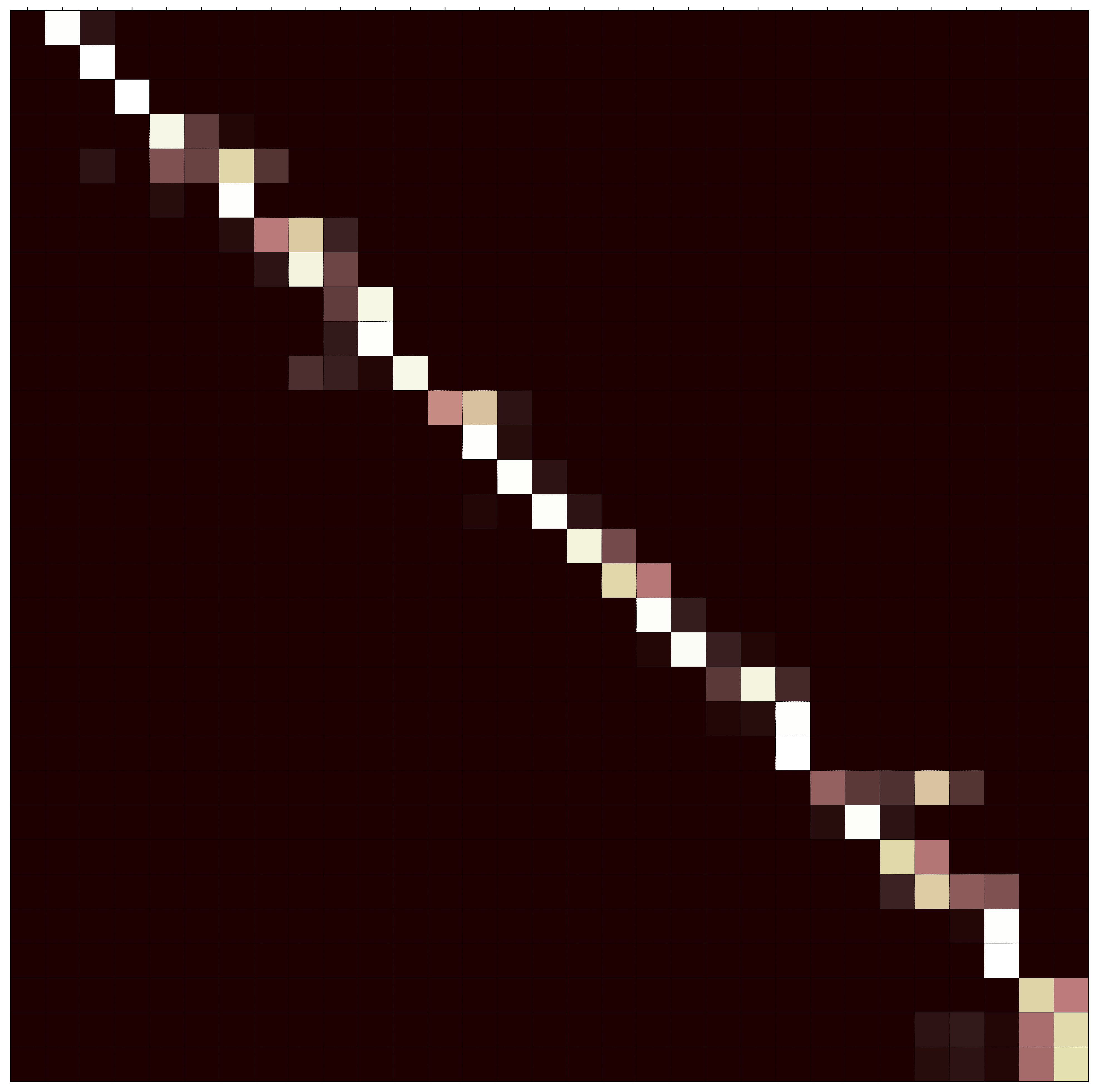}
\includegraphics[width=0.15\textwidth]{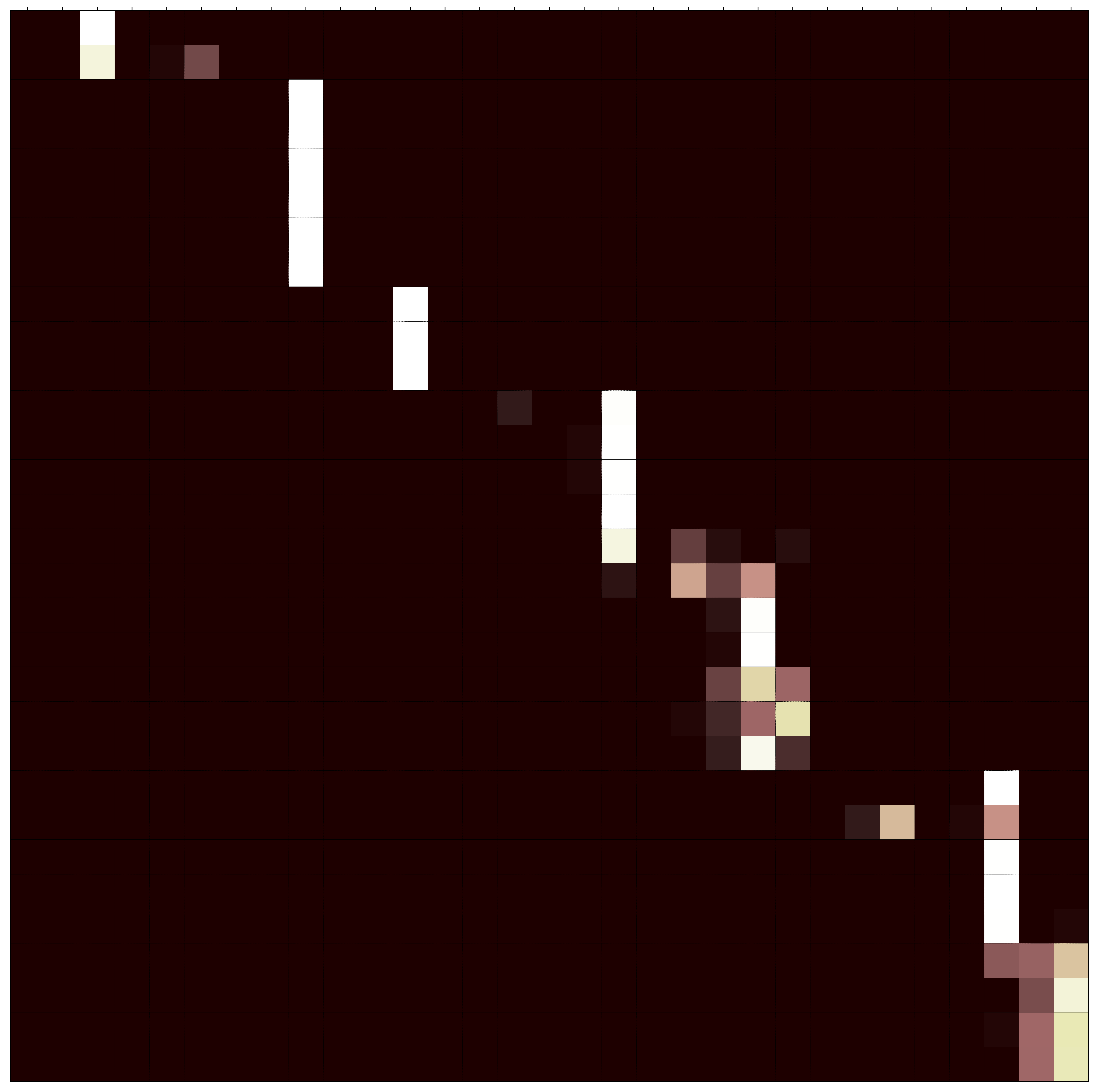}
\includegraphics[width=0.15\textwidth]{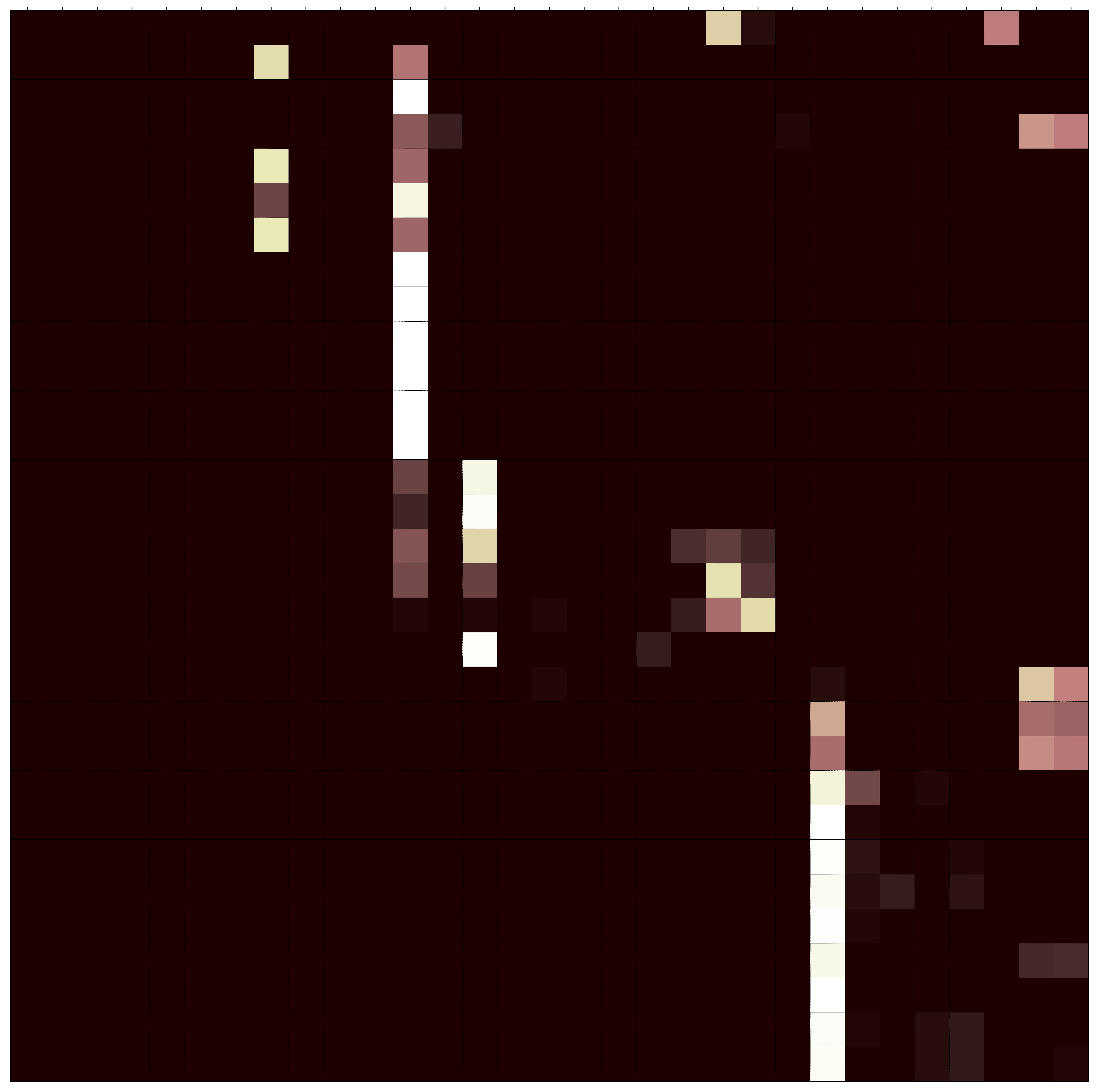}
\includegraphics[width=0.15\textwidth]{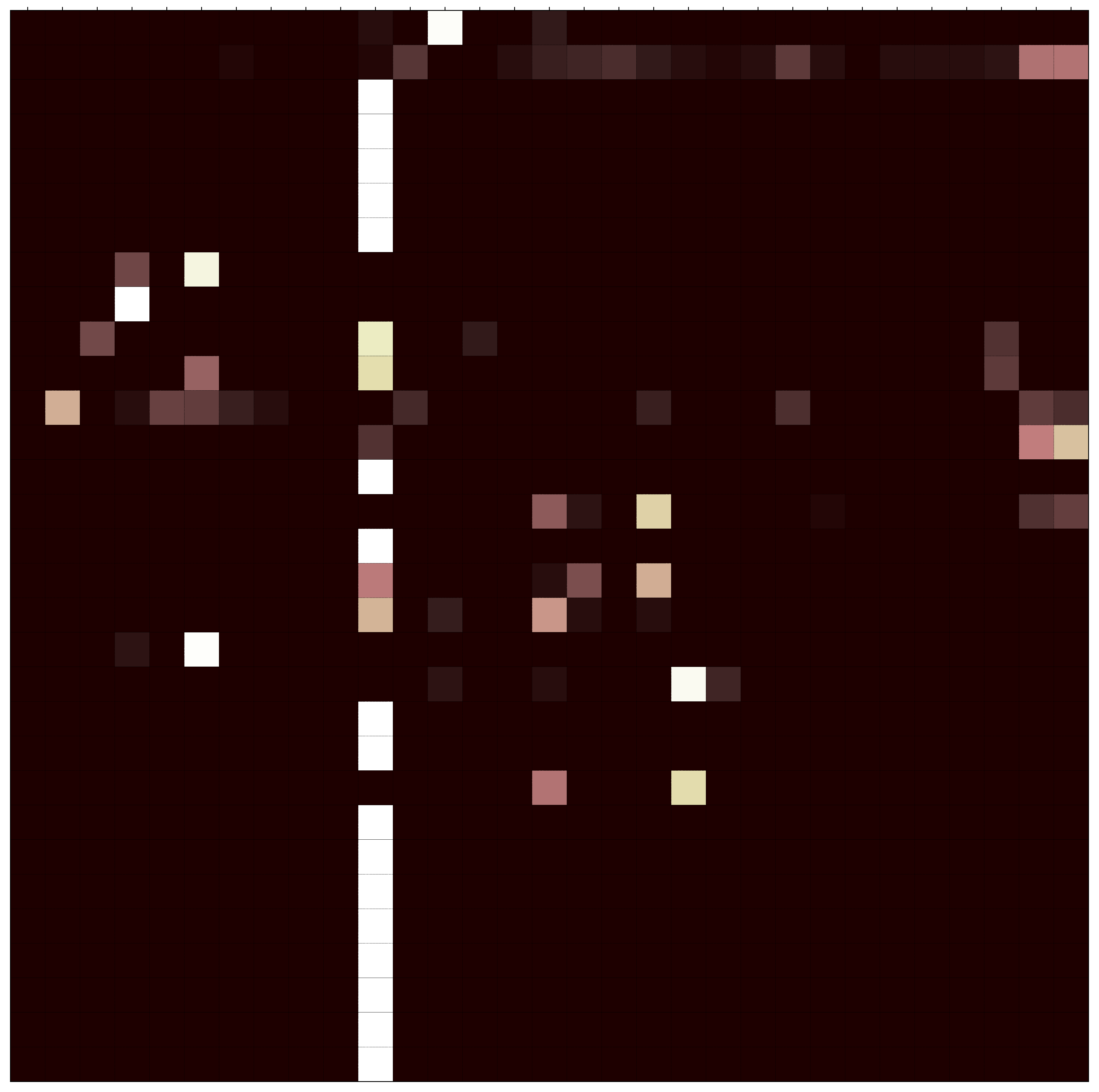}
\end{center}
\caption{Layer 6}
\end{figure}

\end{document}